\documentclass[accepted]{uai2021} 

\usepackage[american]{babel}

\usepackage{natbib} 
\usepackage{mathtools} 
\usepackage{booktabs} 
\usepackage{tikz} 

\usepackage{titling}
\usepackage{standalone}
\usepackage{mathrsfs}
\usepackage{booktabs}
\usepackage[switch]{lineno} 
\usepackage{xcolor}
\usepackage{amsmath,amssymb}
\usepackage{calrsfs}
\DeclareMathAlphabet{\pazocal}{OMS}{zplm}{m}{n}
\usepackage{upquote}

\newcommand{\swap}[3][-]{#3#1#2} 

\title{Uncertainty in Minimum Cost Multicuts for Image and Motion Segmentation}

\author{\href{mailto:Amirhossein Kardoost <kardoostamirhossein@gmail.com>?Subject=Your UAI 2021 paper}{Amirhossein Kardoost}{}\qquad Margret Keuper\\Data and Web Science Group
    
    University of Mannheim, Germany\\
}

\begin{document}


\maketitle

\begin{abstract}
  The minimum cost lifted multicut approach has proven practically good performance in a wide range of applications such as image decomposition, mesh segmentation, multiple object tracking and motion segmentation. It addresses such problems in a graph-based model, where real valued costs are assigned to the edges between entities such that the minimum cut decomposes the graph into an optimal number of segments. Driven by a probabilistic formulation of minimum cost multicuts, we provide a measure for the uncertainties of 
   the decisions made during the optimization. We argue that the access to such uncertainties is crucial for many practical applications and conduct an evaluation by means of sparsifications on three different, widely used datasets in the context of image decomposition (BSDS-500) and motion segmentation (DAVIS$_{2016}$ and FBMS$_{59}$) in terms of variation of information (VI) and Rand index (RI).
\end{abstract}

\section{Introduction}
\label{sec:introduction}

The minimum cost (lifted) multicut problem (\cite{chopra-1993})
, also known as correlation clustering, has been widely applied in computer vision for applications ranging from image segmentation~(\cite{Keuper2015}) to multiple person tracking~(\cite{margret_tpami_2020}) and pose estimation~(\cite{pishchulin2016cvpr}). The formulation is flexible and can easily be adapted to a new clustering problem. Specifically entities are represented by nodes in a graph and real valued costs are assigned to edges where positive costs indicate that the adjacent nodes are similar and negative costs indicate that the adjacent nodes are dissimilar. 
The minimum cost multicut then provides a decomposition of the graph into an optimal number of segments.
\citet{andres-2012-globally}, relate the minimum cost multicut to a probabilistic model, which was further extended in \cite{Keuper2015} to \textit{lifted} multicuts.
They show that, if the edge costs are set according to logits of the cut probability between nodes, the minimum cost multicut provides the maximum a posteriori probability (MAP) estimate. Motivated by this finding, edge cost definitions based on cut probability estimates have become common practice, even in settings where the final solution is estimated using primal feasible heuristics~\cite{Keuper_15,fusionMoves} to account for the NP-hardness of the problem~(\cite{Bansal2004}). 

\begin{figure}[t]
    \centering 
        \begin{tabular}{@{}c@{}c@{}c@{}c@{\hspace{0.1cm}}l@{}}
            \includegraphics[height=1.45cm]{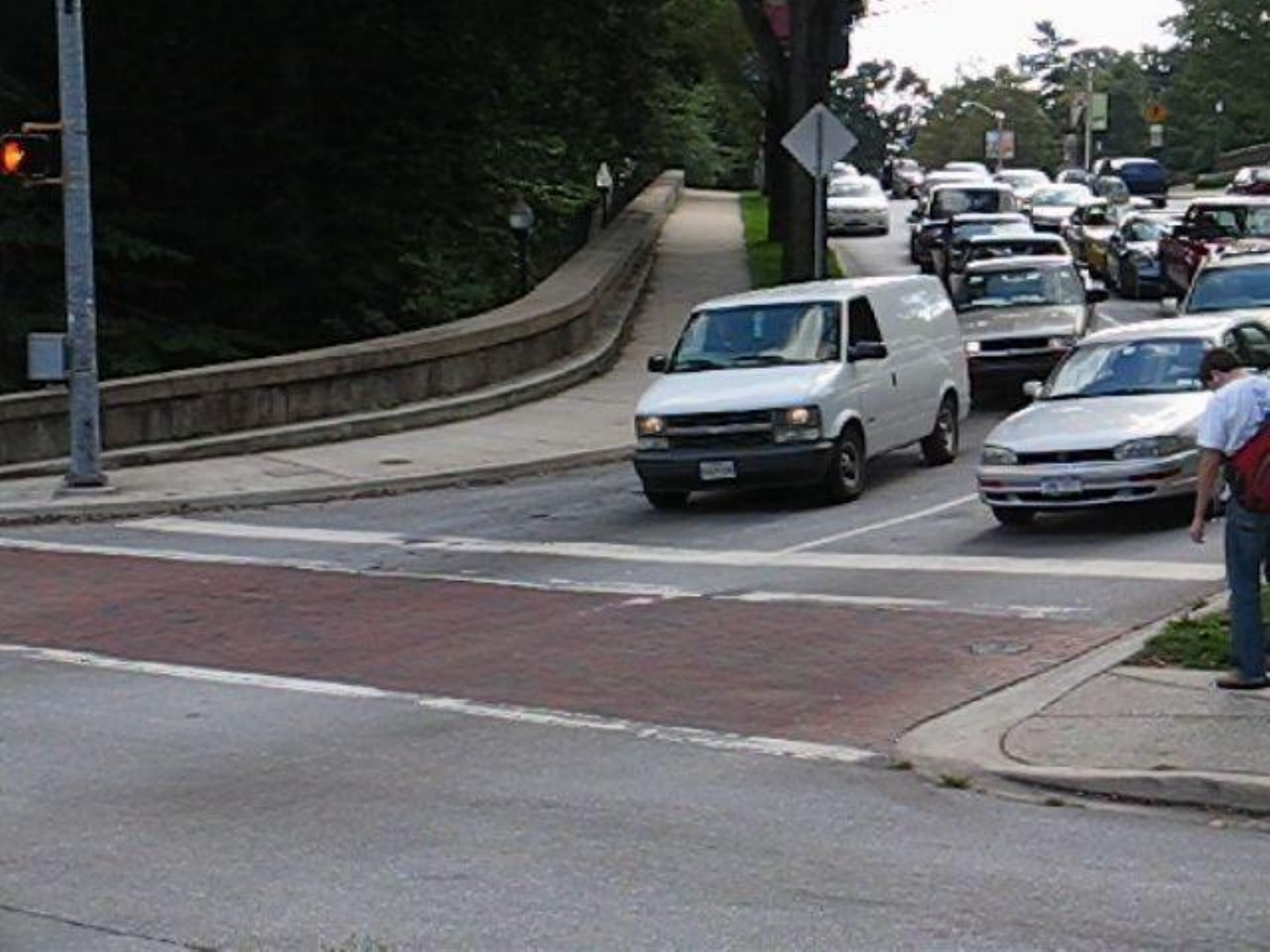}&
            \includegraphics[height=1.45cm]{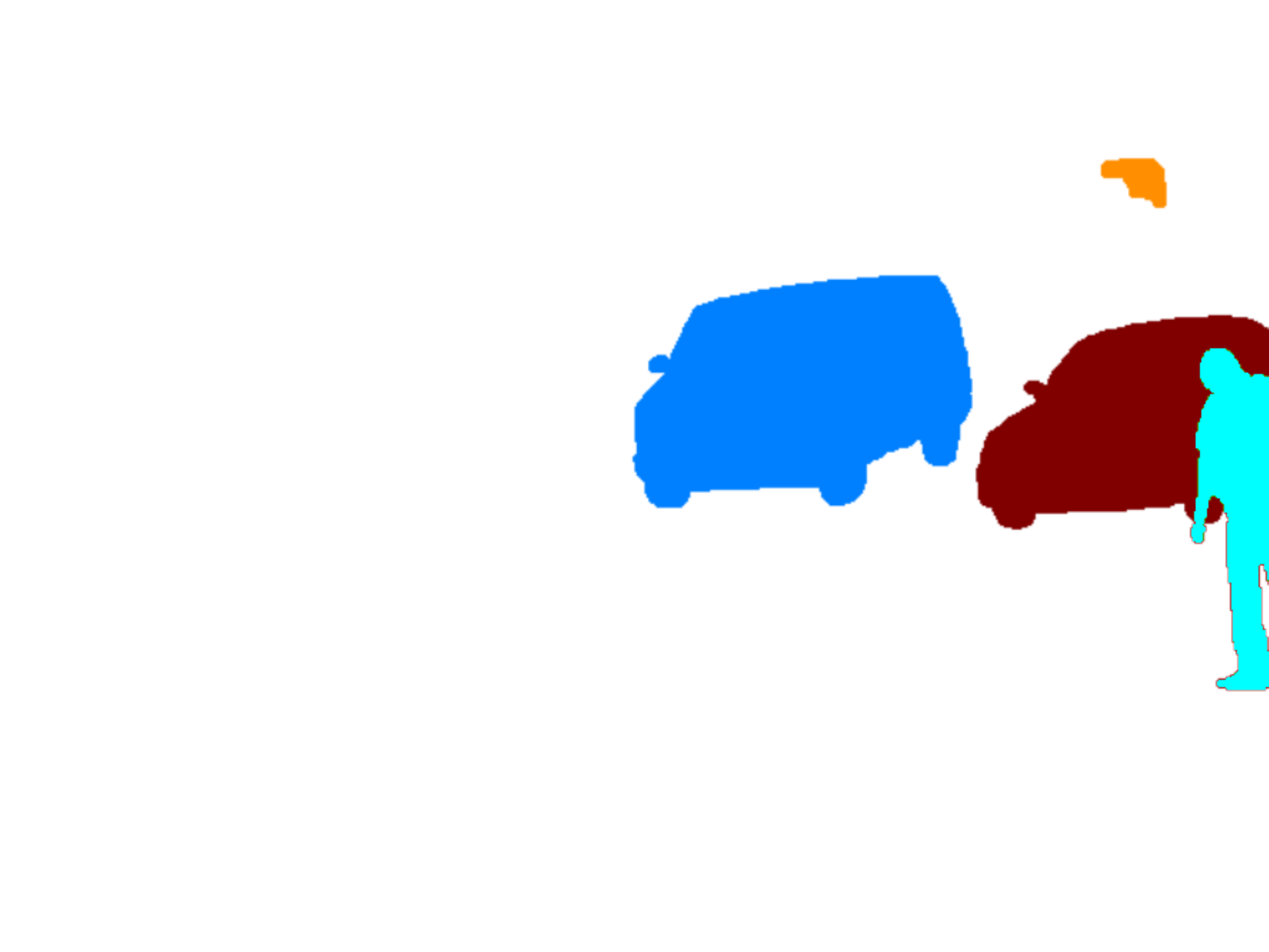}&
            \includegraphics[height=1.45cm]{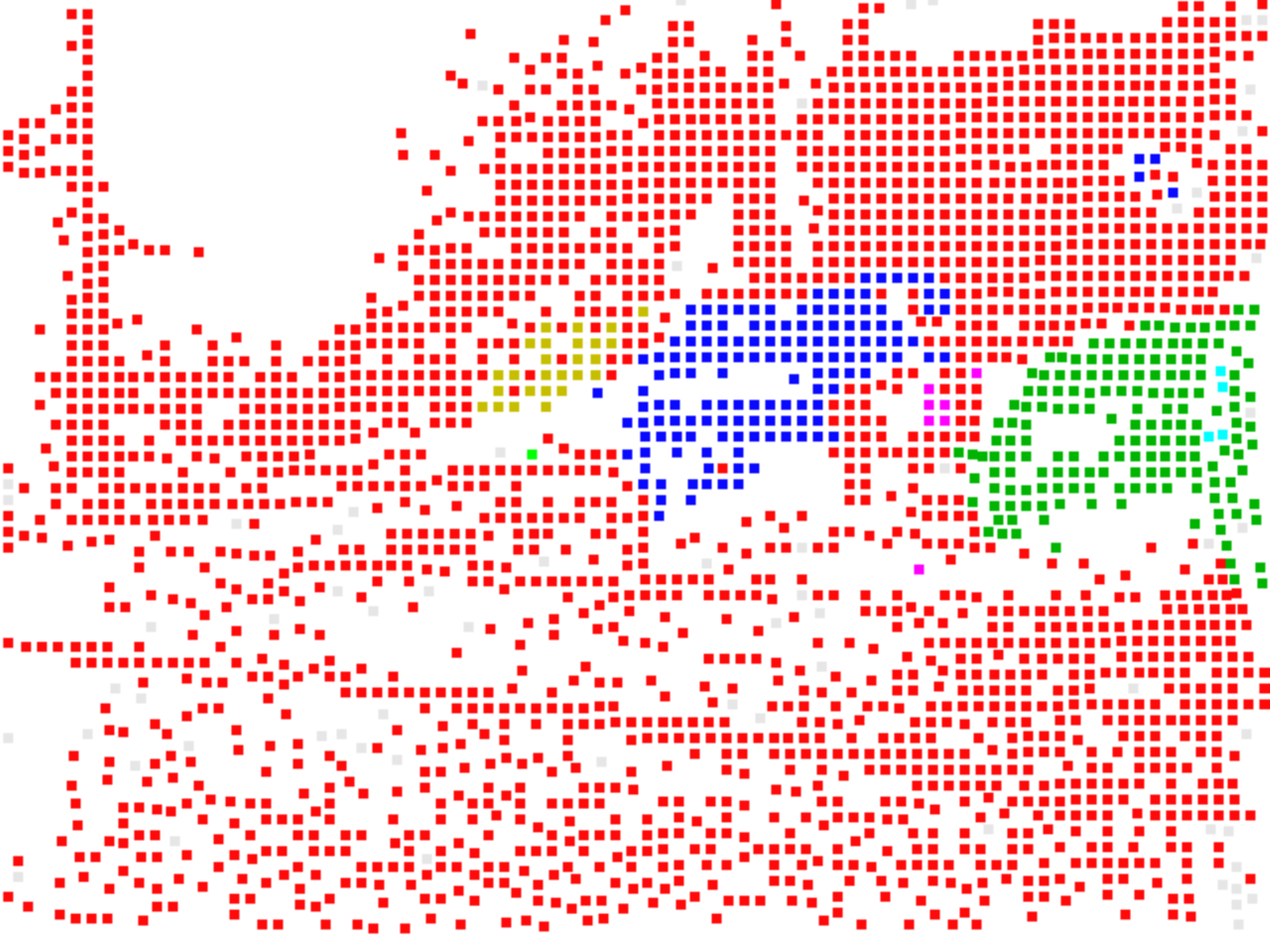}&
            \includegraphics[height=1.45cm]{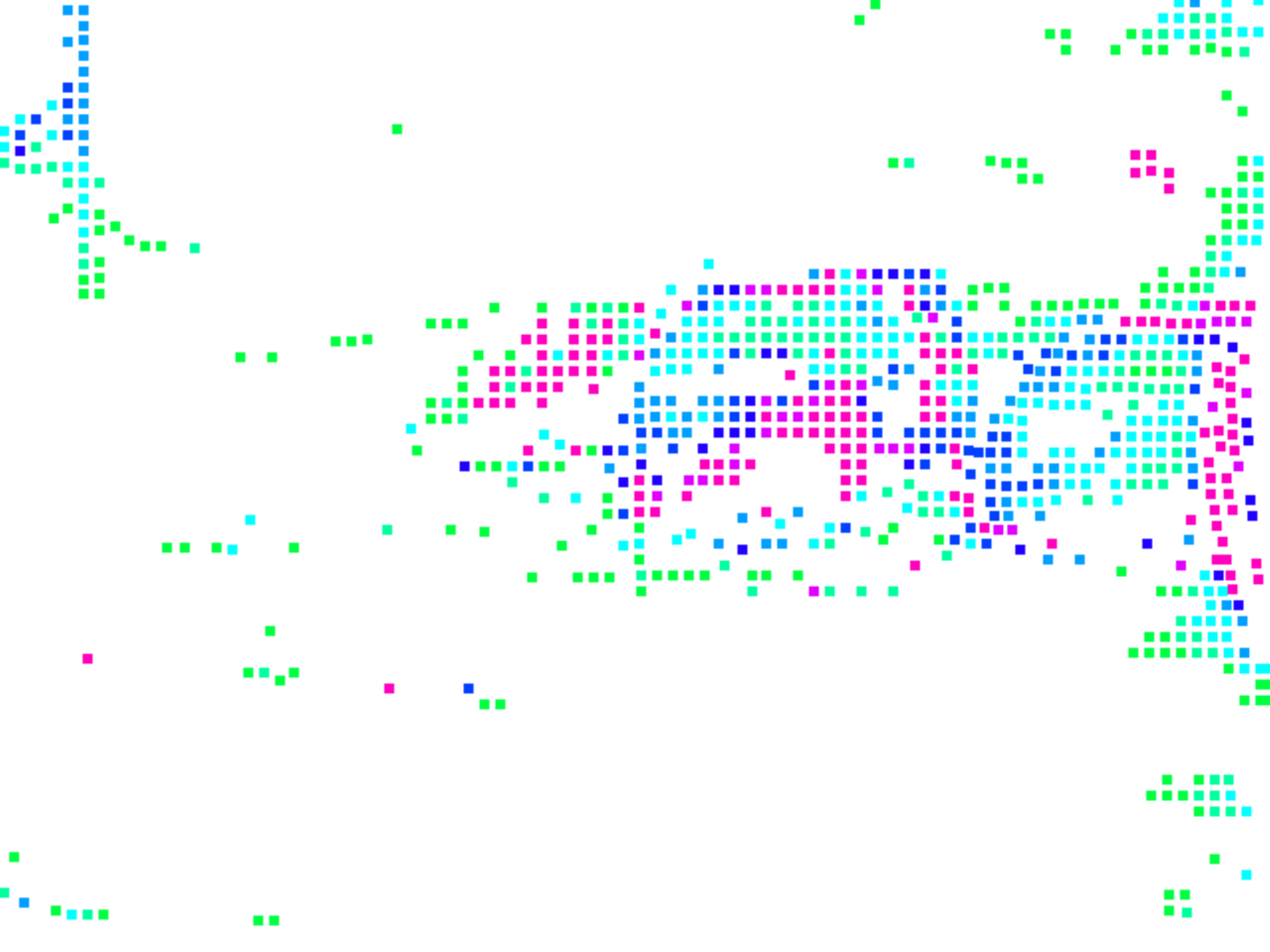}&
            \includegraphics[height=1.45cm]{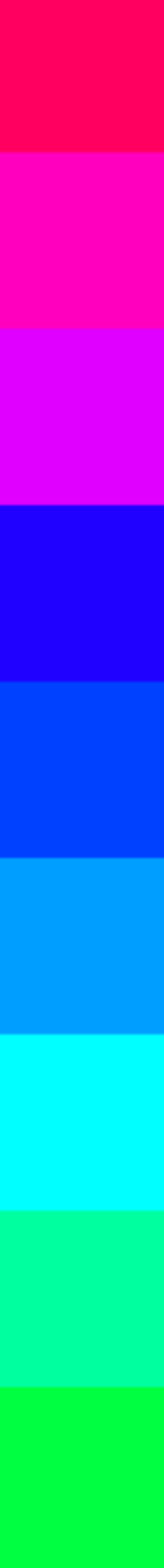}\\
            image&GT&multicut&uncertainty
        \end{tabular}
    \caption{Motion segmentation and the proposed uncertainty measure on a street scene. The uncertainty is high on incorrectly segmented points, specifically on the missed person.}
    \label{fig:teaser}
\end{figure}

In this paper, we argue that considering solely the (approximate) MAP solution is not satisfying in all scenarios. For example 
in an automotive setting, it might be required to assess the uncertainty in the prediction of the number of moving objects and their shapes. Therefore, we propose to employ the probabilistic model from \cite{andres-2012-globally} and derive, for a proposed solution, a measure for the uncertainty of each node-to-label assignment. 
An example is provided in Fig.~\ref{fig:teaser} for sparse motion segmentation. Based on sparse point trajectories, we apply the motion model derived by~\cite{Ochs14} and provide the minimum cost multicut solution (\cite{Keuper_15}). Depending on the exact parameters employed, the person standing nearby the street can not be correctly segmented. Yet, the uncertainty indicates potential mistakes in the prediction.

The proposed uncertainty measure is directly derived from the probabilistic multicut formulation and can be applied to \emph{any} given decomposition. We evaluate our approach in the context of minimum cost multicuts for sparse motion segmentation using the model from \cite{Keuper_15} on the datasets FBMS$_{59}$~\cite{Ochs14} and DAVIS$_{2016}$~\cite{davis_16}. 
Further, we investigate potential benefits of the predicted uncertainties for the generation of dense motion segmentations from sparse ones. Such densifications can be computed using convolutional neural networks trained in a self supervised way for example using the approach by~\cite{selfsupervised}.
Last, we evaluate the proposed uncertainty measure in the context of minimum cost \emph{lifted} multicuts for image decomposition on BSDS-500~\cite{BSDS500}. In both applications, motion and image segmentation, we show via sparsification plots that subsequently removing uncertain predictions from the solution improves segmentation metrics. The proposed measure is thus a robust indicator for the uncertainty of a given solution. 

\section{Related Work}
\label{sec:related_work}
We first review prior work related to the minimum cost multicut problem, then, we summarize related work w.r.t. the uncertainty estimation and last, we summarize prior work on multicuts in the considered application domains.

\paragraph{Correlation Clustering}
\label{subsec:multicut}
The minimum cost multicut problem~\cite{chopra-1993}, also known as correlation clustering, is a binary \emph{edge} labeling problem defined on a graph with real valued edge costs. 
The feasible solutions of the multicut problem propose decompositions of the graph and the optimal solution corresponds to the maximum a posteriori probability (MAP) estimate \cite{andres-2012-globally}.
Yet, the problem is shown to be APX-hard~\cite{Bansal2004}. While optimal solutions or solutions within bounds of optimallity can be found for small instances~\cite{demaine-2006,swoboda-2017,andres-2011,andres-2012-globally}, most practical applications depend on heuristic solvers \cite{Keuper2015,CGC,fusionMoves}. 
Our proposed uncertainty measure is defined on the formulation of the minimum cost (\textit{lifted}) multicut problem and can be applied on any given solution. We perform a thorough evaluation in the context of the widely used KLj and GAEC heuristics from \cite{Keuper2015} (compare the supplementary material for GAEC). 
Probabilistic clustering and segmentation algorithms, such as the minimum cost (\textit{lifted}) multicut have been studied on different applications, such as motion segmentation~\cite{margret_tpami_2020}, image decomposition~\cite{Keuper_15,andres-2011,HOsegMC,node_agglom}, multiple object tracking~\cite{Keuper_17,hornakova-2017,People_Tracking,Lifted_Disjoint}, connectomics~\cite{neurite_seg}, cell tracking~\cite{Rempfler} and instance segmentation~\cite{InstanceCut}. For all of these applications, one can easily imagine use-cases that benefit from a measure of uncertainty. 
We evaluate the proposed approach on image and motion segmentations.

%
\paragraph{Uncertainty Estimation} 
\citet{Torr} proposed a measure of uncertainty for the graph cut solutions using the min-marginals associated with the label assignments in a MRF. With respect to minimum cost multicuts, \citet{multicut_perturb} measure the uncertainty of image partitions by an approximate marginal distribution using cost perturbations and induced \emph{edge} label flips. The proposed approach is complementary to this method as it does not measure the uncertainty of a binary edge labeling but assesses the uncertainty of the induced \emph{node} labeling. 
In~\cite{Low_Dim_Perturb}, the Perturb and MAP (PM) approach is used for learning a restricted Boltzmann machine. In this method, each of the observable and hidden variables are flipped in order to check the change in the energy function. 
We propose a measure of uncertainty on the cut/join decisions of the graph in minimum cost (\textit{lifted}) multicut formulation. To show the applicability of our uncertainty measure in real-world scenarios we evaluate on two \textit{motion segmentation} and one \textit{image decomposition} benchmarks. 

%
\paragraph{Motion Segmentation}
\label{subsec:related_motion_seg}
Formulations of minimum cost multicut problems have been successfully used for motion segmentation for example in \cite{Keuper_15,Keuper_17}. The goal in this application is to segment motion patterns of the foreground objects with respect to the scene and irrespective of the camera motion, scaling movements and out-of-plane rotation of the objects~\cite{Ochs14}.
One widely used paradigm to tackle this problem is to define spatio-temporal curves, called \textit{motion trajectory}. The trajectories are created by tracking the points through consecutive frames using the optical flow estimation. Due to the high computational complexity, one often considers sparse motion segmentation~(\cite{Ochs14}), where point trajectories are not sampled at every point but at a defined density. Therefore, the segmentations do not cover all the image pixels. The trajectories are then cast to the nodes in a graph and their affinities are used to compute costs on the edges.
%
%
In this setting, the solution to the motion segmentation problem is sparse. There are different methods to provide dense motion segmentations from the sparse results, like the variational approach from~\cite{Ochs14} and the self-supervised deep learning based approach from~\cite{selfsupervised}. We study the proposed uncertainty measure on motion segmentation in the datasets FBMS$_{59}$~\cite{Ochs14} and DAVIS$_{2016}$~\cite{davis_16}. Additionally, we use uncertainties to improve the training of the densification model of~\cite{selfsupervised}.

\vspace{0.1cm}
\noindent\textbf{Image Decomposition.\quad}
\label{subsec:related_image_decompose}
For image decomposition, the minimum cost multicut problem is defined over sets of pixels or superpixels 
for example in~\cite{BSDS500,Keuper2015,HOsegMC,globally_optimal,andres-2011}. In this scenario, the pixels act as nodes in the graph and their connectivity defined by edges~\cite{andres-2011}, i.e. directly neighboring pixels are connected. The extension of this problem using \textit{lifted} mulitucts is defined in~\cite{Keuper2015}, where lifted edges are used to encode long-range information while the connectivity of the original graph is preserved. This can lead to an improved pixel-level clustering behavior. \citet{HOsegMC} use a higher order multicut model on superpixels for the task. 
Orbanz and Buhmann~\cite{Bayesian_img_seg} provide a non-parametric Bayesian model for histogram clustering to determine the number of image segments, utilizing Dirichlet process mixture model. Cutting plane and integer programming techniques are used in~\cite{globally_optimal} for image decomposition. Further they proposed an approximate solution by solving a polynomial LP.

\section{Uncertainties in Minimum Cost (Lifted) Multicuts}
\label{sec:method}
	
In this section, we formally define the minimum cost multicut problem (MP)~(\cite{chopra-1993}) and its generalization, the minimum cost \emph{lifted} multicut problem (LMP)~(\cite{Keuper2015}). Then, we summarize the probabilistic model from \cite{andres-2012-globally} and deduce the proposed uncertainty measure. 


\subsection{The Minimum Cost Multicut Problem} 
\label{sub:mcmp}

Given a graph $G = (V, E)$, a cost function $c: E \rightarrow \mathbb{R}$ and \emph{edge} labels $y:~E~\rightarrow~\{0,~1\}$, the optimization problem in \eqref{eq:MP} is an instance of the minimum cost multicut problem (MP) with respect to the graph $G$ and real-valued costs $c$
\begin{align}
\min\limits_{y \in \{0, 1\}^E}
\sum\limits_{e \in E} c_e y_e
\label{eq:MP}
\end{align}
\begin{align}
s.t. \quad \forall C \in cycles(G) \quad \forall e \in C : y_e \leq \sum\limits_{e^\prime \in C\backslash\{e\}} y_{e^\prime} .
\nonumber
\end{align}
The edge labeling $y$ induces a decomposition of graph $G$, which is ensured by the inequality constraints stated over all cycles of $G$. \citet{chopra-1993} showed that is sufficient to consider all chordless cycles. The minimum cost multicut problem allows to assign a cost ($c_e$) for every edge $e\in E$, where a positive cost encourages the edge to be joined while a negative cost encourages the edge to be cut.
	
\subsection{Minimum Cost Lifted Multicuts} 
\label{sub:mclmp}
The minimum cost \emph{lifted} multicut problem (LMP) is defined with respect to a graph $G=(V,E)$ and a lifted graph $G'=(V,E')$ with $E\subseteq E'$ and a cost function $c': E' \rightarrow \mathbb{R}$~(\cite{Keuper2015}). The cost function $c'$ allows to assign to every edge in $E'$ a cost for being cut. The decompositions of graph $G$ relate one-to-one to the feasible solutions of the problem, similar to the minimum cost multicut problem.
    
For any undirected graph $G=(V,E)$, any $F=\binom{V}{2} \setminus E$ and any $c':E'=E\cup F \rightarrow \mathbb{R}$, the linear program written in Eq.~\eqref{eq:LMC1} - \eqref{eq:LMC4} is an instance of the minimum cost \textit{lifted} multicut problem with respect to the graph $G$, lifted edges $F$ and the edge costs $c'$ ~\cite{Keuper2015}.
\begin{align}
    \min\limits_{y \in \{0, 1\}^{E^{\prime}}}
    \sum\limits_{e \in E^{\prime}} c'_e y_e
    \label{eq:LMC1}
\end{align}
\begin{align}
    s.t. \quad \forall C \in cycles(G) \quad \forall e \in C : y_e \leq \sum\limits_{e^\prime \in C\backslash\{e\}} y_{e^\prime}\label{eq:LMC2}\\
%
    \forall vw \in F \quad \forall P \in vw\mbox{-}\mathrm{paths}(G) : y_{vw} \leq \sum\limits_{e\in P} y_{e} \\
%
    \forall vw \in F \quad \forall C \in vw\mbox{-}\mathrm{cuts}(G) : 1 - y_{vw} \leq \sum\limits_{e\in C} (1 - y_e)\label{eq:LMC4}
\end{align}
The linear inequalities in Eq. \eqref{eq:LMC2} - \eqref{eq:LMC4} constrain $y$ such that $\{ e\in E |y_e=1\}$ is a multicut of $G$. Further, they ensure that, for any edge $uv\in F, y_{uv} = 0$ if and only if there exists a path ($uv$-path) in the original graph $G$, along which all edges are connected, i.e. labeled as 0. The set $F$ of lifted edges is utilized to modify the cost function of the multicut problem without changing the set of feasible solutions. The lifted edges can thus connect non-neighboring nodes in graph $G$ and are commonly used to introduce non-local information.

\subsection{Probability Measures} 
\label{sub:optimi_problem}
\begin{figure}
    \centering
    \includegraphics[width=0.6\linewidth]{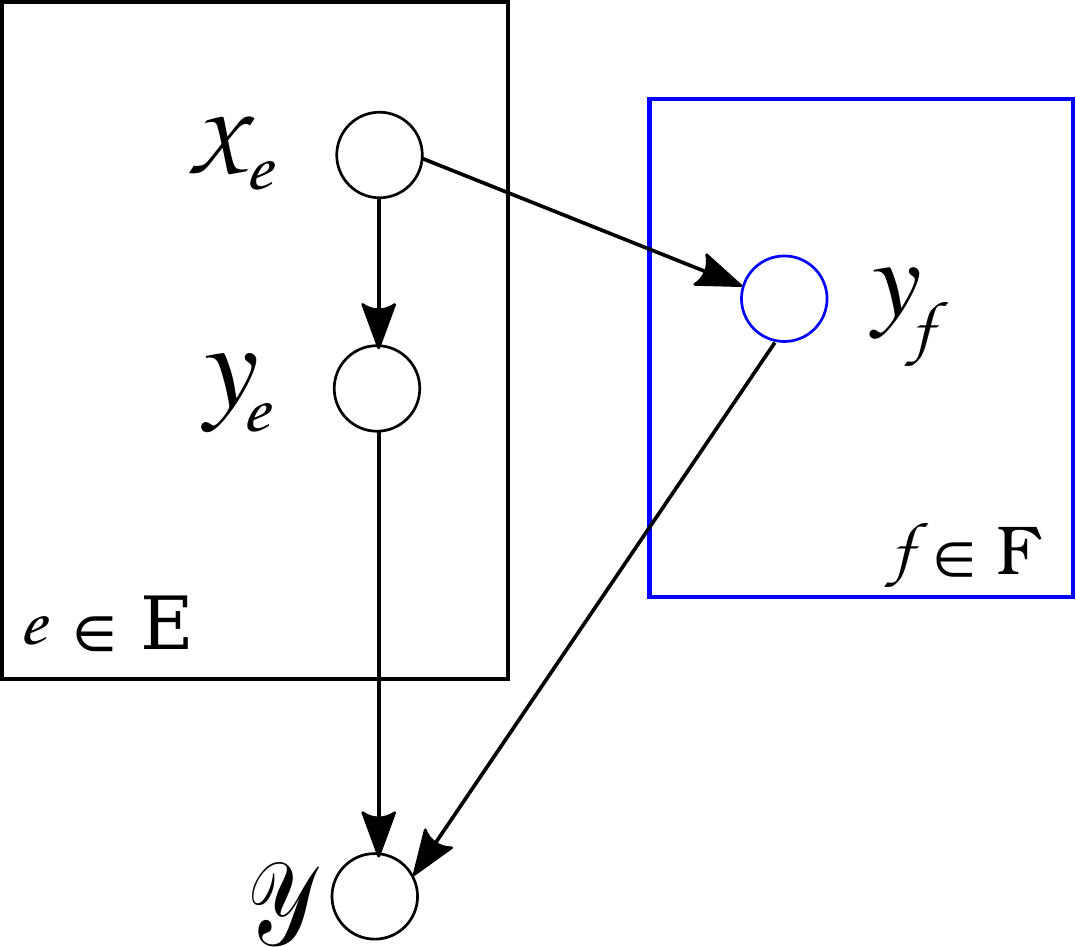}
    \caption{Bayesian Network from \cite{Keuper2015}, defining a set of probability measures on multicuts (black) and \textit{lifted} multicuts (blue).}
    \label{figure:bayesian-network} 
\end{figure}

\citet{andres-2012-globally} show that minimum cost multicuts can be derived from a Bayesian model (see Fig.\ref{figure:bayesian-network}) and their solutions correspond to the maximum a posteriori estimates. In the following, we summarize this probabilistic model and its extension to lifted multicuts from \cite{Keuper2015}, upon which we build the proposed uncertainty estimation.

\paragraph{Probability Measures on MP and LMP}

For a graph $G=(V,E)$, \citet{andres-2012-globally} assume that likelihoods $p_{\pazocal{X}_e|\pazocal{Y}_e}$ 
are computed based on an affinity definition of the nodes in graph $G$. 
Further, the costs assigned to the edges are assumed to be independent of each other and the topology of the graph $G$. Moreover, the prior $p_{(\pazocal{Y}_e)}$ is assumed to be identical for all edges $e\in E$ and is specified by a value $\beta\in(0,1)$, so that $p_{\pazocal{Y}_e=1}=\beta$ and $p_{\pazocal{Y}_e=0}=1-\beta$.
In this setting, they show that the minimum cost multicut maximizes the posterior probability $p_{\pazocal{Y} | \pazocal{X}, \mathscr{Y}}$  of a joint labeling $y\in\{0,1\}^{|E|}$. 
By definition, the Maximum a Posteriori (MAP) is then 
\begin{align}
    p_{\pazocal{Y} | \pazocal{X}, \mathcal{Y}} 
    \ & \propto \ 
    p_{\mathcal{Y} | \pazocal{Y}} \cdot \prod_{e \in E} p_{\pazocal{Y}_e | \pazocal{X}_e} \cdot p_{\pazocal{Y}_e} 
\enspace .
\label{eq:map}
\end{align}
for the MP, Fig.~\ref{figure:bayesian-network} (the black part)~(\cite{andres-2012-globally}), and it can be extended to the LMP, Fig.~\ref{figure:bayesian-network} (the blue part) as follows (\cite{Keuper2015})
\begin{align}
p_{\pazocal{Y} | \pazocal{X}, \mathcal{Y}}
    \propto 
        p_{\mathcal{Y} | \pazocal{Y}} 
        \cdot \prod_{e \in E} p_{\pazocal{Y}_e | \pazocal{X}_e} \cdot p_{\pazocal{Y}_e}
        \cdot \prod_{f \in F} p_{\pazocal{Y}_f | \pazocal{X}_E} \cdot p_{\pazocal{Y}_f}
\enspace .
\label{eq:lmap}
\end{align}
In both cases, $p_{\mathcal{Y} | \pazocal{Y}}$ indicates the feasibility of a solution, i.e. 
\begin{align}
p_{\mathcal{Y} | \pazocal{Y}}(Y_{E'}, y)
\ & \propto \ 
\begin{cases}
1 & \textnormal{if}\ y \in Y_{E'}\\
0 & \textnormal{otherwise}
\end{cases}
\enspace .
\label{eq:renormalization}
\end{align}
where $E'=E$ for the MP and $E'=E\cup F$ for the LMP.

Equations~\eqref{eq:map} and~\eqref{eq:lmap} can be maximized by minimizing their negative log-likelihoods. This leads to the definition of instances of the MP (Eq.~\ref{eq:MP}) and LMP (Eq.~\eqref{eq:LMC1}) by setting edge costs according to 
\begin{align}
\forall e \in E: \quad
c_e = \log \frac{p_{\pazocal{Y}_e | \pazocal{X}_e}(0, x_e)}{p_{\pazocal{Y}_e | \pazocal{X}_e}(1, x_e)} + \log \frac{1-\beta}{\beta}
\enspace .
\label{eq:weights-map}
\end{align}
The value of the scalar $\beta$ is the cut prior and assumed to be 0.5 for a bias-free case.

\subsection{Uncertainty Estimation Model}
\label{sec:uncertain_estim_model}

Based on the probabilistic formulation of the MP \eqref{eq:map} and LMP \eqref{eq:lmap}, we can study the uncertainty of given feasible solutions. More specifically, we aim to assign to every node in $V$ a confidence reflecting the certainty of the assigned label. The simplest attempt to this goal would be to directly assess the posterior probability of the solution. In the following, we briefly sketch this baseline approach before introducing the proposed measure.

\paragraph{Baseline Approach}
For every node $v_i\in V$ 
we propose, as a baseline approach, to draw an uncertainty measure from the proxi to the posterior probability (e.g. Eq.~\eqref{eq:map}). 
Thus, the confidence of the given label $A$ of node $v_i$ is 
\begin{align}
\displaystyle \prod_{e=(v_i,v_j), v_j\in A} p_{\pazocal{Y}_e | \pazocal{X}_e}(0, x_e) . \displaystyle \prod_{e=(v_i,v_j), v_j\in V\setminus A} p_{\pazocal{Y}_e | \pazocal{X}_e}(1, x_e).
\label{eq:uncertainty_eq4_2}
\end{align}
This measure intuitively aggregates the local join probabilities of all edges that are adjacent to $v_i$ and joined (set to 0) in the current solution, and the local cut probabilities of all edges adjacent to $v_i$ and cut (set to 1) in the current solution. Accordingly, Eq.~\eqref{eq:uncertainty_eq4_2} yields a low value if the local probabilities for the given solution are low.  

\textcolor{black}{A potential issue with this simple approach from Eq.~\eqref{eq:uncertainty_eq4_2} is that it under-estimates the confidence in many practical scenarios, especially when the local probabilities are uncertain and the connectivity is dense. Then, the product of local probabilities will issue a low value even if all local cues agree.} Therefore, we describe in the following the derivation of the proposed, calibrated uncertainty measure. 


\paragraph{Uncertainty Estimation.}

Given an instance of the (lifted) multicut problem and its solution, we employ the probability measures in equations \eqref{eq:map} (\textit{MP}) and \eqref{eq:lmap} \textit{(LMP)}. We iterate through nodes $v_i\in\{1, \dots, |V|\}$ in vicinity of a cut, i.e. $\exists e\in \pazocal{N}_{E}(v_i)$ with $e\in E$ and $y_e=1$.
Assuming that $v_i$ belongs to segment $A$ and its neighbour $v_j$ according to $E$ belongs to the segment $B$, the amount of cost change $\gamma_B$ is computed in the linear cost function (defined in~\eqref{eq:MP} and ~\eqref{eq:LMC1}) by moving $v_i$ from cluster $A$ to cluster $B$ as
%
\begin{align}
\gamma_B=\sum_{v_j\in\pazocal{N}_{E'}(v_i)\cap A}c_{(v_i,v_j)} - \sum_{v_j\in\pazocal{N}_{E'}(v_i)\cap B}c_{(v_i,v_j)}.
\label{eq:costDifference}
\end{align}
Thus, in $\gamma_{B}$, we accumulate all costs of edges from $v_i$ that are not cut in the current decomposition and subtract all costs of edges that are cut in the current decomposition but would not be cut if $v_i$ is moved from $A$ to $B$. 
Note that, while the cost change is computed over all edges in $E'$ for lifted graphs, only the uncertainty of nodes with an adjacent cut edge in $E$ can be considered in order to preserve the feasibility of the solution. For each node $v_i$ the number of possible moves depends on the labels of its neighbours $\pazocal{N}_{E}(v_i)$, and Eq.~\eqref{eq:costDifference} allows us to assign a cost to any such node-label change. Altogether, we assess the uncertainty of a given node label by the cheapest, i.e. the most likely, possible move
\begin{align}
\gamma_i = \min_B \gamma_B.
\label{eq:uncertain_node}
\end{align}
and set $\gamma_i$ to $\infty$ if no move is possible. 
The minimization in Eq.~\eqref{eq:uncertain_node} corresponds to considering the local move of $v_i$ which maximizes
\begin{align}
\prod_{e=(v_i,v_j), v_j\in A}
\frac{p_{\pazocal{Y}_e | \pazocal{X}_e}(1, x_e)}{p_{\pazocal{Y}_e | \pazocal{X}_e}(0, x_e)} \cdot\prod_{e=(v_i,v_j), v_j\in B}\frac{p_{\pazocal{Y}_e | \pazocal{X}_e}(0, x_e)}{p_{\pazocal{Y}_e | \pazocal{X}_e}(1, x_e)}.
\label{eq:uncertain_nodeProb}
\end{align}
To produce an uncertainty measure for each node in the graph, we apply the logistic function on \eqref{eq:uncertain_node} 
\begin{align}
\mathrm{uncertainty=\frac{1}{1+\exp{(-\gamma_i})}} 
\label{eq:uncertain_nodeest}
\end{align}
as it is the inverse of the logit function used in the cost computation in \eqref{eq:weights-map}. As shown in the supplementary material, this uncertainty measures the below ratio
%
\fontsize{8}{6}{
\begin{align}
&
\frac{\displaystyle \prod_{e,A} p_{\pazocal{Y}_e | \pazocal{X}_e}(0, x_e) . \displaystyle \prod_{e,B} p_{\pazocal{Y}_e | \pazocal{X}_e}(1, x_e)}{{\displaystyle \prod_{e,A} p_{\pazocal{Y}_e | \pazocal{X}_e}(0, x_e) . \displaystyle \prod_{e,B} p_{\pazocal{Y}_e | \pazocal{X}_e}(1, x_e)+{\displaystyle \prod_{e,B} p_{\pazocal{Y}_e | \pazocal{X}_e}(0, x_e) . \displaystyle \prod_{e,A} p_{\pazocal{Y}_e | \pazocal{X}_e}(1, x_e)}}}
\label{eq:uncertainty_eq4_1}
\end{align}
}%
where the nominator is the product of the local posterior probabilities of the observed solution $A$ at node $v_i$ (compare Eq.~\eqref{eq:map} for $p_{\pazocal{Y}_e}$ const.) and is proportional to the posterior of the chosen node label if $v_i$ has at most two labels in the local neighborhood. The denominator sums trivially to one in the case of $|\pazocal{N}_{E'}(v_i)| =1$. For the more common case of  $|\pazocal{N}_{E'}(v_i)| \geq 1$, expression~\eqref{eq:uncertainty_eq4_1} can be interpreted as a "calibrated" probability. It normalizes the posterior probability of the MAP solution with the sum of posteriors of this solution and the second most likely one. Intuitively, if all solutions have a low absolute posterior probability, the relatively best one can still have a high certainty.

\section{Evaluation and Results}
\label{sec:evaluation}

We evaluate the proposed uncertainty measure in two common application scenarios of minimum cost multicuts: motion segmentation and image segmentation. For motion segmentation, we conduct experiments on the datasets Freiburg-Berkeley Motion Segmentation FBMS$_{59}$~(\cite{Ochs14}) and Densely Annotated Video Segmentation (DAVIS$_{2016}$)~(\cite{davis_16}). On both datasets, we compute point trajectories as well as the segmentation using the method from \cite{Keuper_15}, which computes edges between point trajectories as pseudo-probabilities from a simple motion model. 

Additionally, we highlight the potential benefit of a robust uncertainty measure: It not only allows us to produce several hypotheses of solutions. We can also employ the estimated uncertainties in the densification framework from \cite{selfsupervised} to compute improved, dense video segmentations from sparse ones. 

Last, we evaluate the proposed uncertainty measure on the image segmentation task posed by the BSDS-500 dataset by ~\cite{BSDS500}, employing the minimum cost lifted multicut instances from 
\cite{Keuper2015}.
\begin{figure}[t]
    \centering 
        \begin{tabular}{@{}l@{}l@{}l@{}l@{}l@{}}
            \includegraphics[height=1.45cm]{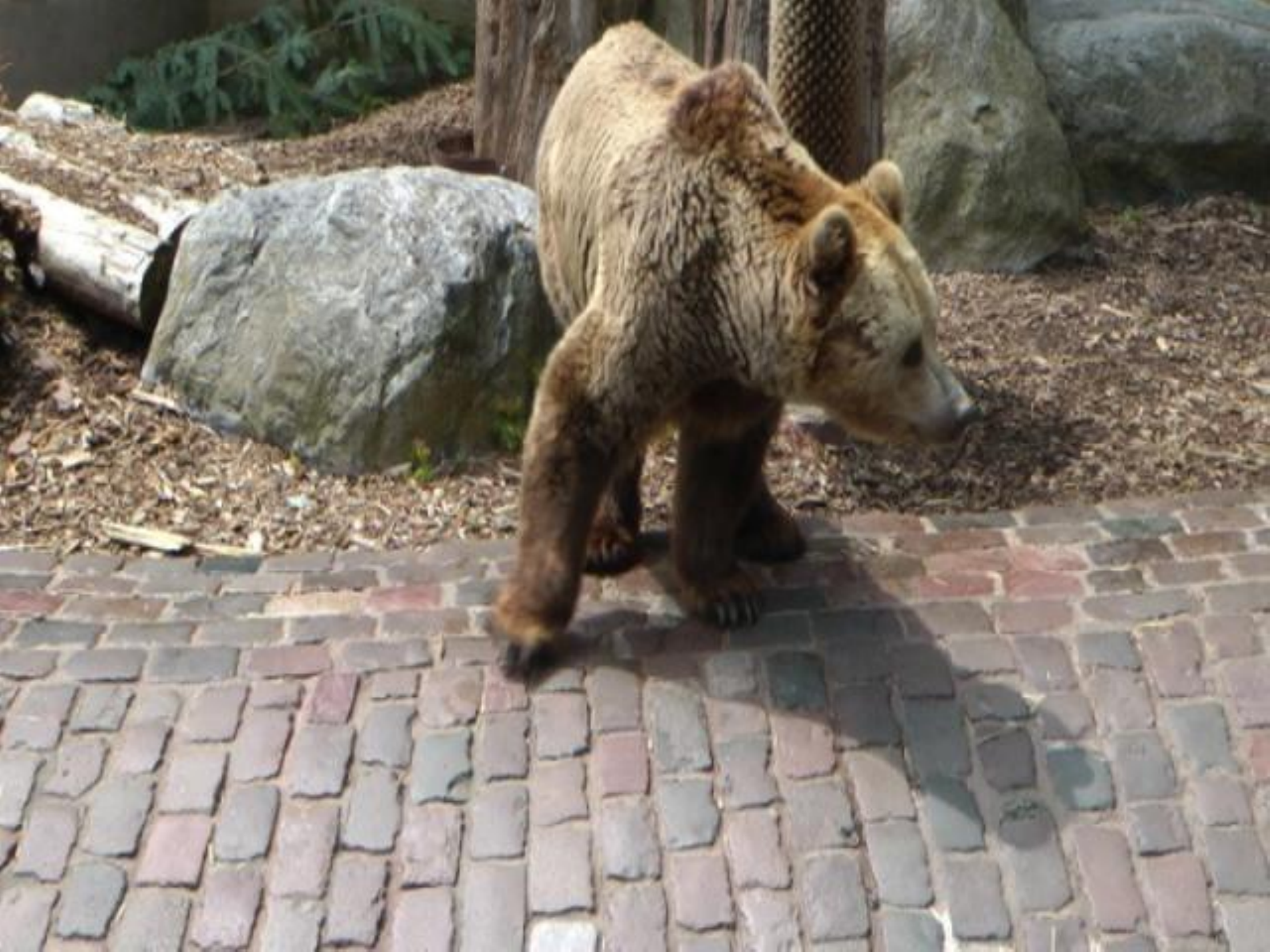}
            \includegraphics[height=1.45cm]{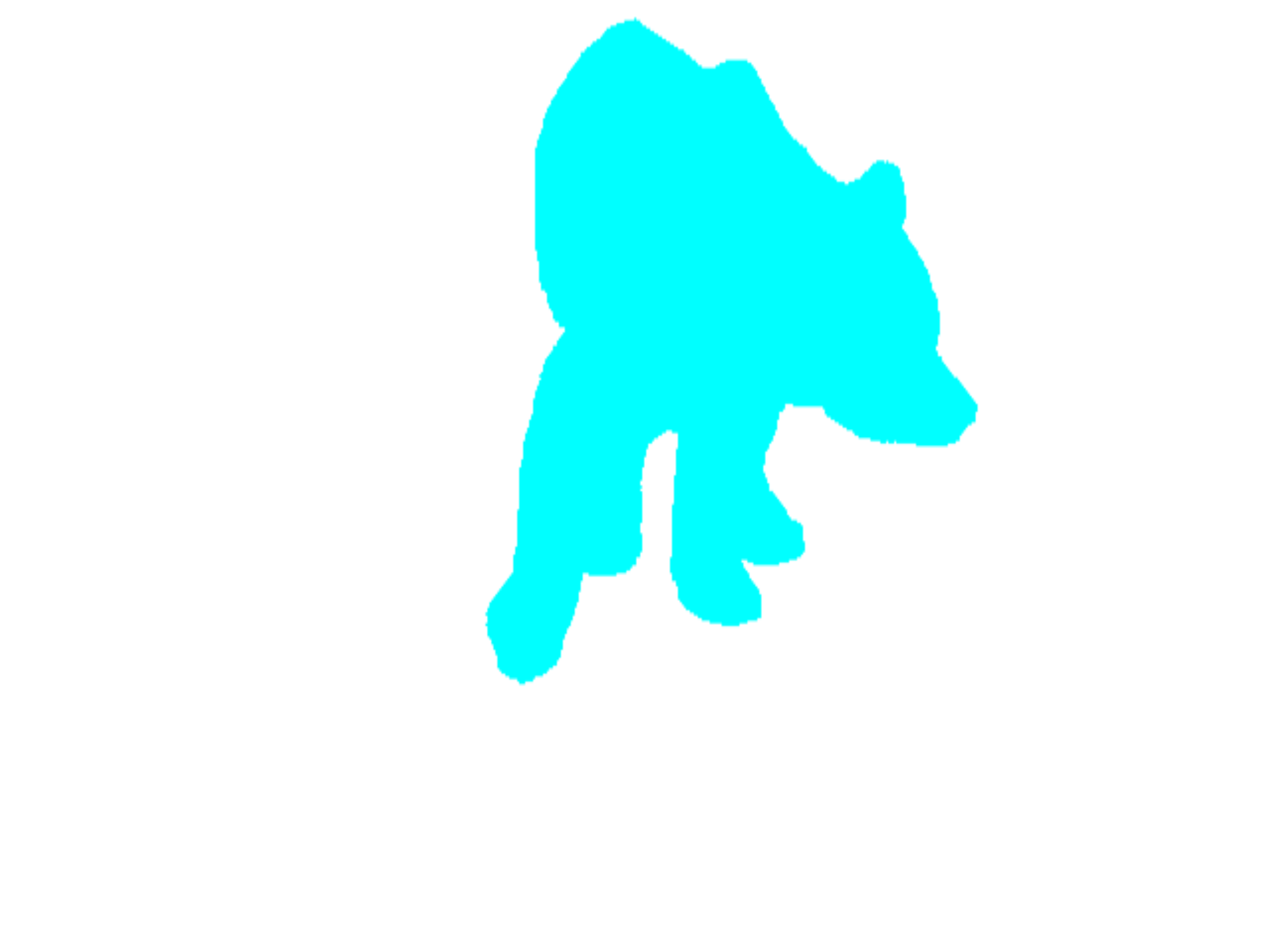}&
            \includegraphics[height=1.45cm]{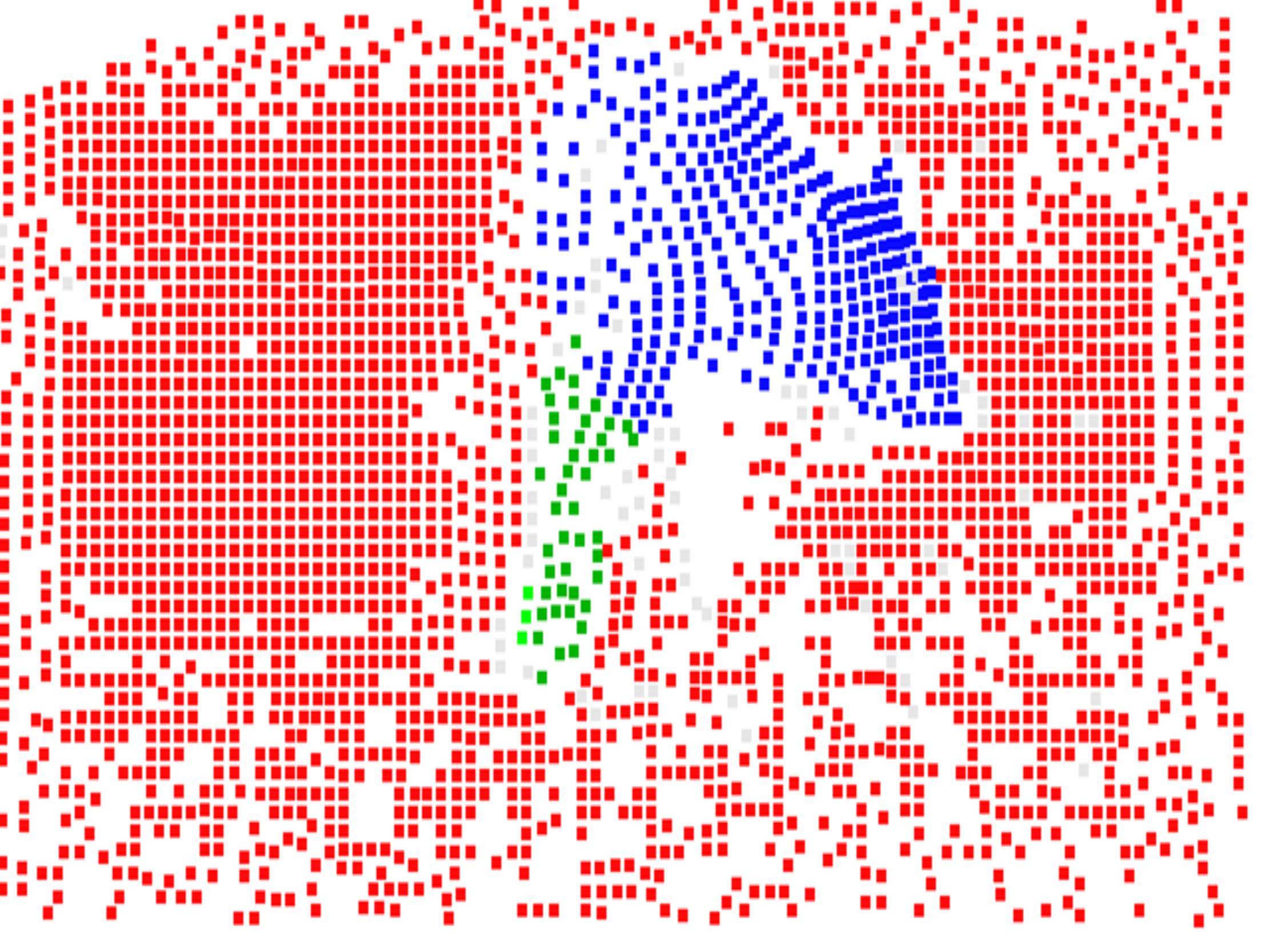}&
            \includegraphics[height=1.45cm]{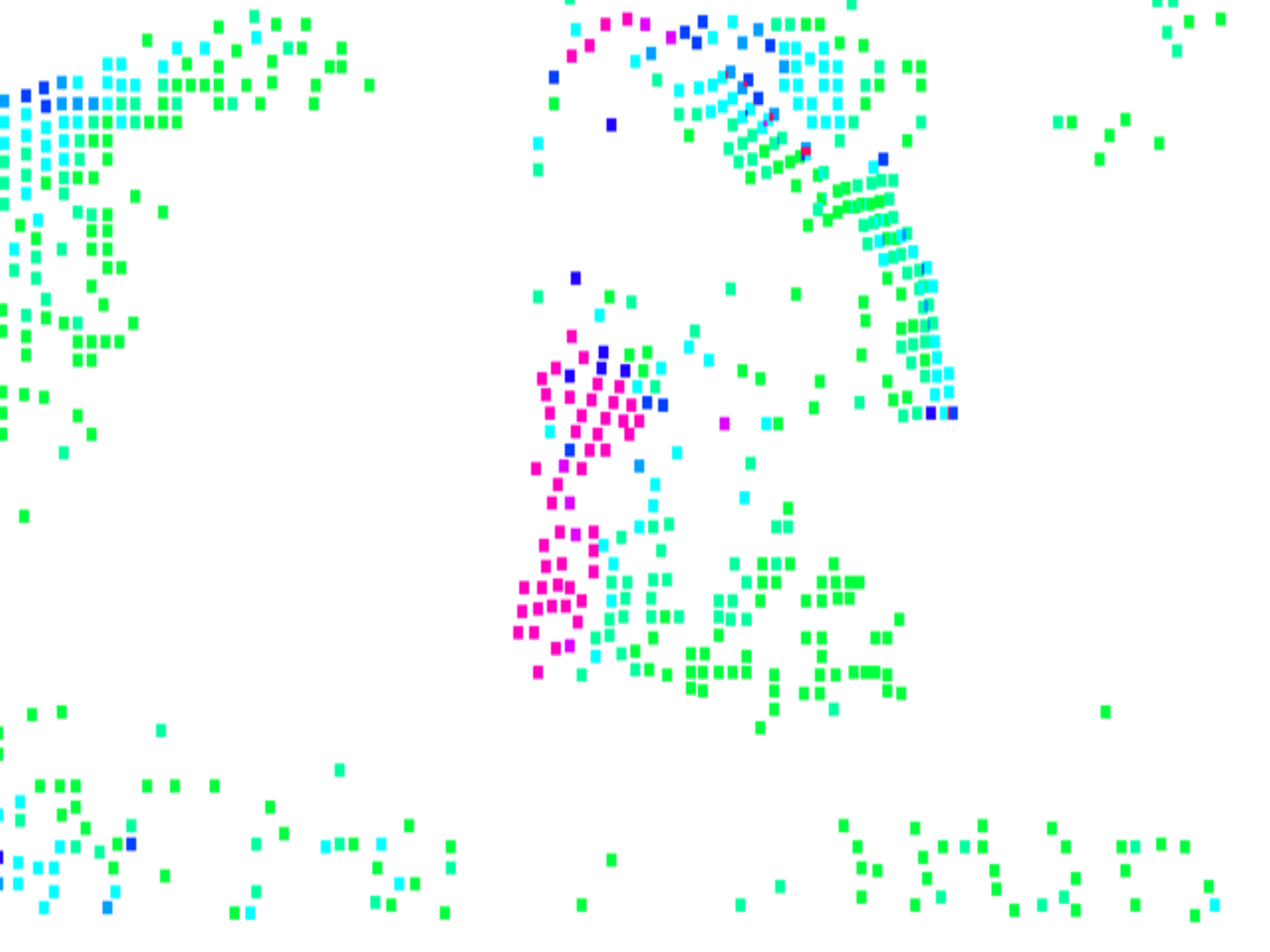}&
            \includegraphics[height=1.45cm]{1_resized_colormap.ppm.pdf}\\

            \includegraphics[height=1.45cm]{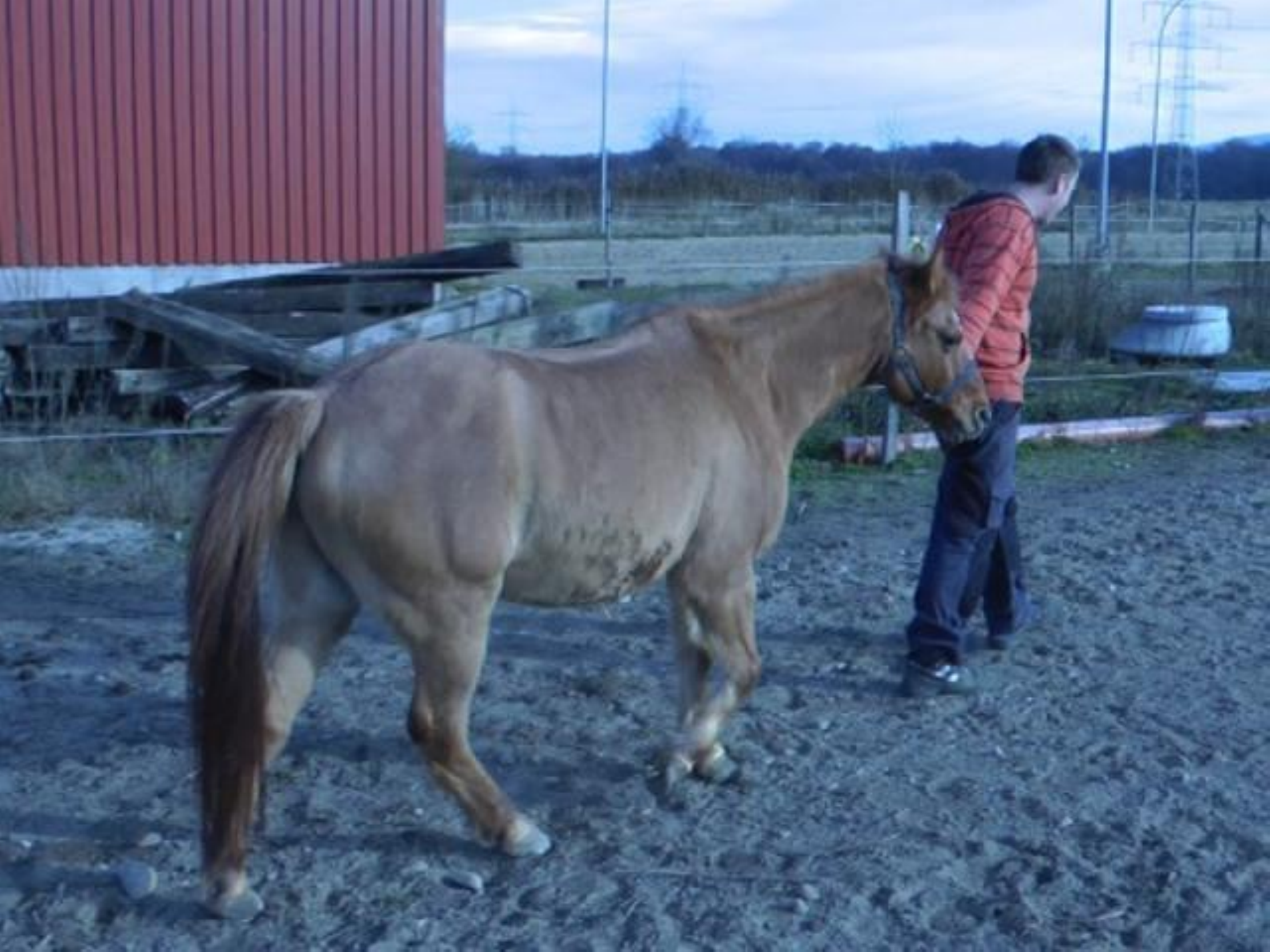}
            \includegraphics[height=1.45cm]{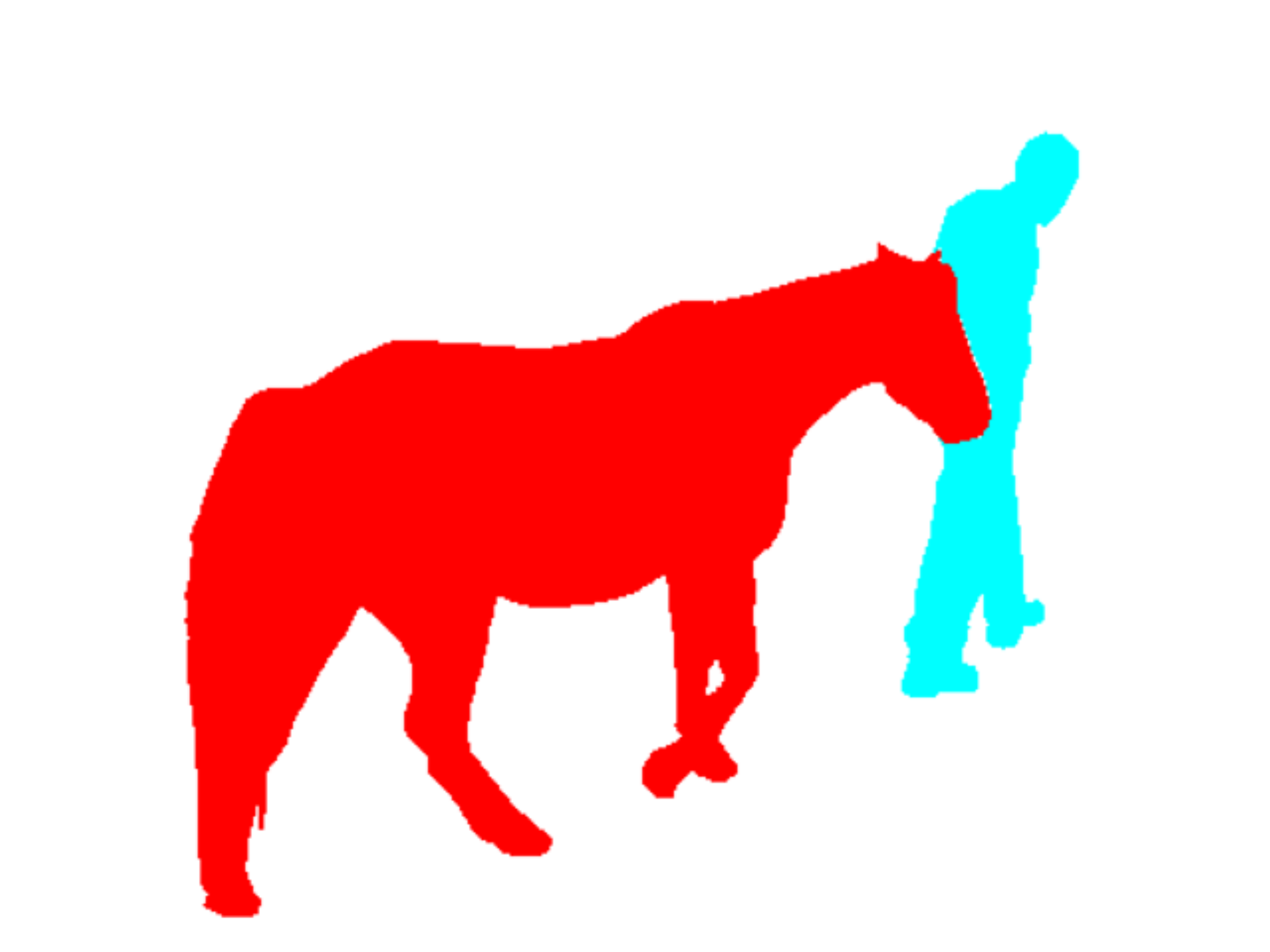}&
            \includegraphics[height=1.45cm]{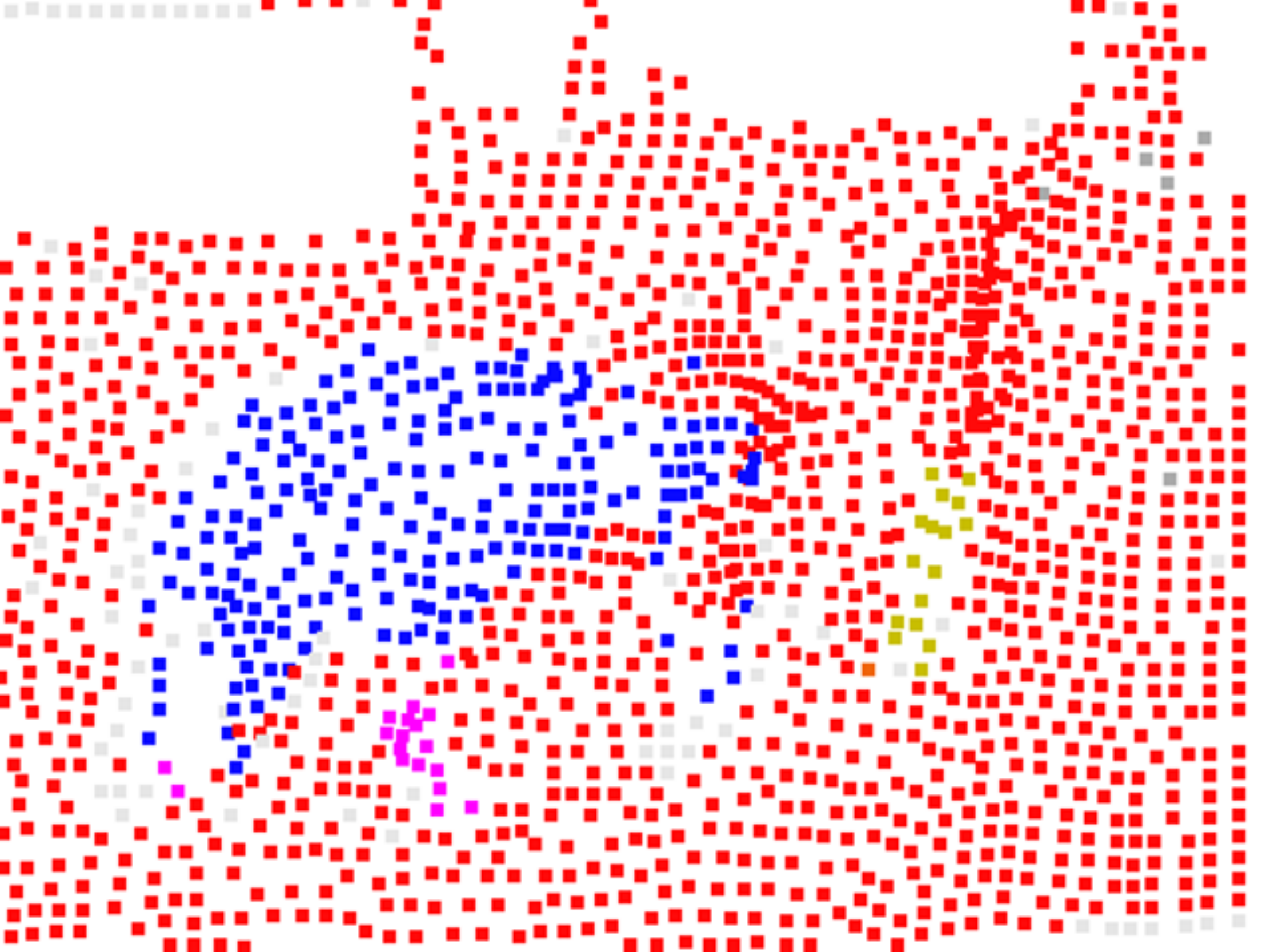}&
            \includegraphics[height=1.45cm]{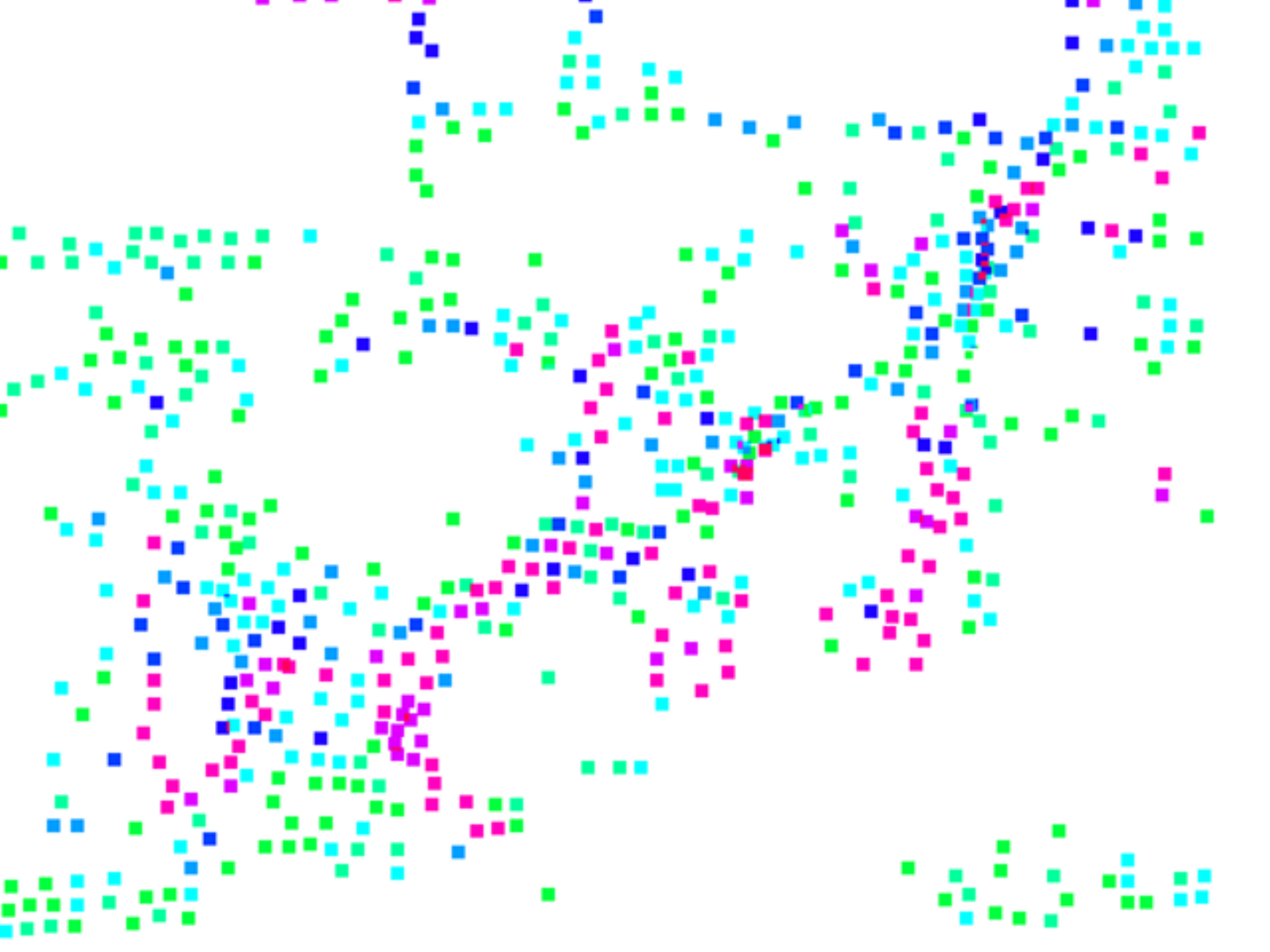}\\
            \includegraphics[height=1.45cm]{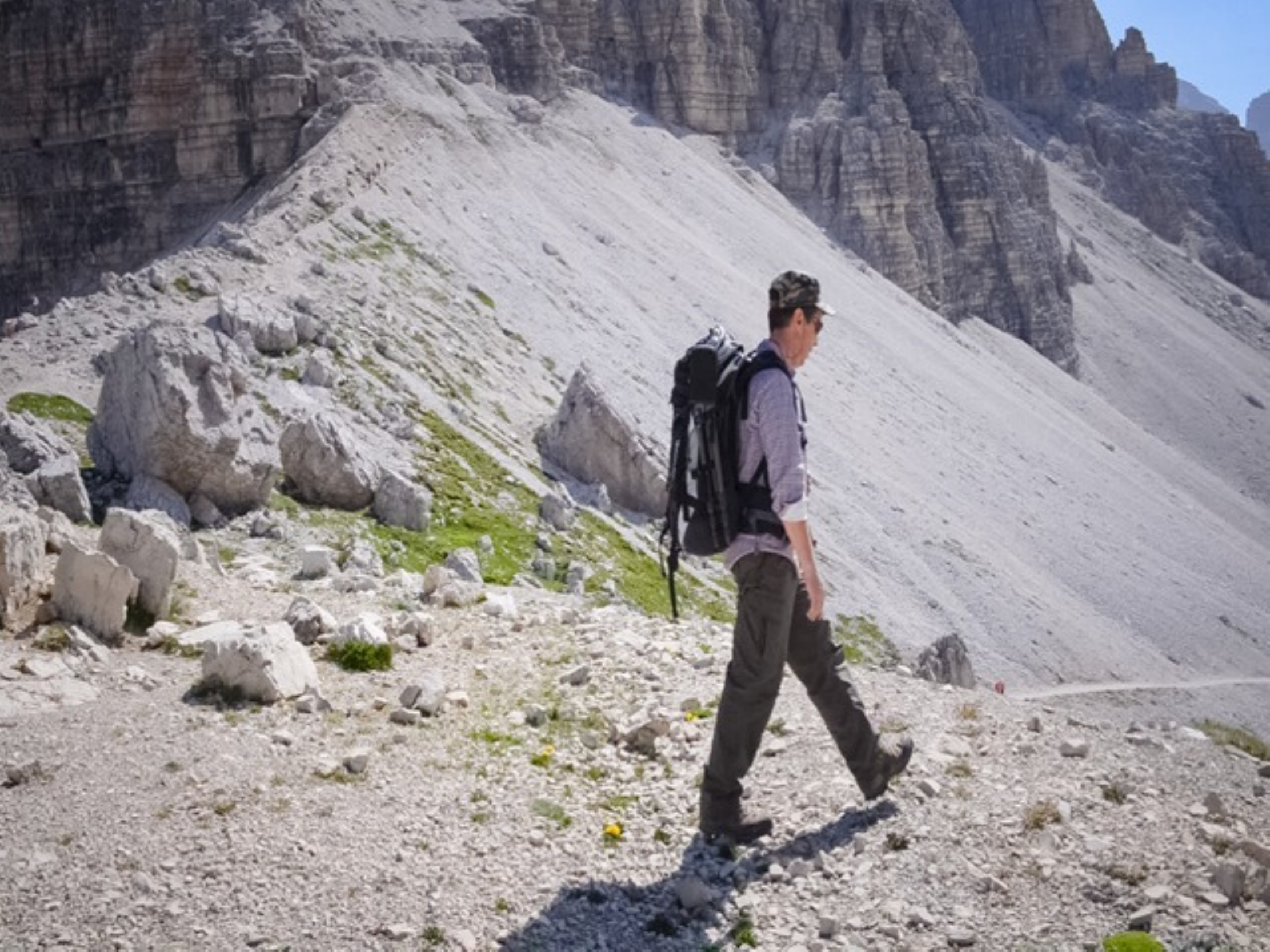}
            \includegraphics[height=1.45cm]{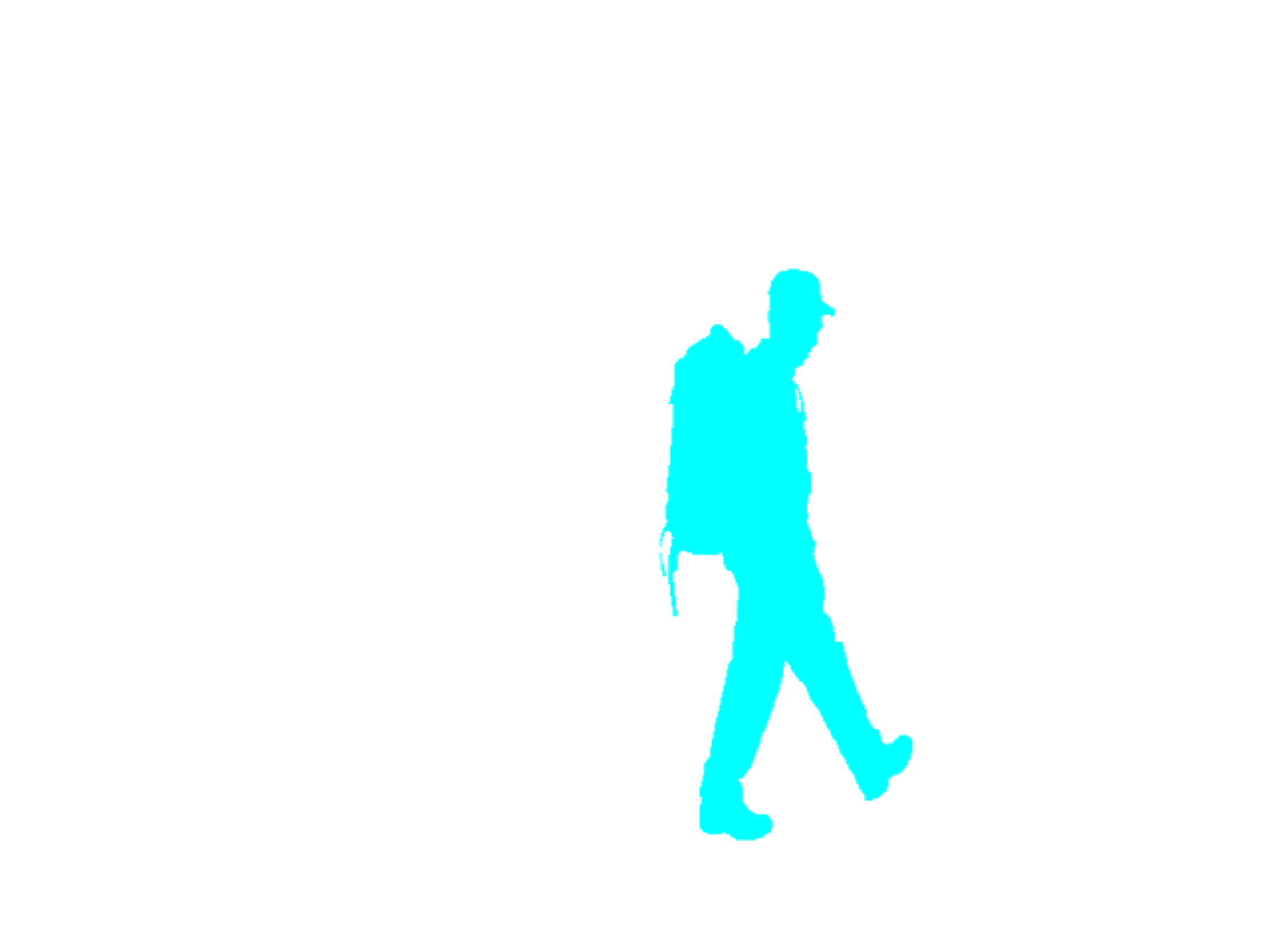}&
            \includegraphics[height=1.45cm]{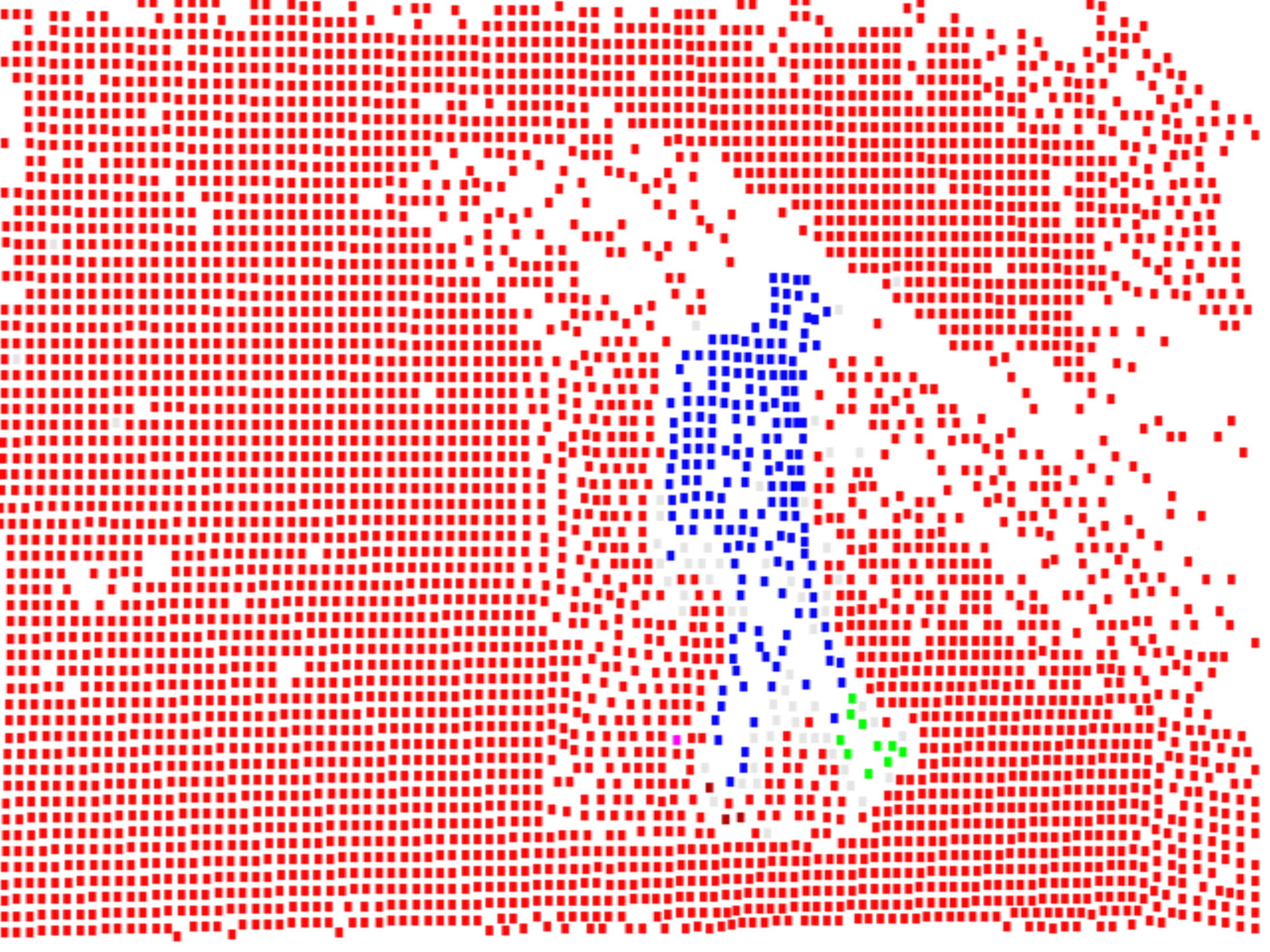}&
            \includegraphics[height=1.45cm]{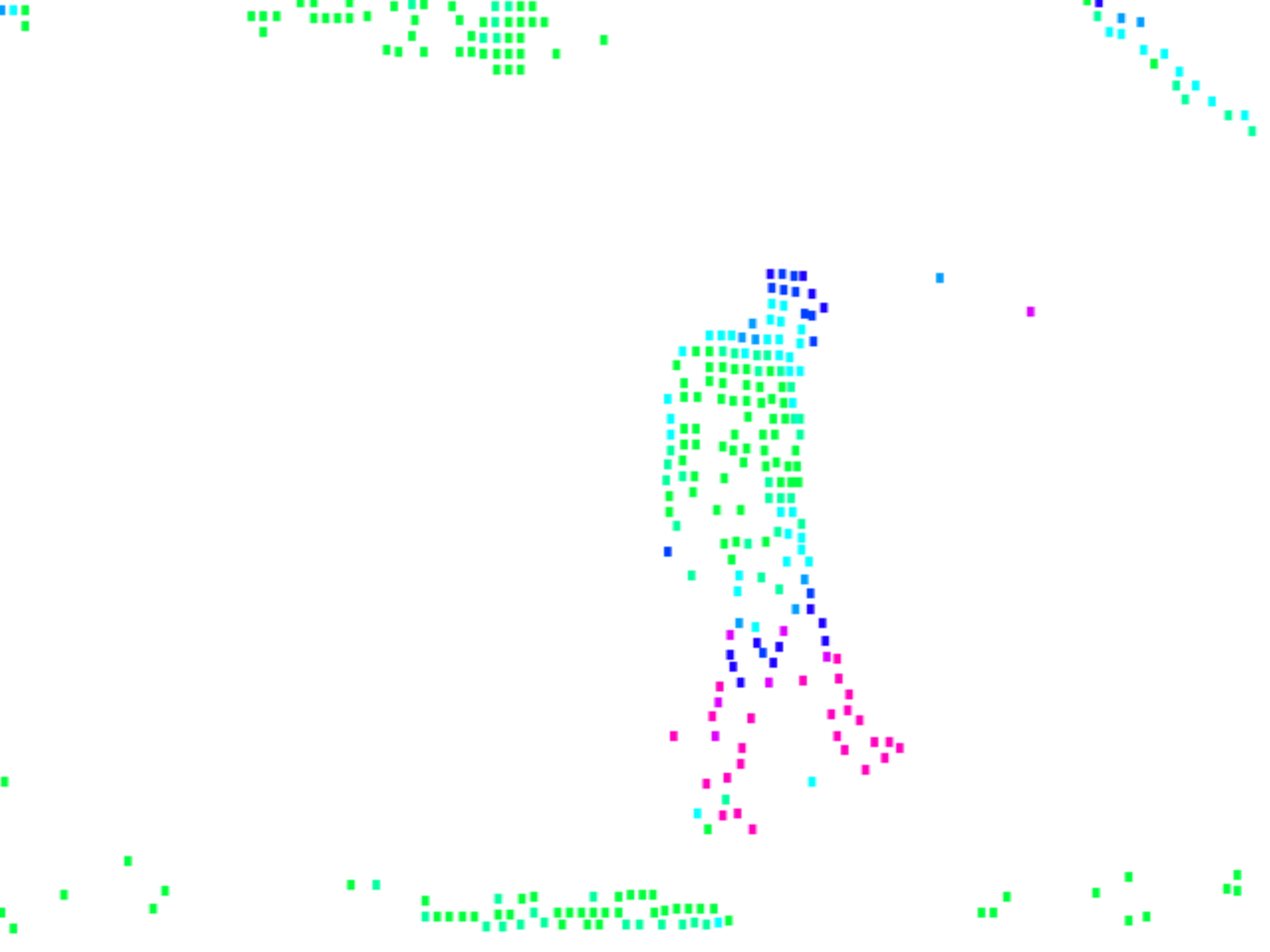}\\
            \includegraphics[height=1.45cm]{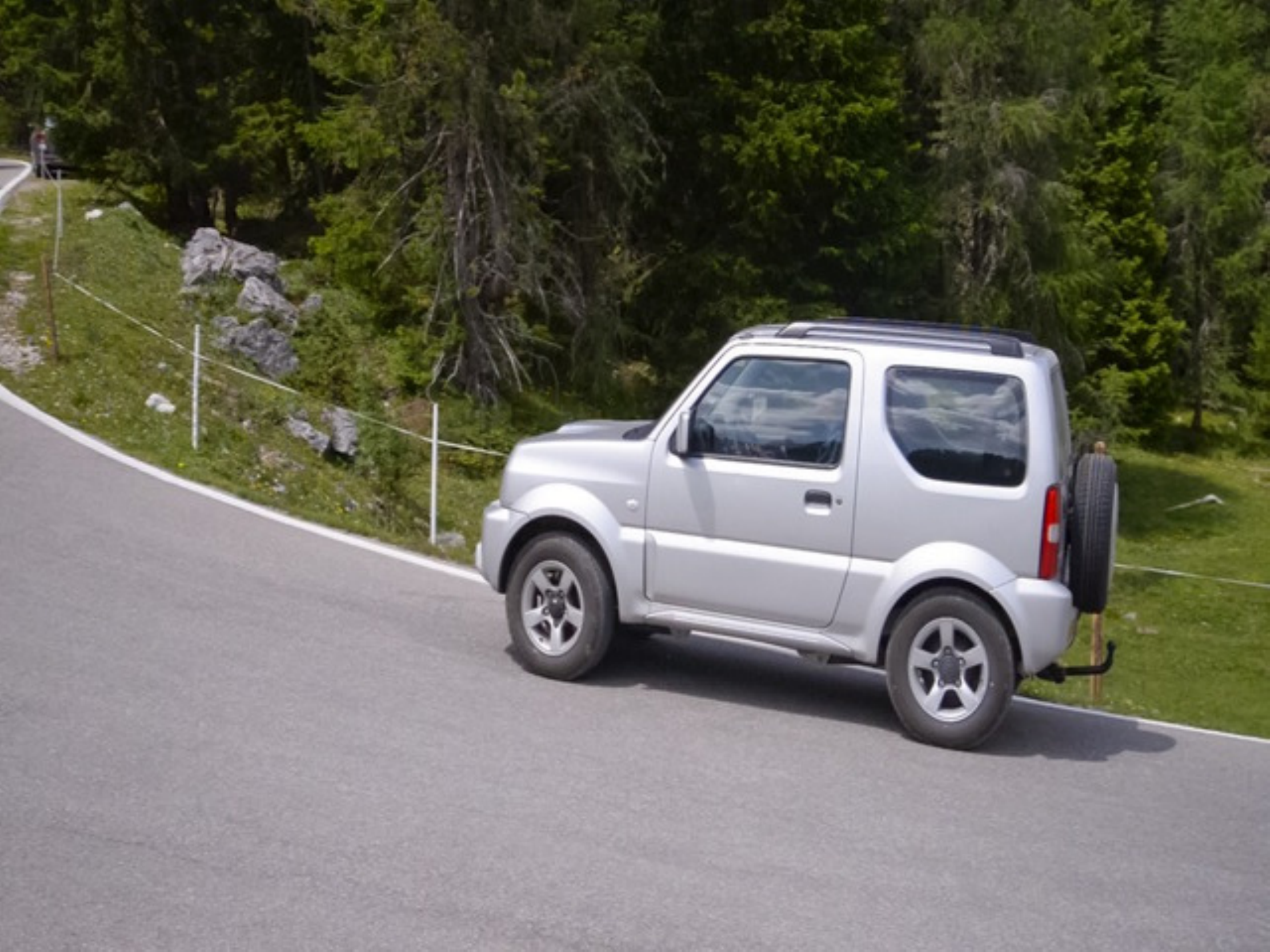}
            \includegraphics[height=1.45cm]{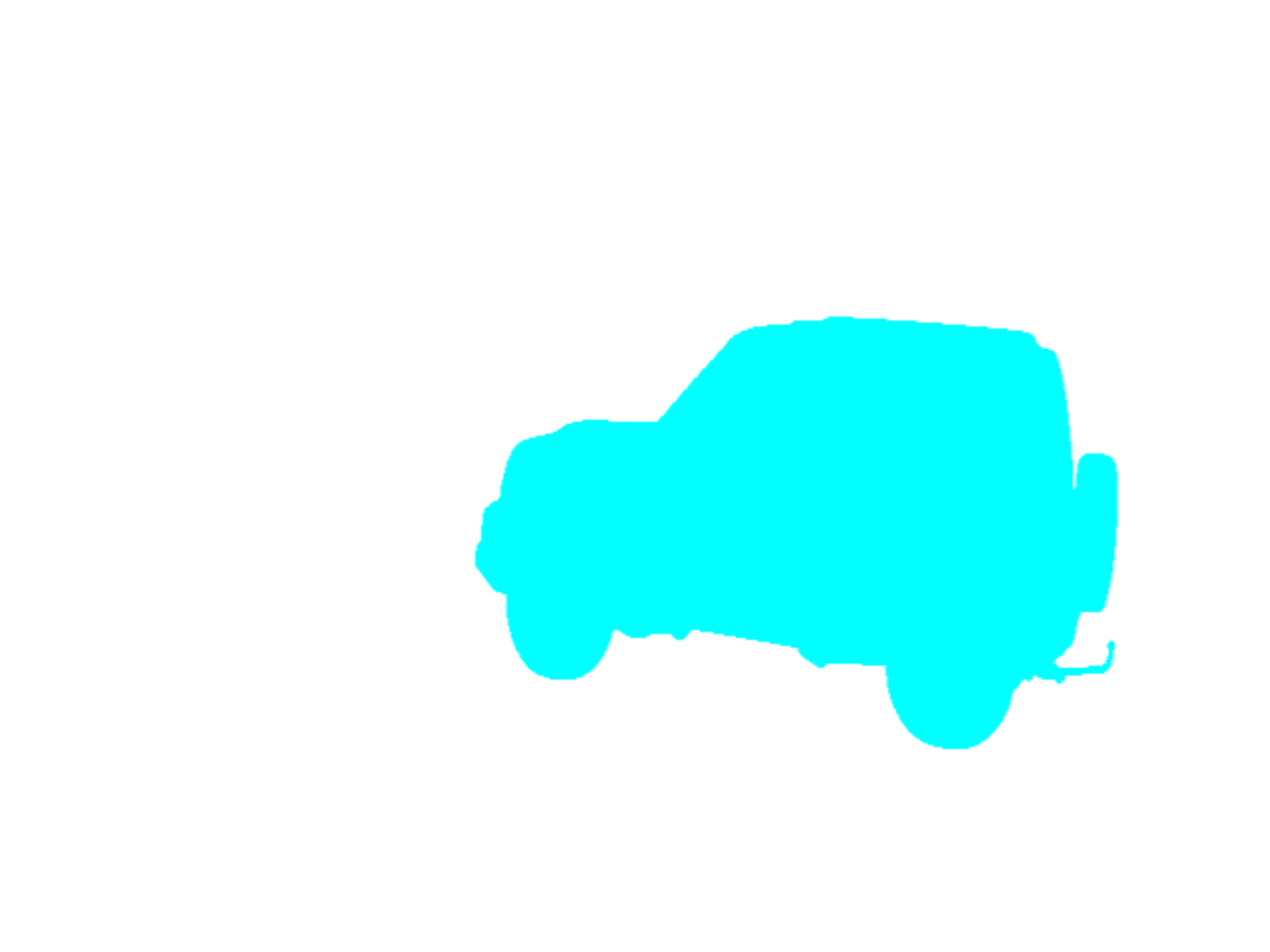}&
            \includegraphics[height=1.45cm]{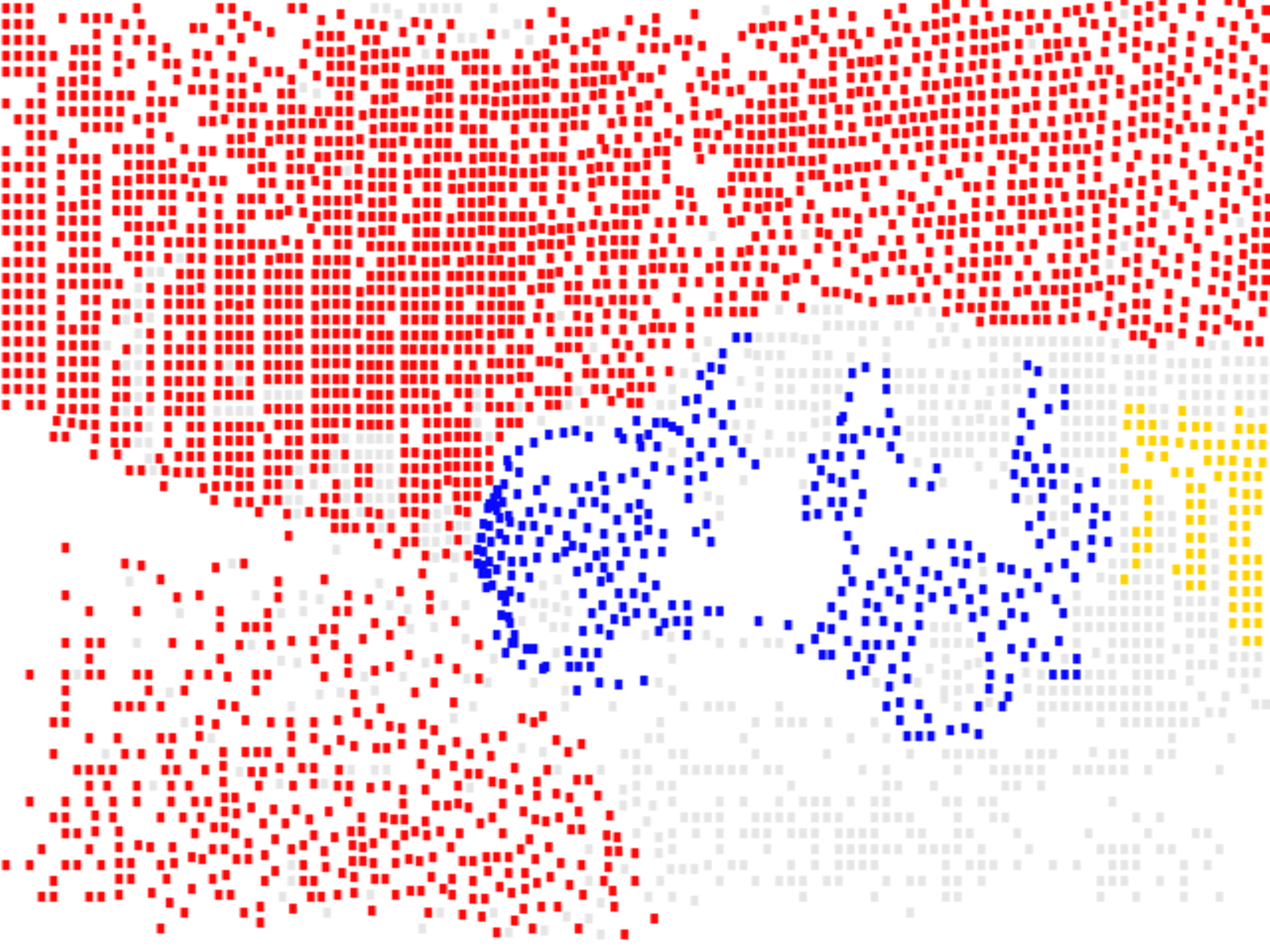}&
            \includegraphics[height=1.45cm]{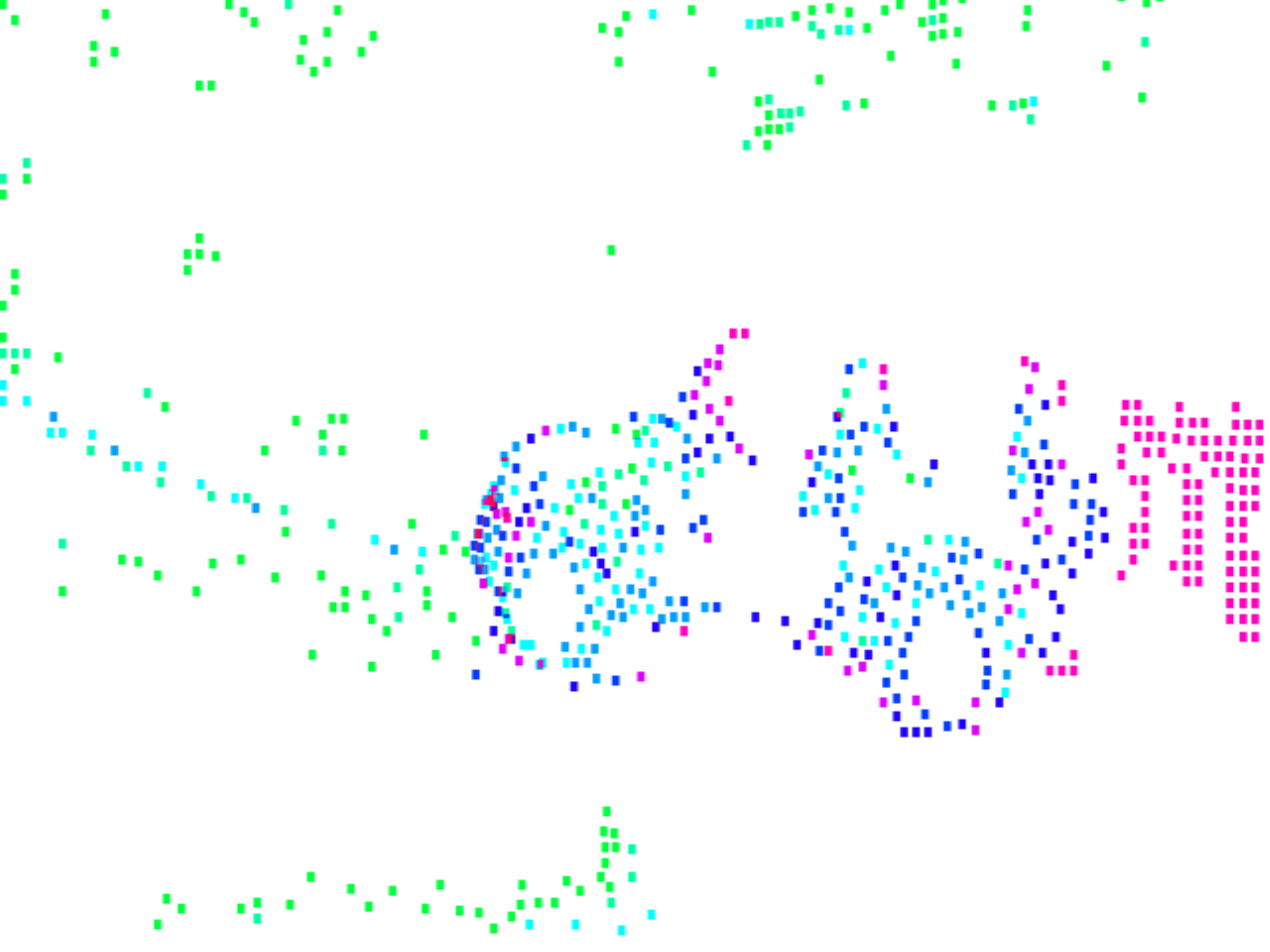}\\
                 \end{tabular}
    \caption{Uncertainty measure on point trajectories on two sequences from FBMS$_{59}$ (first two rows) and DAVIS$_{2016}$ (last two rows). In each row, from left to right, we provide an image, its ground-truth, the segmentation and our uncertainty estimation. The uncertainty values are discretized and color coded for visualization purposes. White areas 
    correspond to the trajectories with high certainty. The uncertainty on thin, articulated object parts is high. 
    }
    \label{fig:davis_fbms_imgs_nodes}
\end{figure}
\paragraph{Freiburg-Berkeley Motion Segmentation} The FBMS$_{59}$~(\cite{Ochs14}) dataset contains 29 train and 30 test sequences. They are between 19 to 800 frames long and show motion of possibly multiple foreground objects. The sequences contain camera shaking, rigid/non-rigid object motion as well as occlusion/dis-occlusion events. The ground-truth segmentation is provided for a subset of the frames in each sequence. 

\paragraph{Densely Annotated Video Segmentation Dataset} The DAVIS$_{2016}$ dataset by~\cite{davis_16} has been introduced for binary video object segmentation. It contains 30 train and 20 validation sequences where objects undergo occlusion and dis-occlusion and rigid as well as non-rigid movements. The pixel-wise binary ground truth segmentation is provided per frame for each sequence. The DAVIS$_{2016}$ dataset has been proposed for \textit{video instance segmentation}. Yet, as each sequence only contains one annotated object with eminent motion with respect to the scene, the dataset can be used to evaluate motion segmentation tasks. Our method is evaluated on train and test sequences of FBMS$_{59}$ and train and validation sequences of DAVIS$_{2016}$.

\paragraph{Berkeley Segmentation Dataset} The BSDS-500~dataset~(\cite{BSDS500}) consists of 200 train, 200 test and 100 validation images, where for each images five different human made annotations are provided. We evaluate our method on the test images.

\paragraph{Parameter Setting}
In the MP and LMP, the value $\beta$ (as in the Eq.~\eqref{eq:weights-map}) is set to $0.5$ (unbiased) to be comparable with the baseline methods in the evaluated datasets. The LMP on images requires the value $\tau$ to be set, which corresponds to the radius until which to insert lifted edges. We set $\tau=20$ as in \cite{Keuper2015}. 
\paragraph{Metrics} In both applications, we evaluate our uncertainty measure based on the variation of information (VI)~(\cite{metrics}) and Rand index (RI)~(\cite{metrics}). We study the effect of measured uncertainties by means of sparsification. The most uncertain nodes are removed subsequently and in each step VI and RI are measured. Ideally, VI should drop and RI should increase monotonically as more uncertain nodes are removed. In~\cite{metrics} it is shown that VI is a more reliable indicator than the RI with respect to the density of the results, since the RI depends heavily on the granularity of the decomposition. 
Specifically, the more pixels or motion trajectories are removed during sparsification, the less reliable the RI becomes, which is why both metrics are reported in all our experiments.
It is important to notice that high uncertainty of a node indicates that the label assigned to the node after the termination of the solver has a tendency to flip. Therefore, removing those nodes from the decomposition should improve the quality of the decomposition in terms of VI and RI. We compared our approach with the proposed baseline approaches.

\begin{figure}[t]
    \begin{center}
    \small
    \begin{tabular}{@{}c@{}c@{}}
    \includegraphics[width=0.51\linewidth]{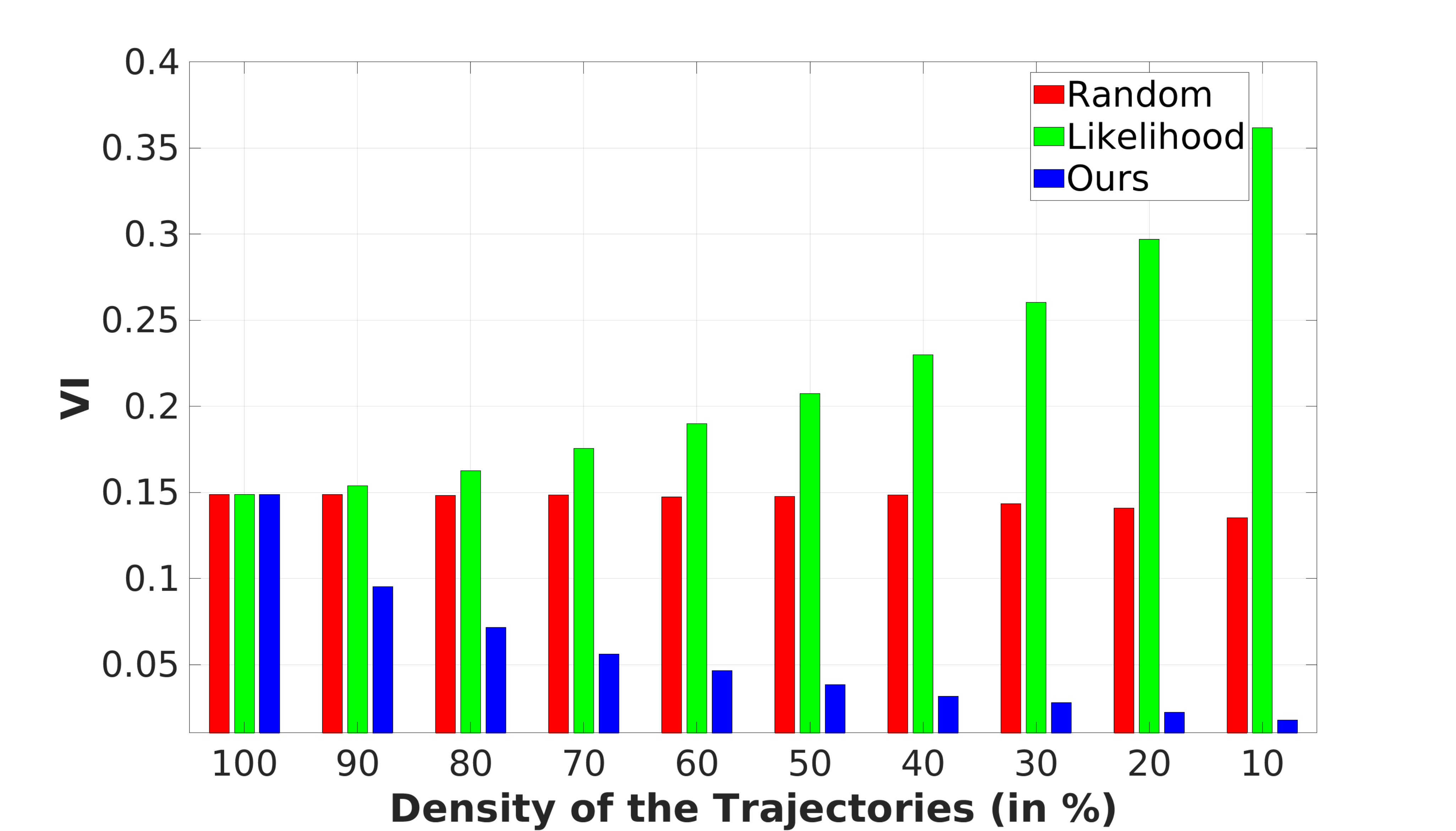}&
    \includegraphics[width=0.51\linewidth]{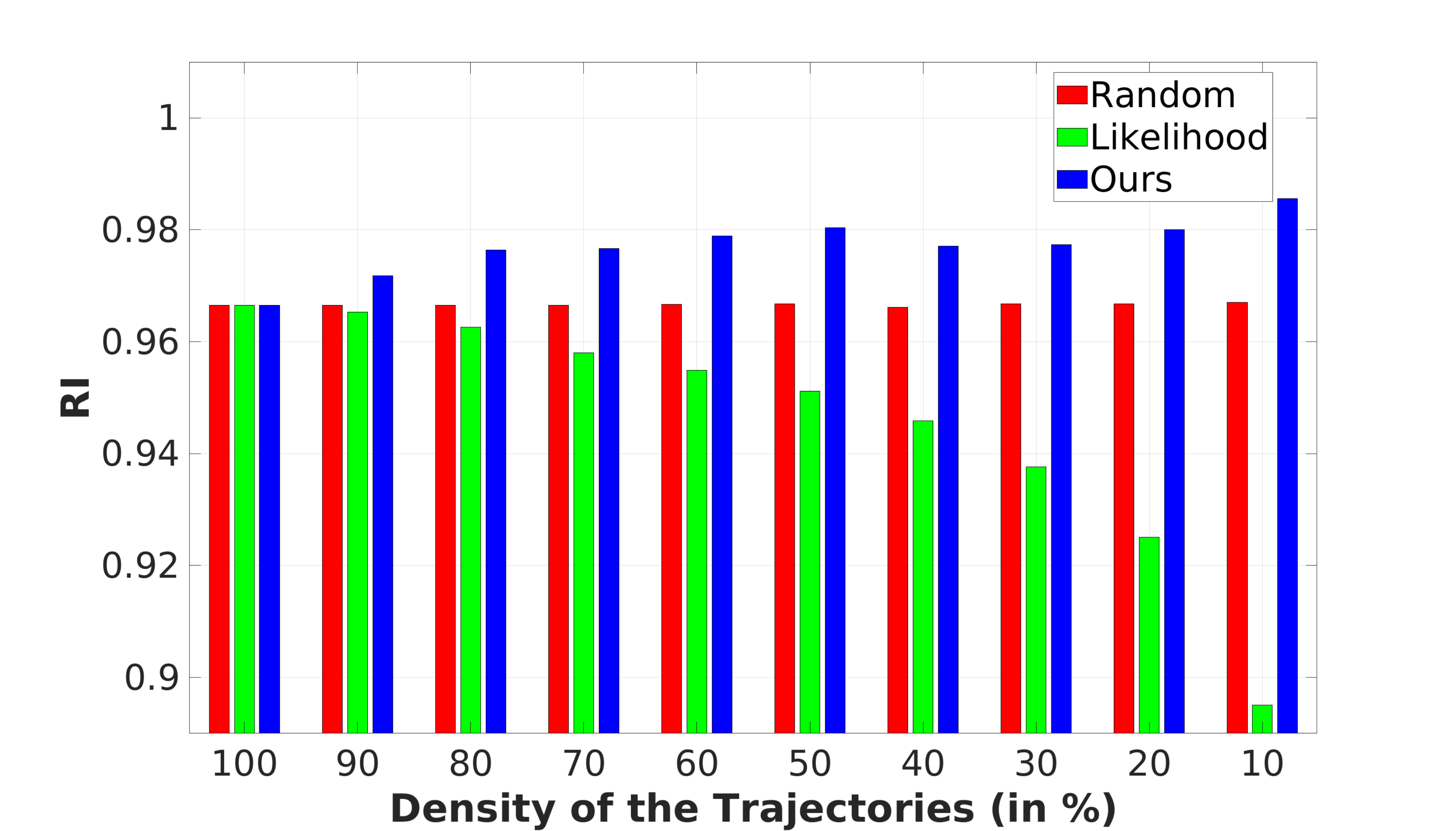}\\
     \includegraphics[width=0.51\linewidth]{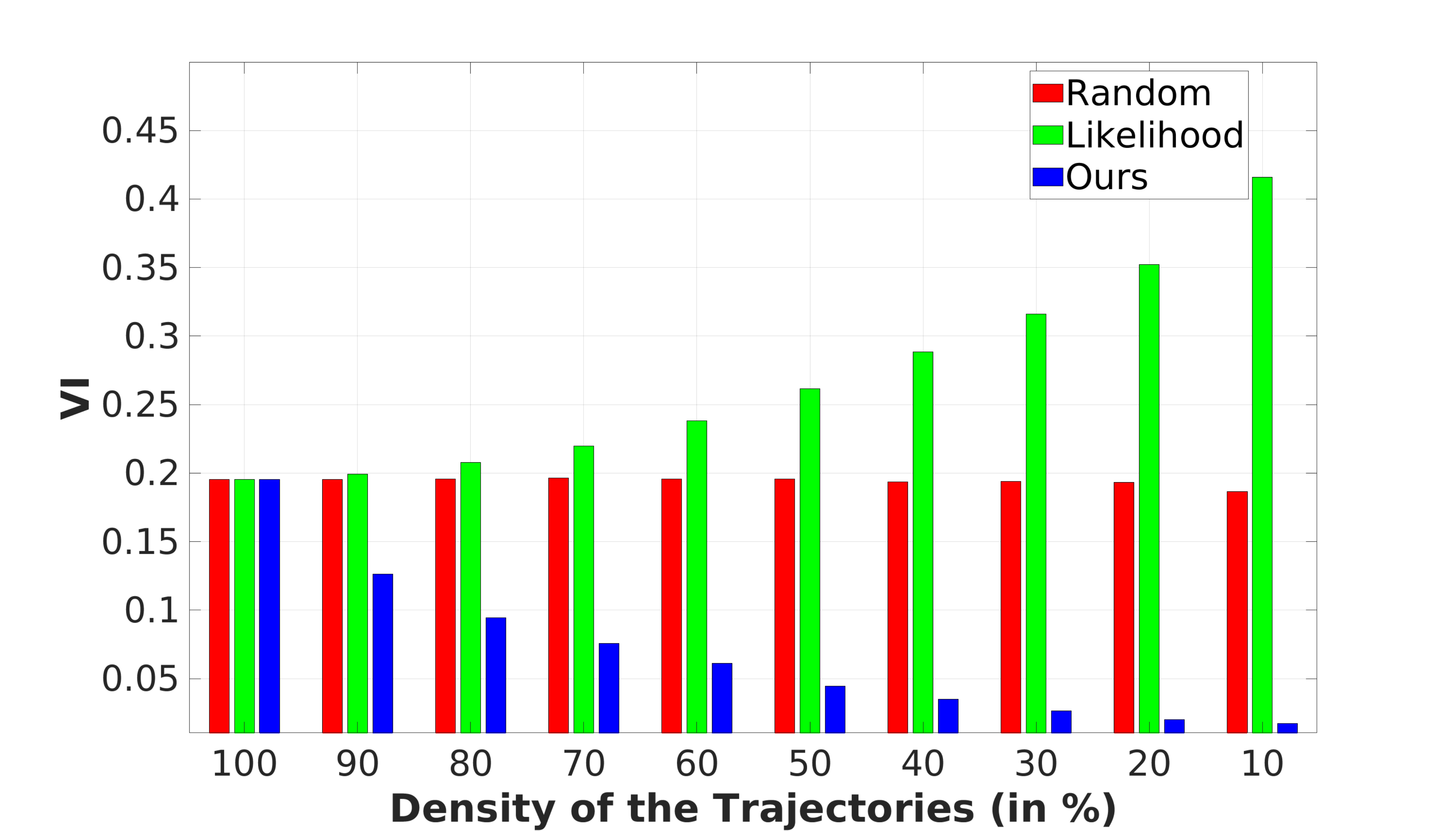}&
    \includegraphics[width=0.51\linewidth]{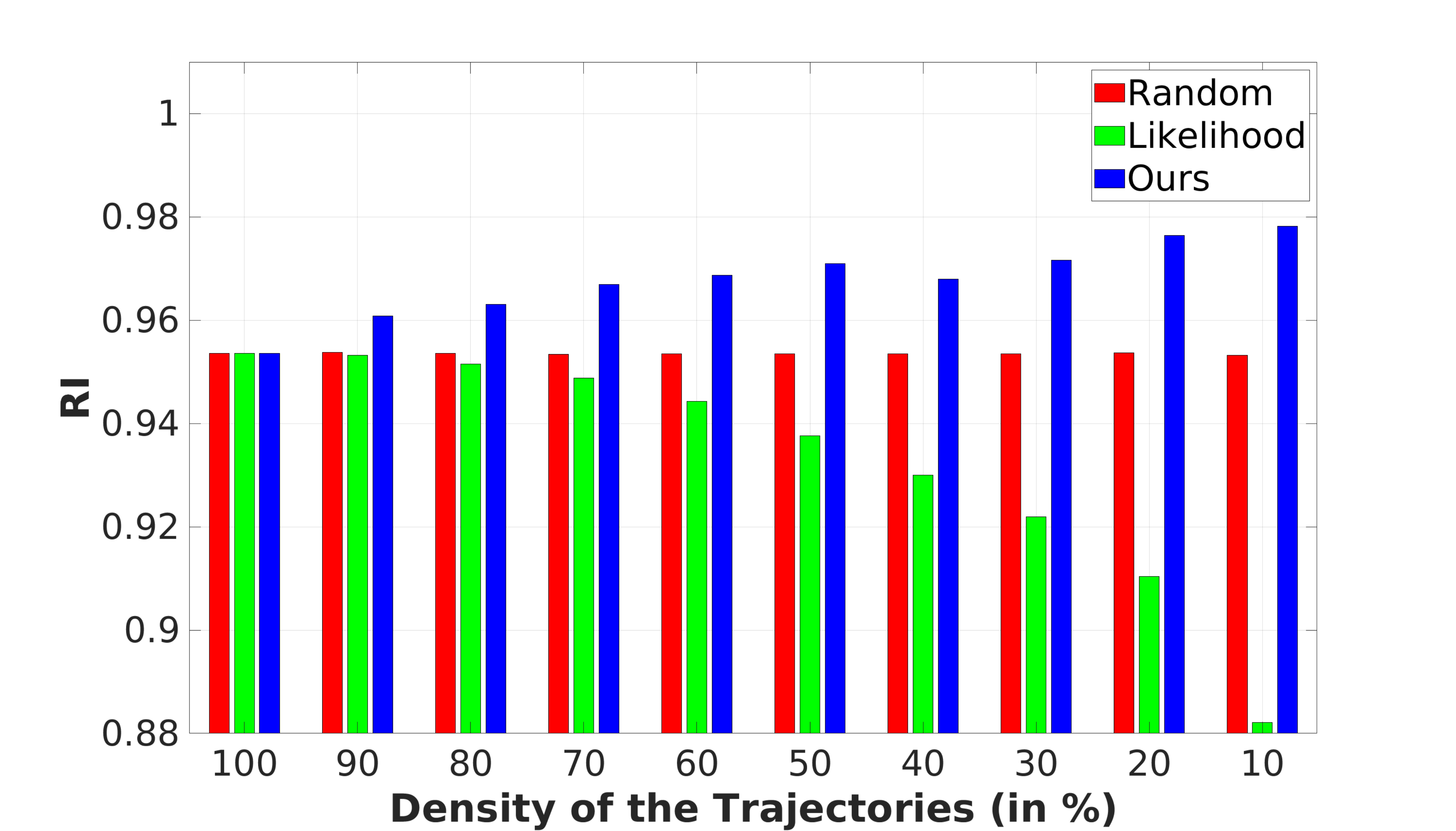}\\
       VI (lower is better)&RI (higher is better)
    \end{tabular}
    \end{center}
    \caption{Study on motion trajectory uncertainties on VI (\textit{left}) and RI (\textit{right}) on the train set (top row) and on the test set (bottom row) of FBMS$_{59}$. The metrics improve by removing trajectories according to the proposed uncertainty measure. Notice that removing uncertain trajectories according to the likelihood baseline deteriorates both VI and RI.} 
    \label{fig:fbms_train_ri_vi_baselines}
\end{figure}
\begin{figure}[t]
    \begin{center}
    \small
    \begin{tabular}{@{}c@{}c@{}}
    \includegraphics[width=0.51\linewidth]{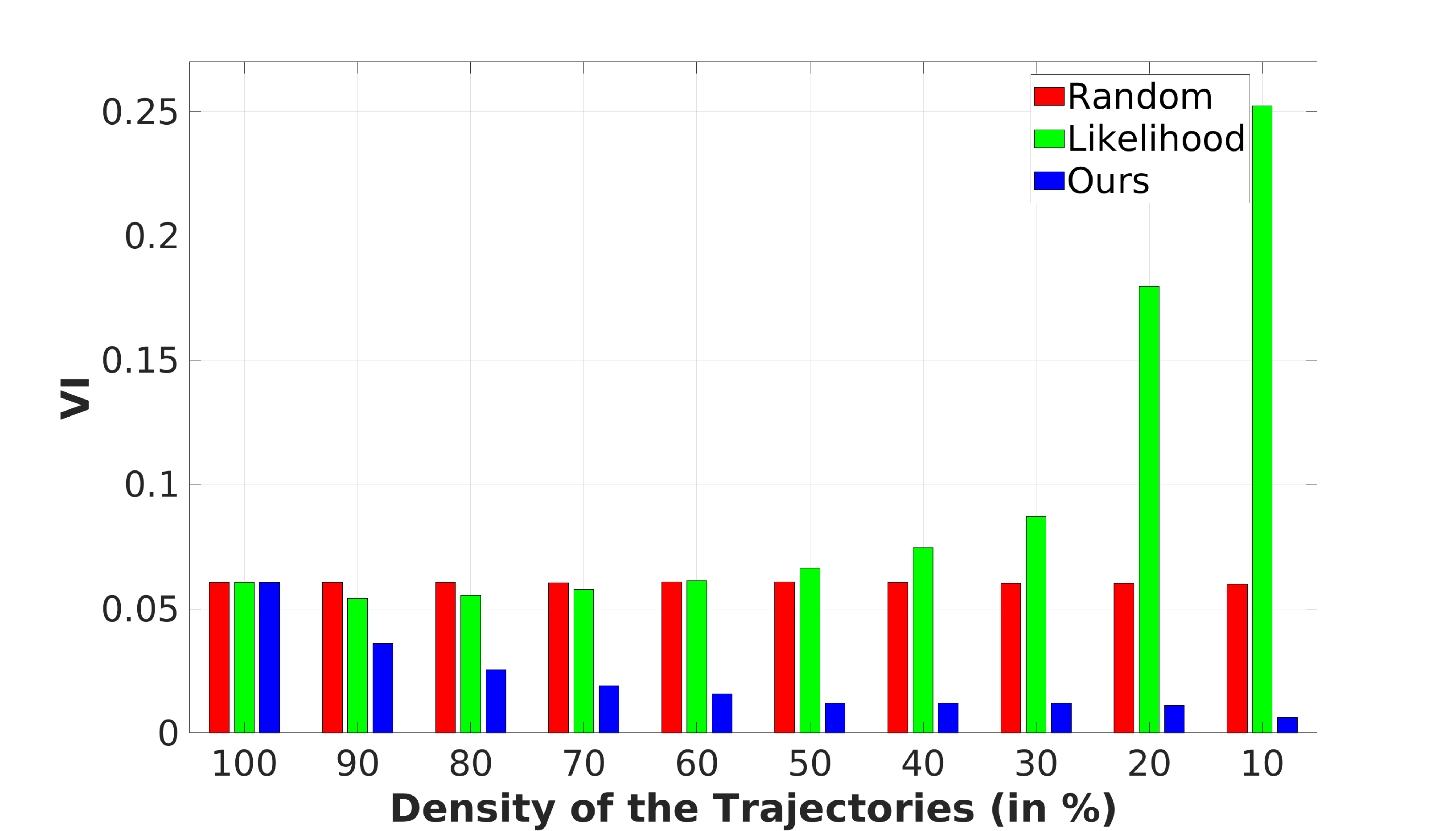}&
    \includegraphics[width=0.51\linewidth]{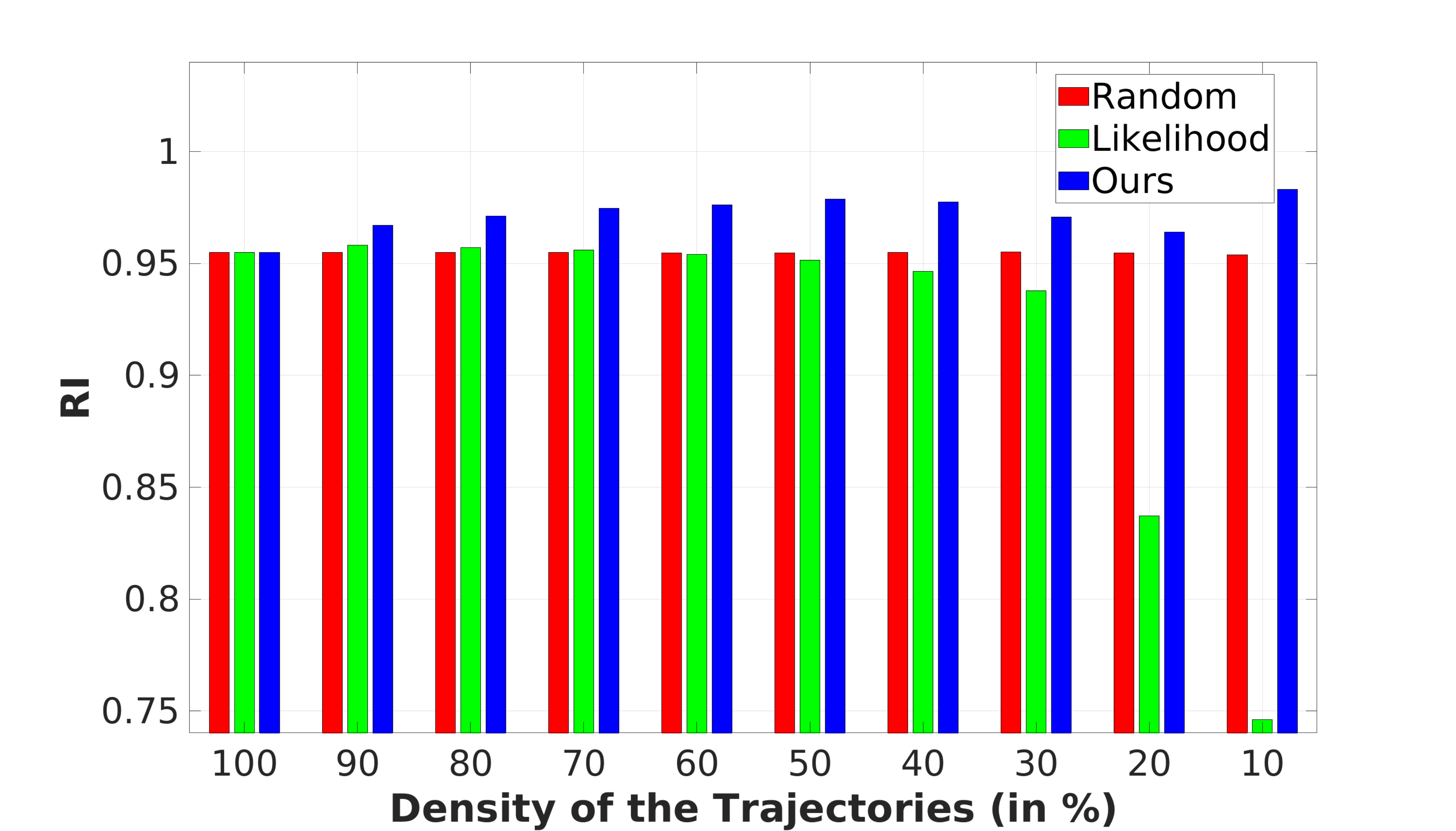}\\
        \includegraphics[width=0.51\linewidth]{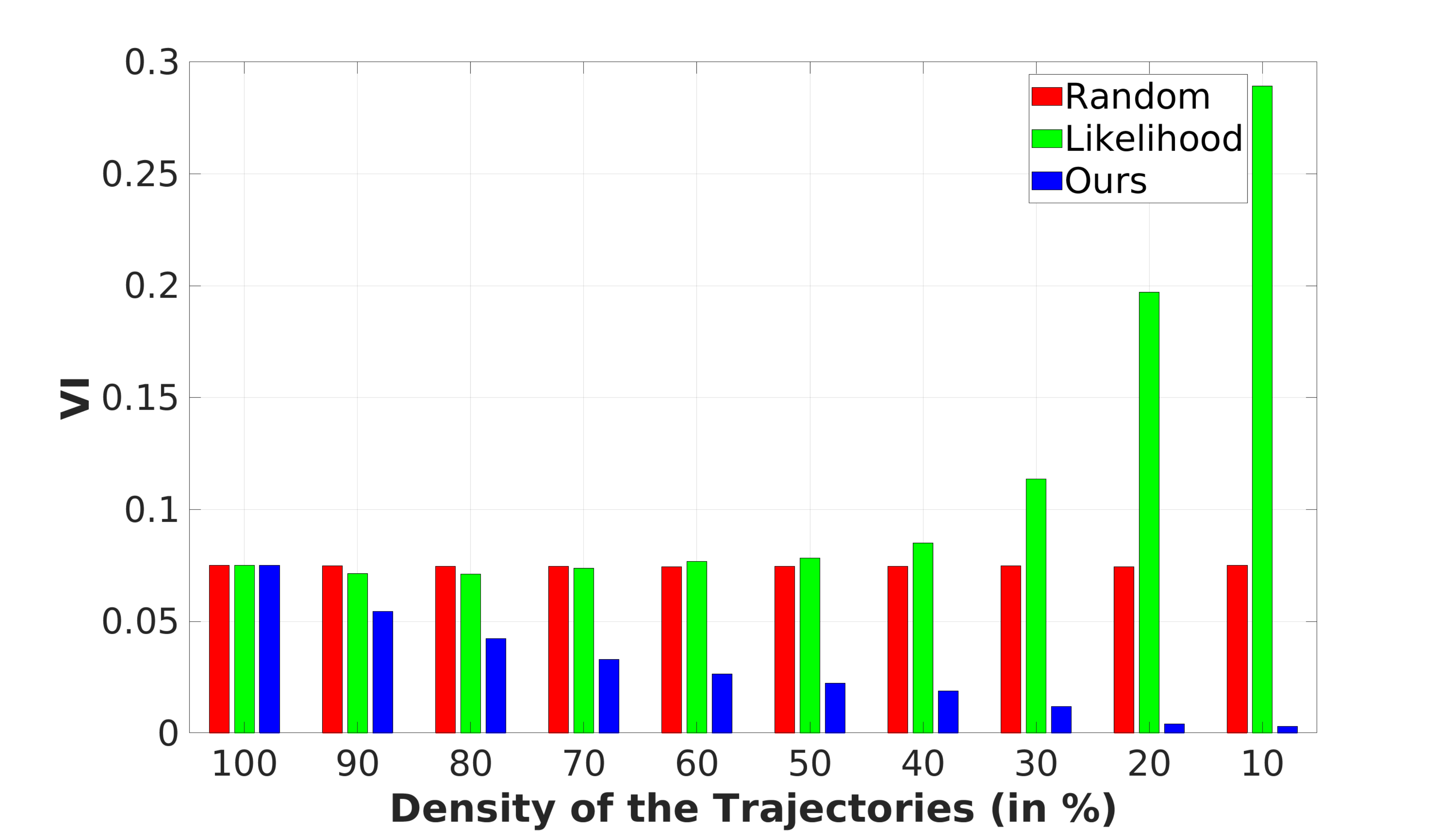}&
    \includegraphics[width=0.51\linewidth]{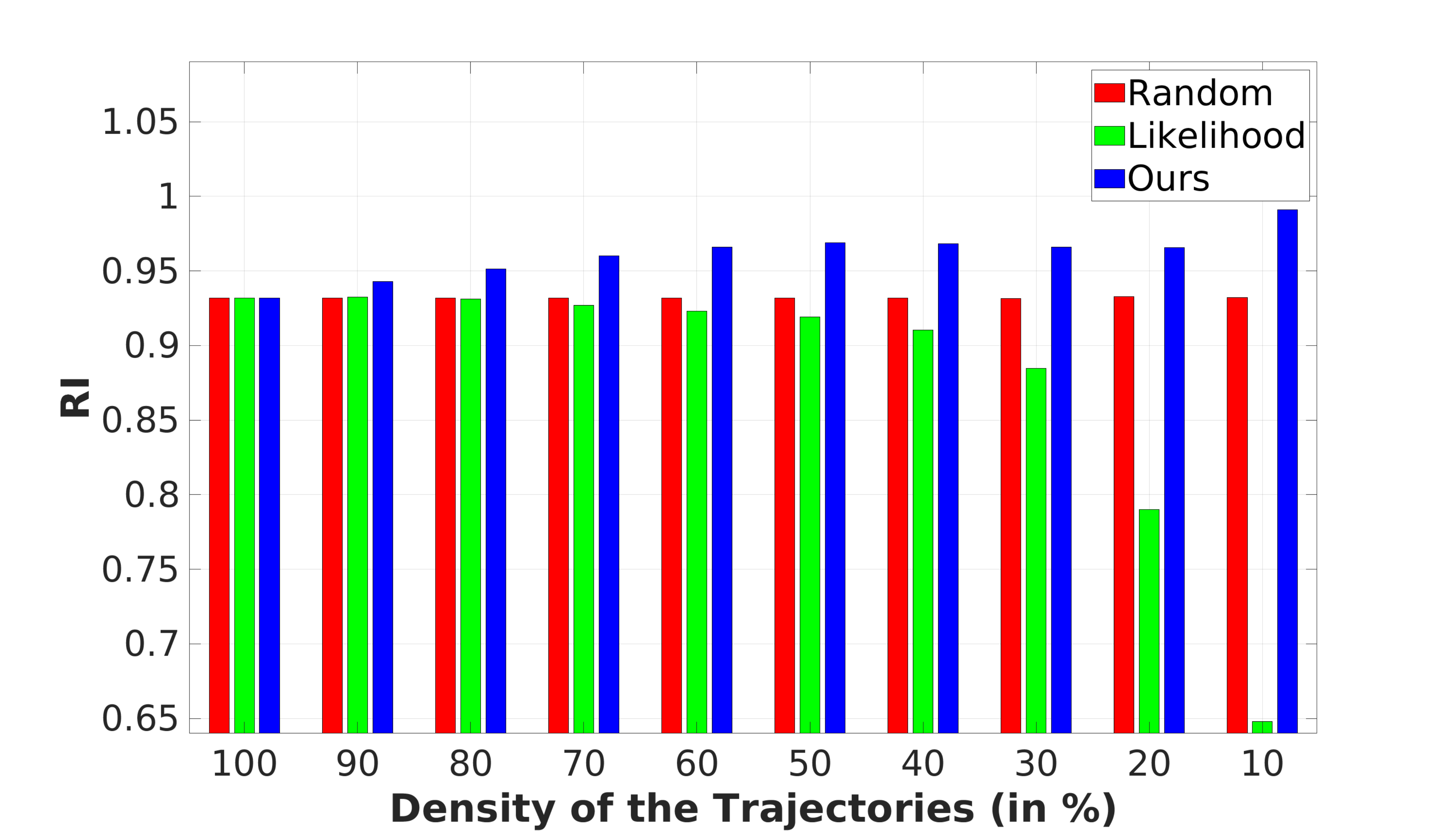}\\
     VI (lower is better)&RI (higher is better)
    \end{tabular}
    \end{center}
    \caption{Study on the motion trajectory uncertainties on VI (\textit{left}) and RI (\textit{right}) on train set (top row) and validation set (bottom row) of DAVIS$_{2016}$~\cite{davis_16}. Our results improve significantly over the baselines.} 
    \label{fig:davis_train_ri_vi_baselines}
\end{figure}
\subsection{Motion Segmentation Results}
\label{sec:experiments}
To study the proposed uncertainty measure, we first compute minimum cost multicuts on the motion segmentation instances of FBMS$_{59}$ and DAVIS$_{2016}$. On these decompositions, we compute the proposed uncertainties. In Fig. \ref{fig:davis_fbms_imgs_nodes}, we depict example images from sequences of FBMS$_{59}$ and DAVIS$_{2016}$, their ground truth segmentation, trajectory segmentation and the proposed uncertainty. The uncertainty measures are in the range of 0 (lowest uncertainty) to 1 (highest uncertainty). As can be seen, the sparse segmentations show good overall accuracy but tend to have issues on dis-occlusion areas for example behind the driving car in the last row, and on object parts under articulated motion such as the legs of the horse in the second row and the legs of the the person in the third row. In all these regions, the estimated uncertainty is high.  
%
%
%
%
Next, we assess the proposed uncertainties in terms of sparsification plots as done in \cite{optFlow_uncertain}, i.e. by subsequently removing nodes in the order of their decreasing uncertainty from the solution. 
%
%
%
%
We compare our results to the \textit{Likelihood} baseline defined in Eq.~\eqref{eq:uncertainty_eq4_2} as well as to a random baseline.
%
In the random baseline, the trajectories 
are removed subsequently at \textit{random}. In this case, neither improvement or decay of the segmentation metric should be expected.

%

\paragraph{FBMS$_{59}$} The Variation of Information (VI) and the Rand index (RI) are provided in Fig. \ref{fig:fbms_train_ri_vi_baselines} and 
FBMS$_{59}$-train and FBMS$_{59}$-test. The most uncertain trajectories are removed until 10\% of the density of the trajectories. Interestingly, removing highly uncertain trajectories accounts for reduction in VI (lower is better) and improves the RI indicating the effectiveness of the proposed uncertainty measure. The VI continuously improves as more uncertain trajectories are removed. Yet, the RI becomes unsteady when getting closer to the lowest density. Recall that the RI depends on the granularity of the solution (see \cite{metrics}) so this behavior is to be expected as more and more ground truth segments are not considered in the sparse solution. 

\paragraph{DAVIS$_{2016}$} The evaluation on the DAVIS$_{2016}$ train and validation sets in Fig.~\ref{fig:davis_train_ri_vi_baselines} 
indicates a similar behaviour as seen on FBMS$_{59}$ and shows the effect of our approach. Notice that our approach provides better uncertainty measures than the baselines (\textit{Random Baseline} and \textit{Likelihood Approach} (refer to~\ref{sec:uncertain_estim_model})). Again, the likelihood baseline even shows an increase in VI during sparsification, proving the importance of using the denominator in the Eq.~\eqref{eq:uncertainty_eq4_1}.
\begin{figure}[t!]
    \begin{center}
    \small
    \begin{tabular}{@{}c@{}c@{}c@{}c@{}}
    \includegraphics[width=0.24\linewidth]{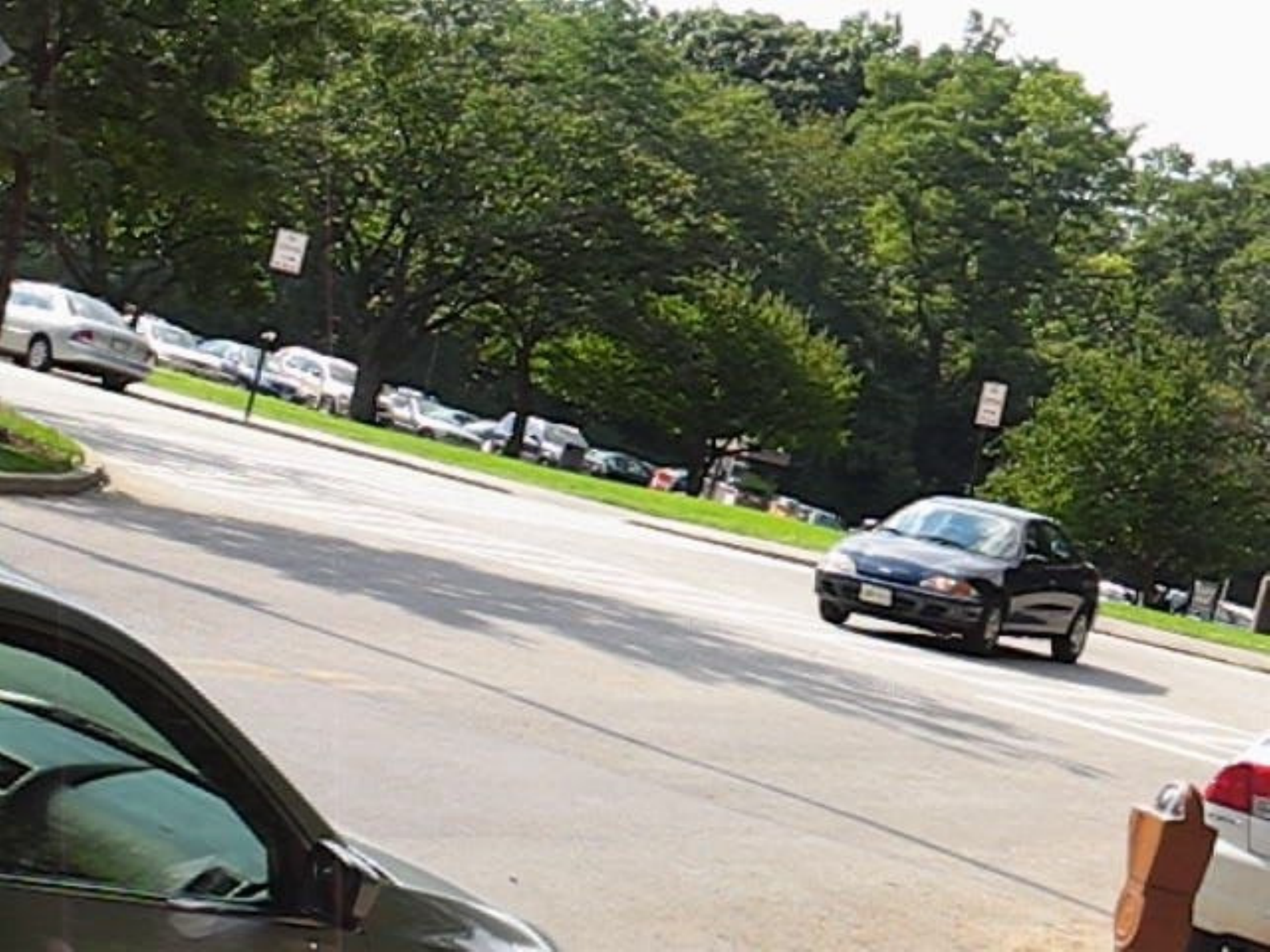}&
    \includegraphics[width=0.24\linewidth]{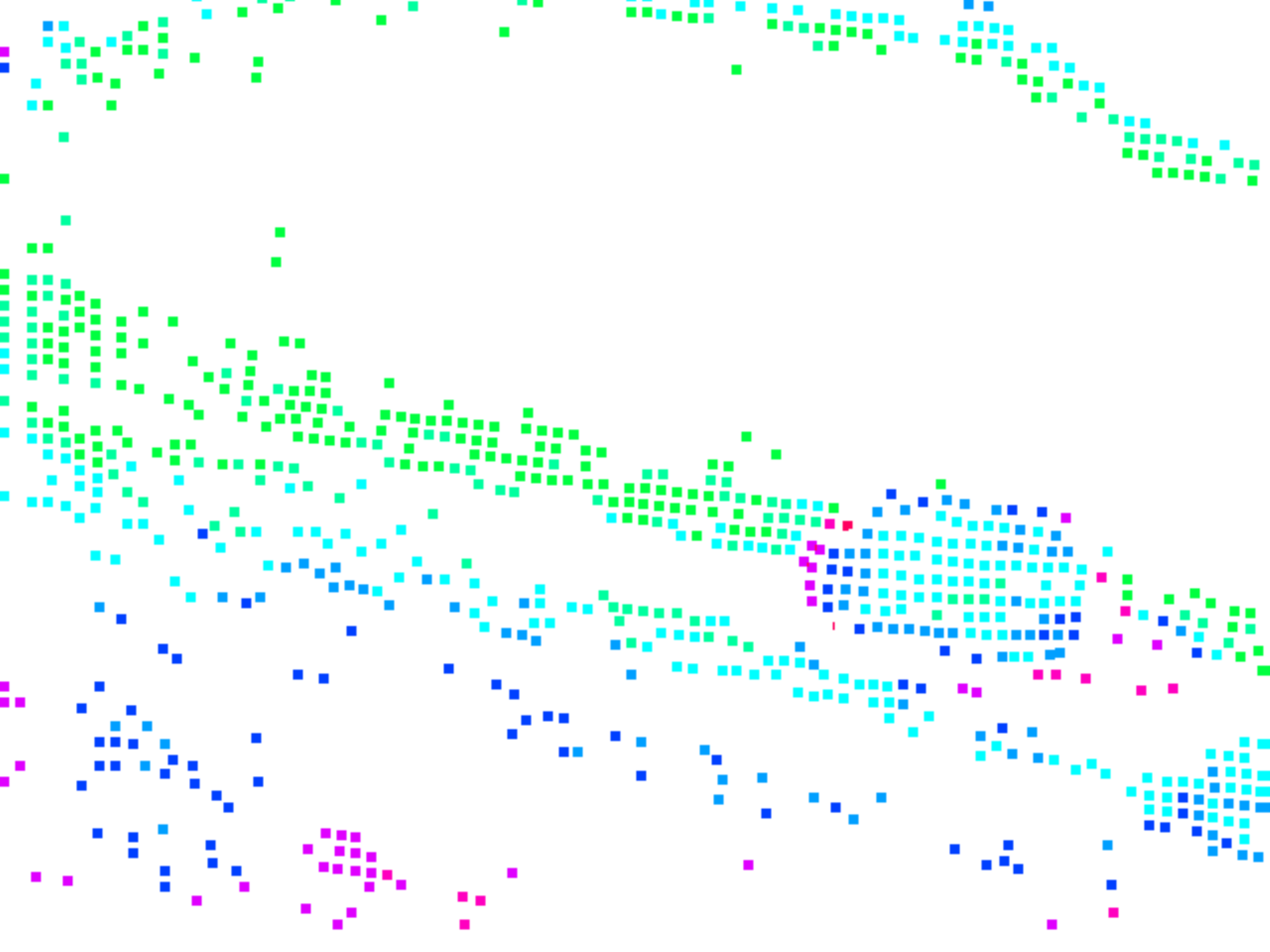}&
    &
    \\
    image&uncertainty&\\
    \includegraphics[width=0.24\linewidth]{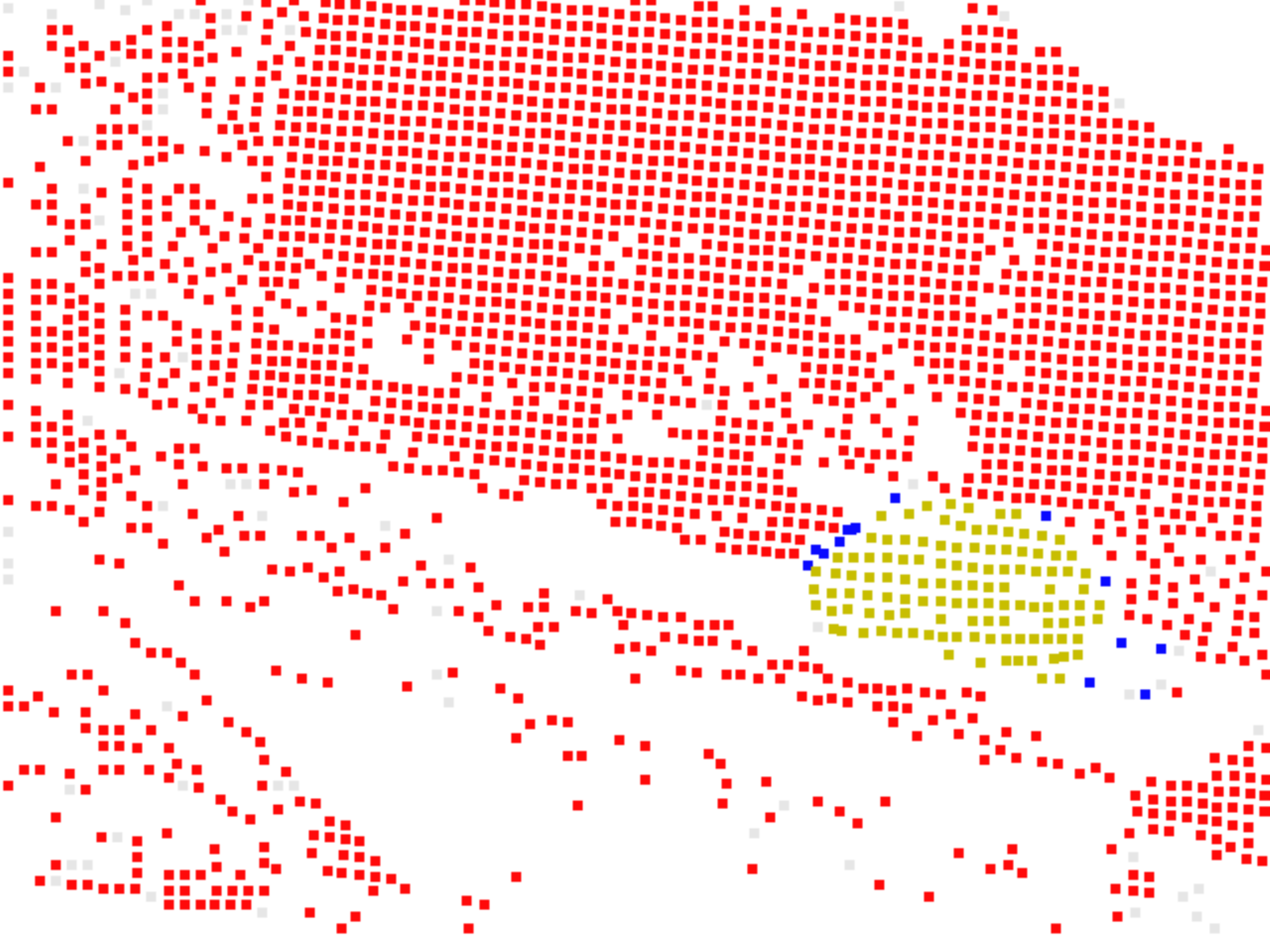}&
    \includegraphics[width=0.24\linewidth]{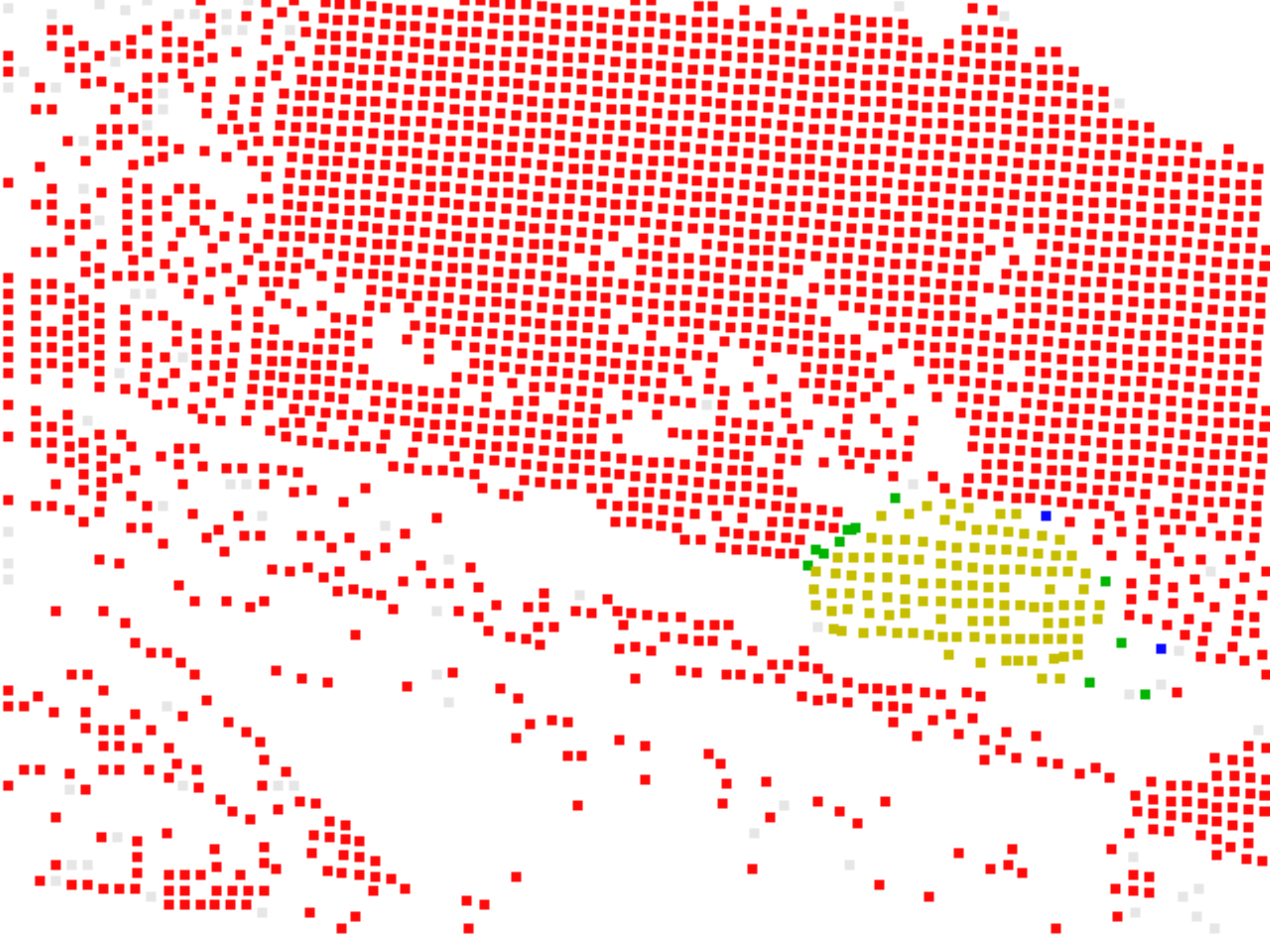}&
    \includegraphics[width=0.24\linewidth]{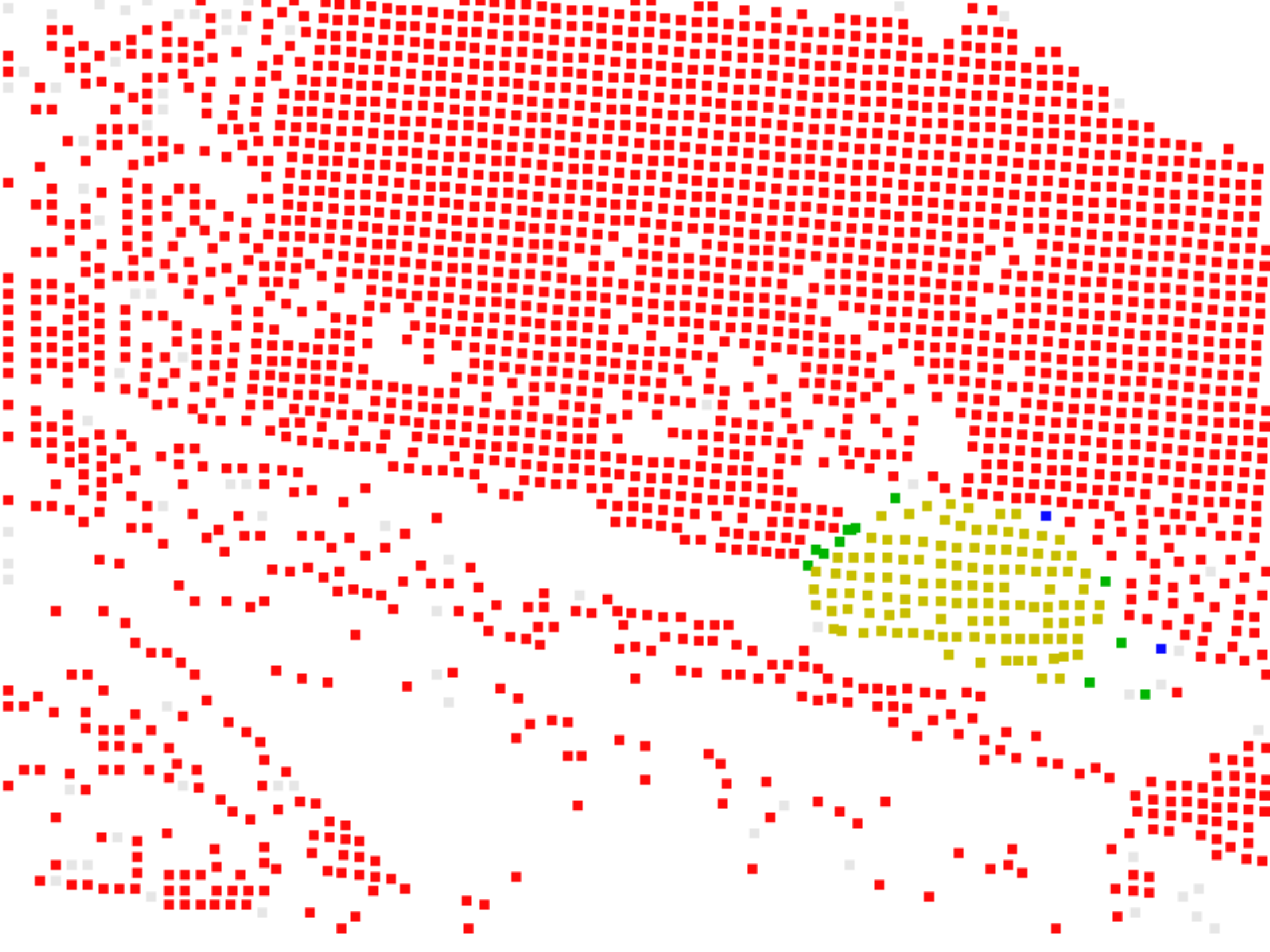}&
    \includegraphics[width=0.24\linewidth]{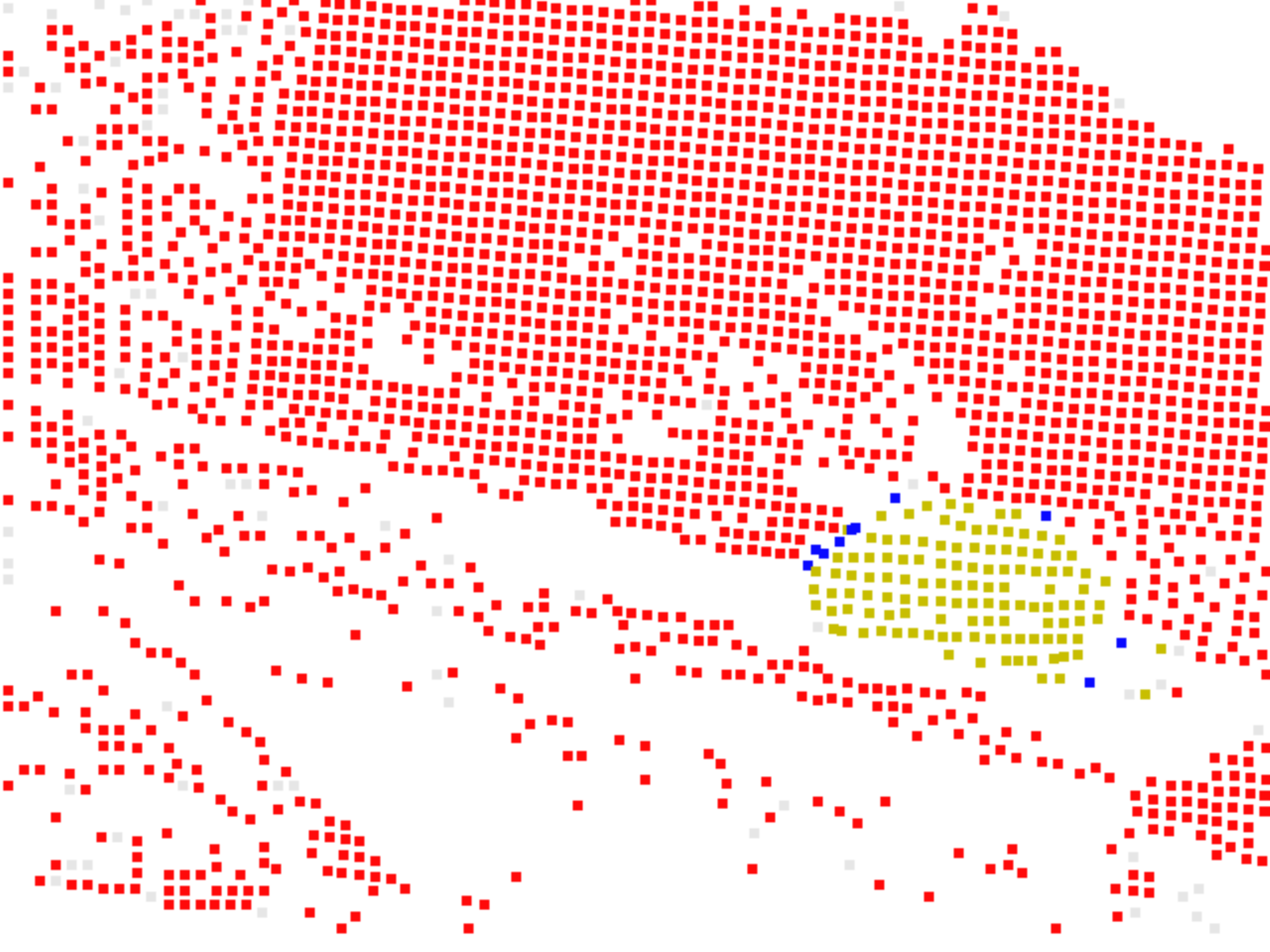}
    \\
    -127910  & -127894  &   -127894 & -127205 \\
    \includegraphics[width=0.24\linewidth]{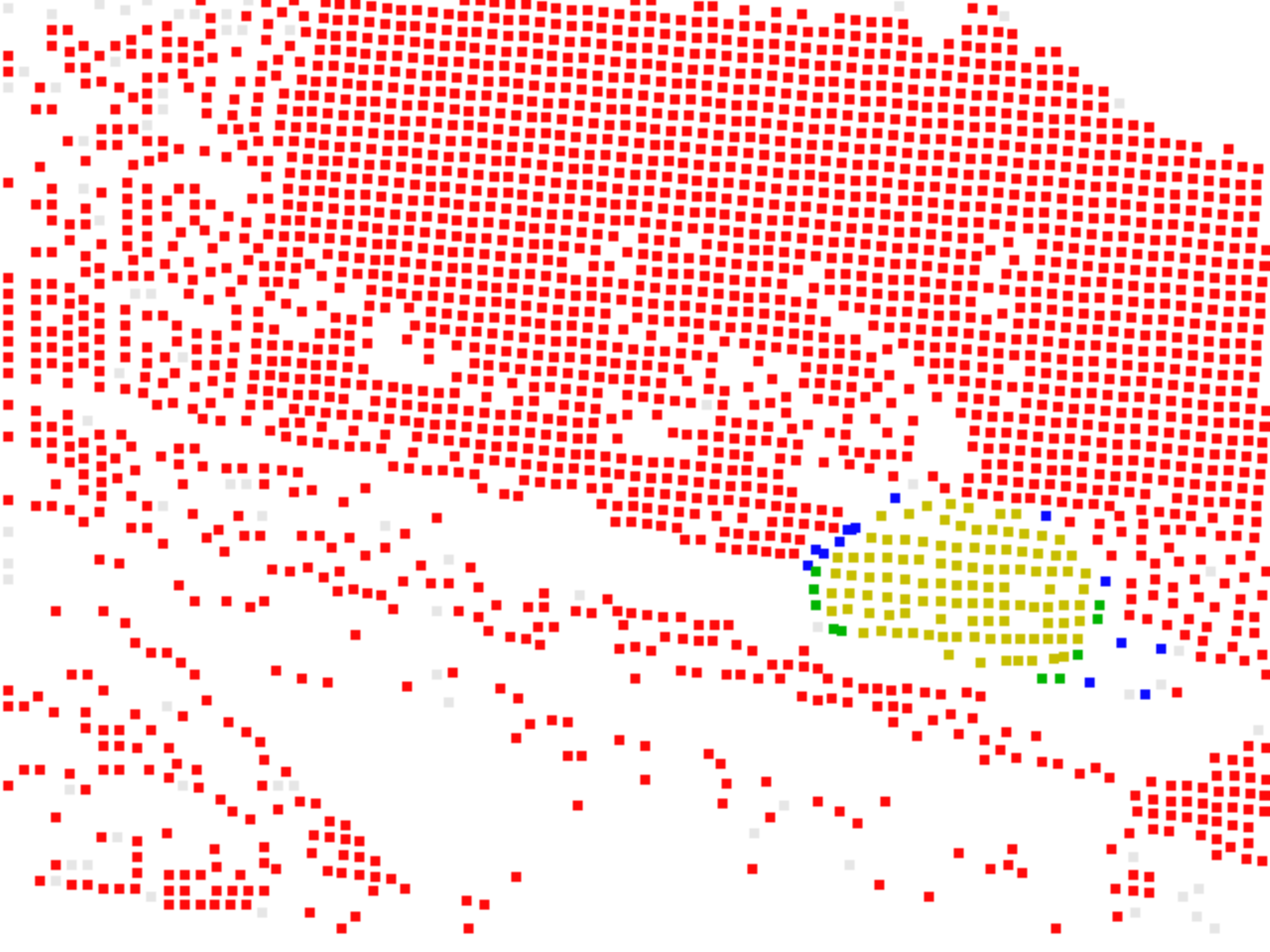}&
    \includegraphics[width=0.24\linewidth]{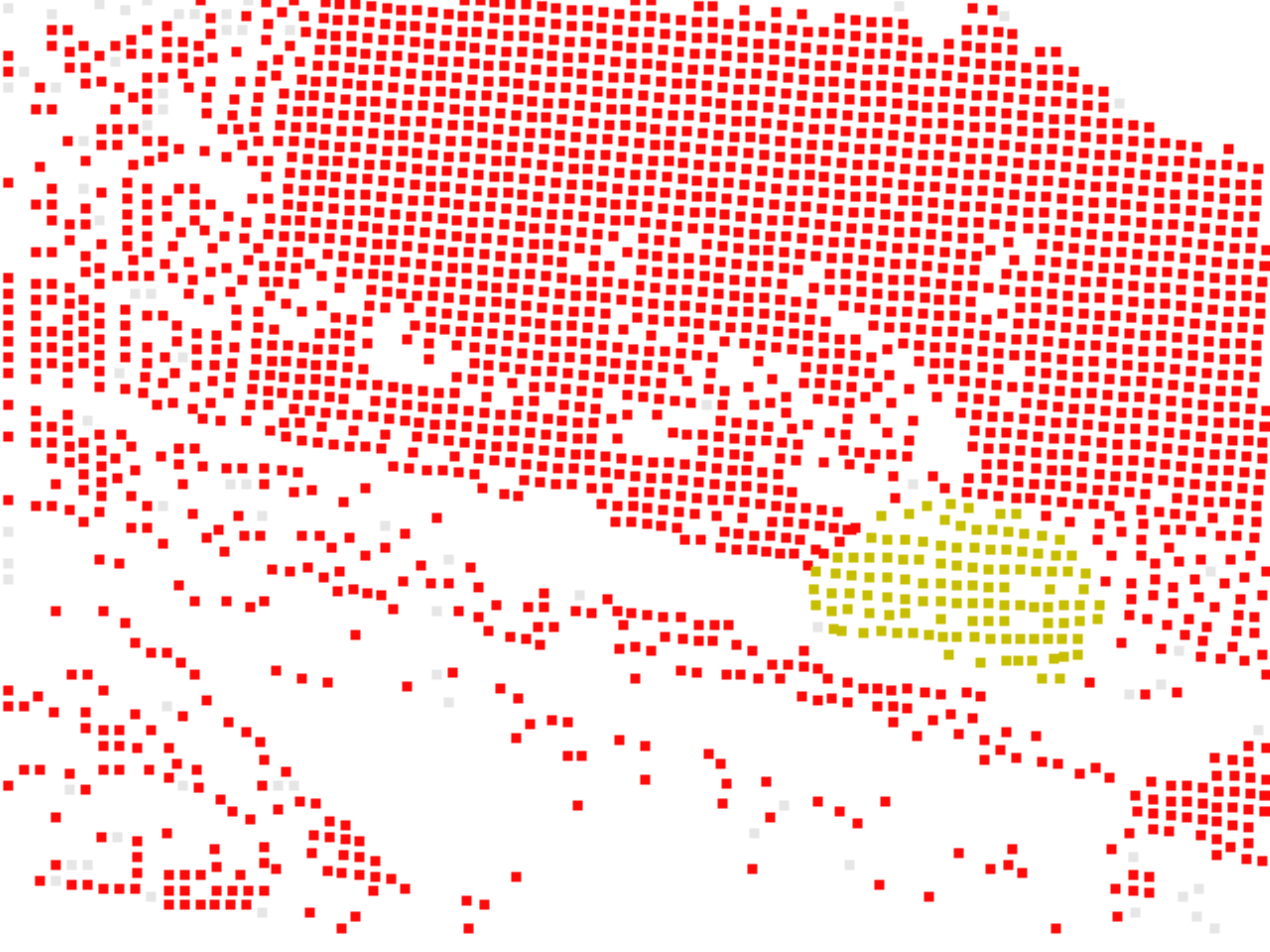}&
    \includegraphics[width=0.24\linewidth]{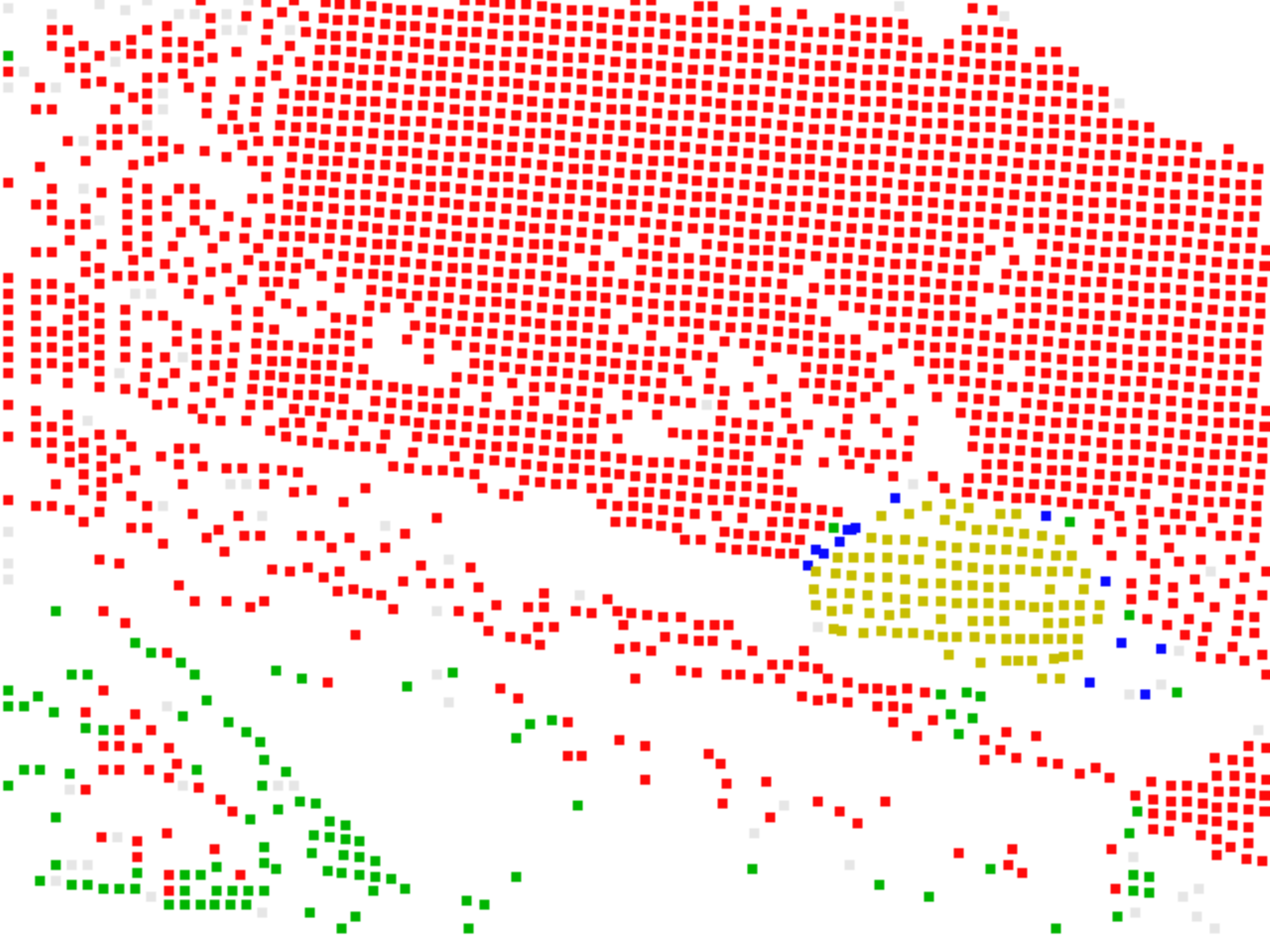}&
    \includegraphics[width=0.24\linewidth]{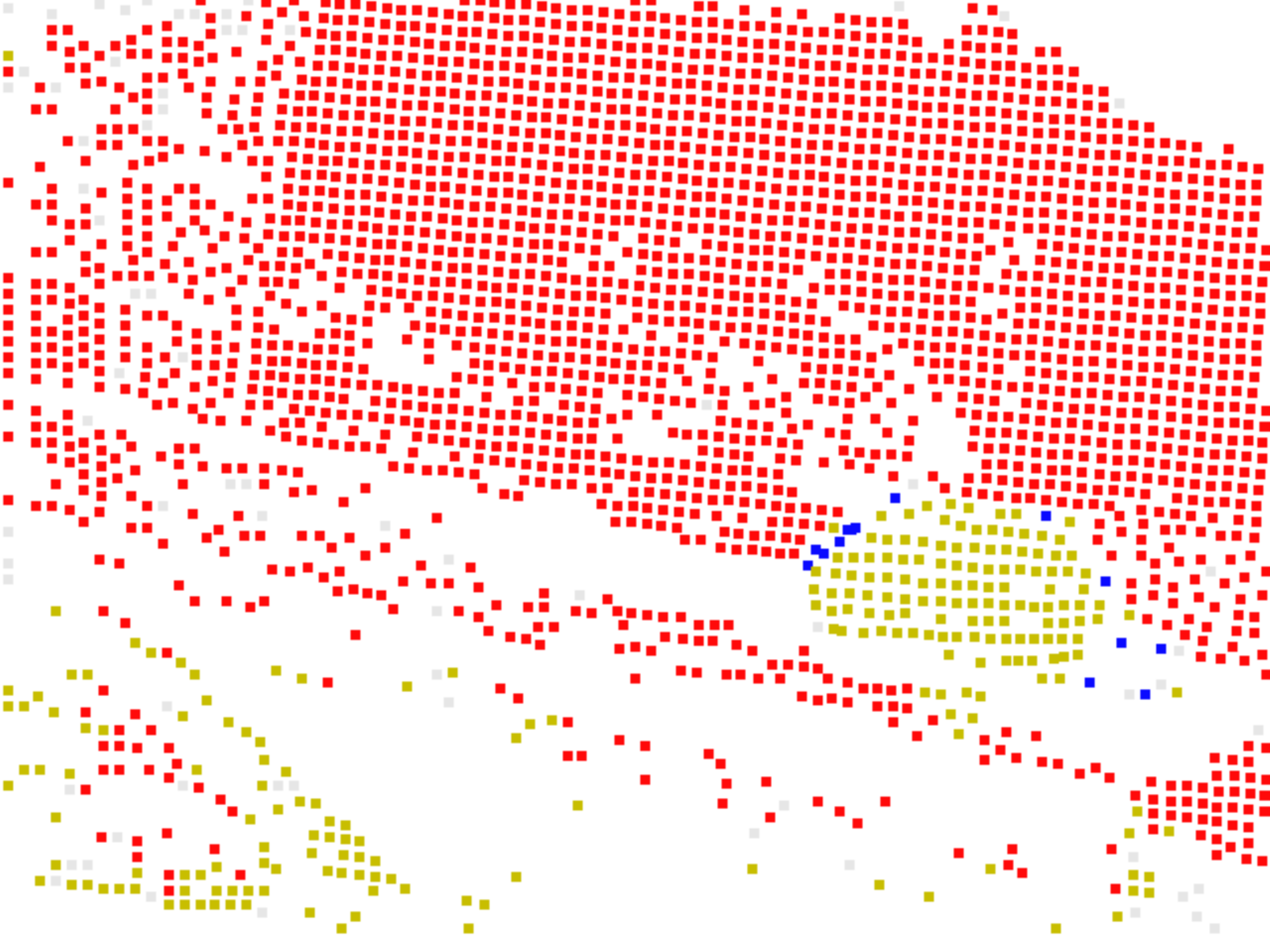} \\
    -126093 & -121703 & -111809 & -108825 \\
    \end{tabular}
    \end{center}
    \caption{Visualization of eight different likely solutions and their energies (Eq.~\eqref{eq:LMC1}, lower is better), as they can be generated by the proposed method. 
    The different solution candidates vary mostly along object boundaries.
    The best segmentation w.r.t. the ground truth corresponds to the second image (from left) in the last row.}
    \label{fig:fbms_uncerainty_ordered}
\end{figure}

\begin{figure}[t]
    \begin{center}
    \includegraphics[width=0.6\linewidth]{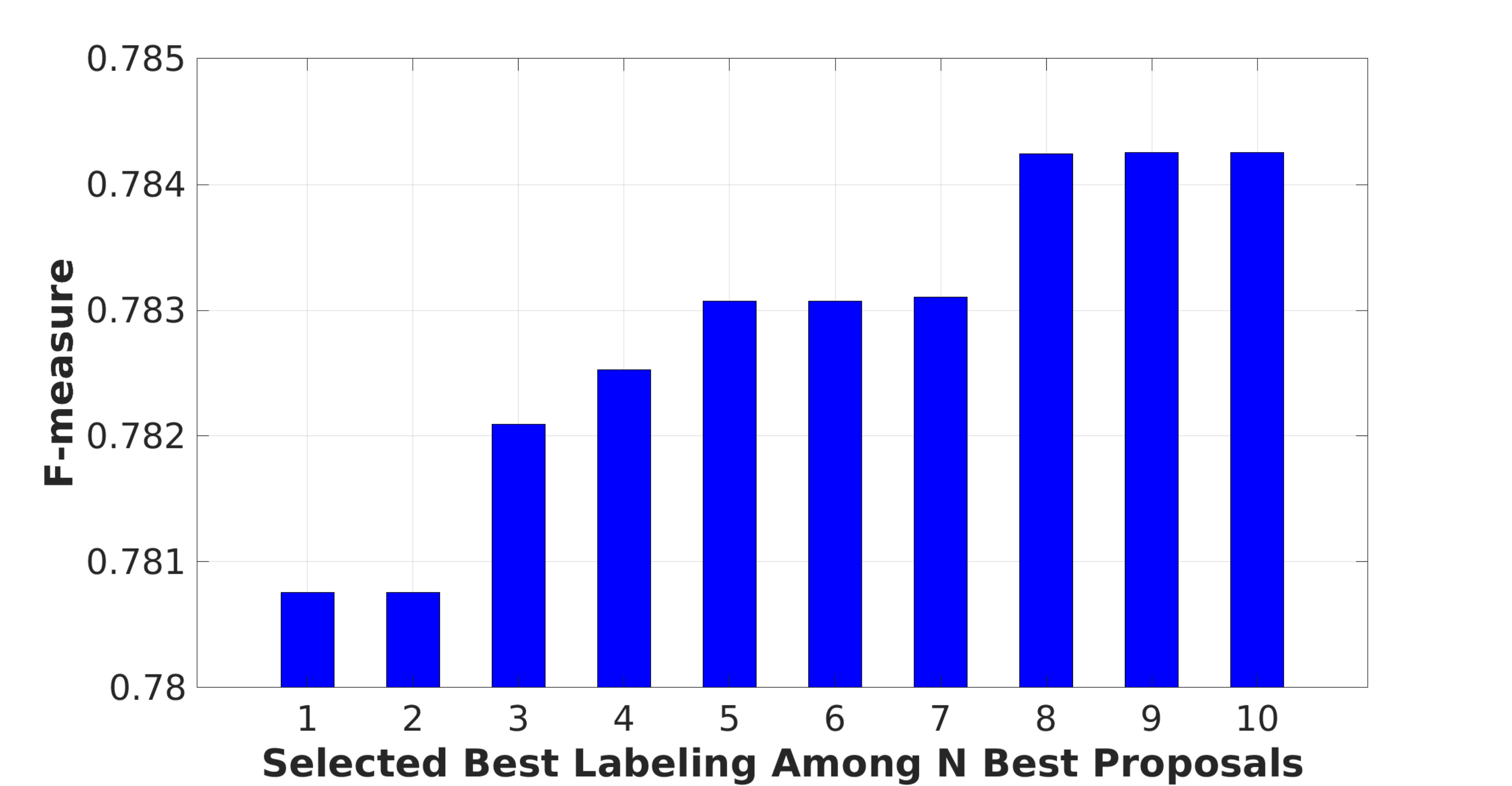}\\
    \end{center}
    \caption{ F-measure on the train set of FBMS$_{59}$ when selecting $n$ best segmentation proposals. The F-measure improves as the number of segmentation candidates increases.} 
    \label{fig:fbms_train_best_label}
\end{figure}

\subsection{Multimodal Motion Segmentation}
\label{subsec:multimodal_moseg}
In the following, we briefly sketch how the proposed approach can be leveraged to generate multiple, likely solutions. After the termination of the heuristic solver, the nodes in the graph $G = (V, E)$ are moved between different clusters to compute the cost differences in Eq. \eqref{eq:costDifference}. Given the $N$ clusters, the node from cluster $A$ is moved to other $N-1$ clusters and the cost change is computed. Therefore, for each node, a vector with the number of different partitions minus one ($N-1$) is computed. This leads to $\dfrac{N\times(N-1)}{2}$ vectors, each with $|V|$ values. To produce multiple segmentations, such vectors are ordered based on their magnitude, i.e. their associated energy increase.  Each such vector 
holds the costs for moving nodes from cluster $A$ to cluster $B$. The label of the node $v$ is changed from cluster $A$ to cluster $B$ if $\Delta cost(A,B)_v \approx 0$. With this approach, it is possible to generate $n$ best clusterings, which differ in the labels of uncertain points. 
In Fig.~\ref{fig:fbms_uncerainty_ordered}, we visualize an example of eight such potential solutions and their energies. The variance is strongest on the object boundaries and on the car in the foreground. It coincides with the estimated uncertainties.  
In Fig.~\ref{fig:fbms_train_best_label}, we allow the evaluation script of FBMS$_{59}$ to choose the best among the $n$ most likely solutions for $n \in \{1, ..., 10\}$. By taking more solutions into account, the F-measure improves.

\subsection{Densified Motion Segmentation}
\label{subsec:effect_on_moseg}

\begin{figure}[t!]
    \begin{center}
    \small
    \begin{tabular}{@{}c@{}c@{}c@{}c@{}c@{}}
    \includegraphics[width=0.2\linewidth]{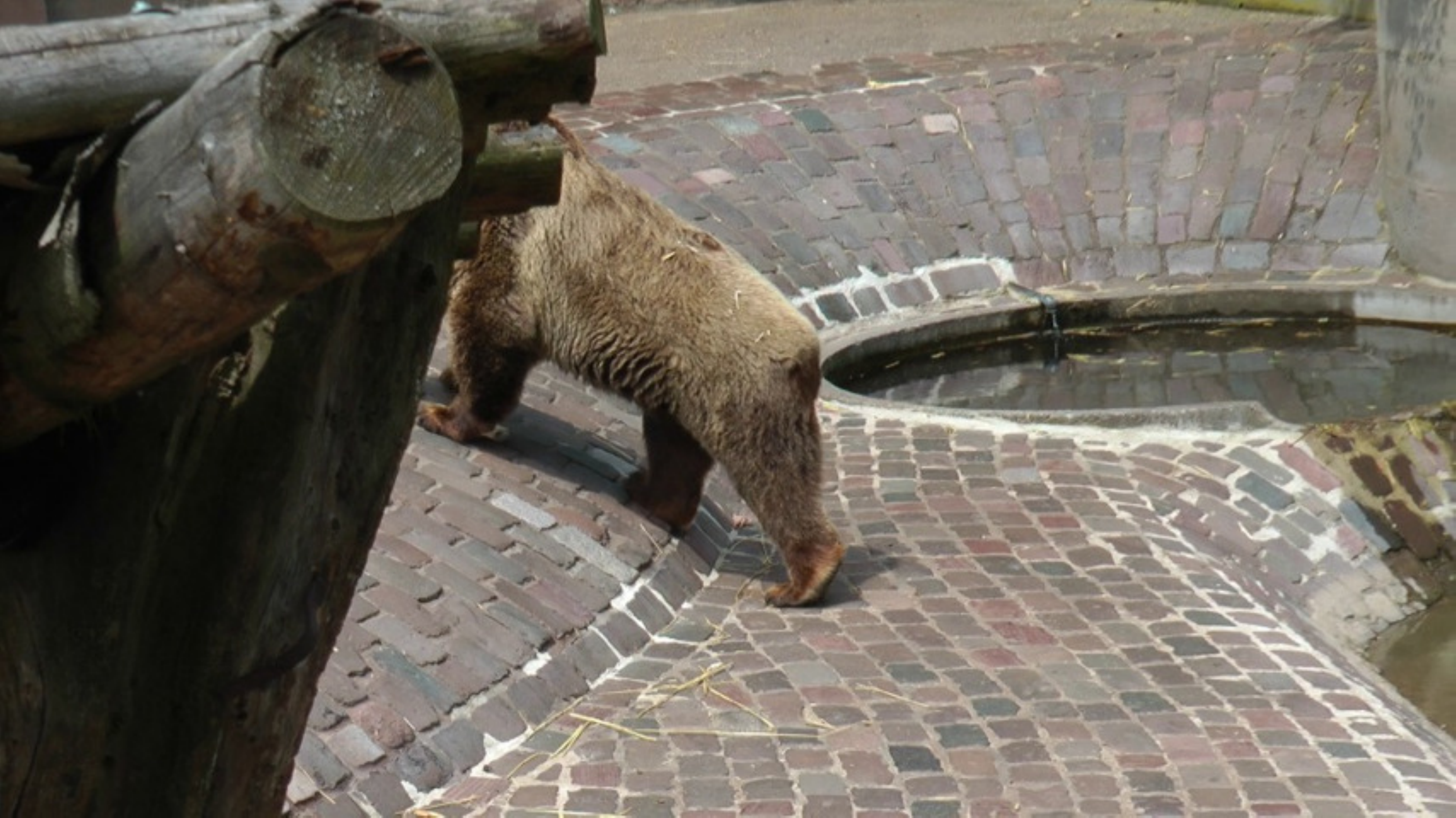}&
    \includegraphics[width=0.2\linewidth]{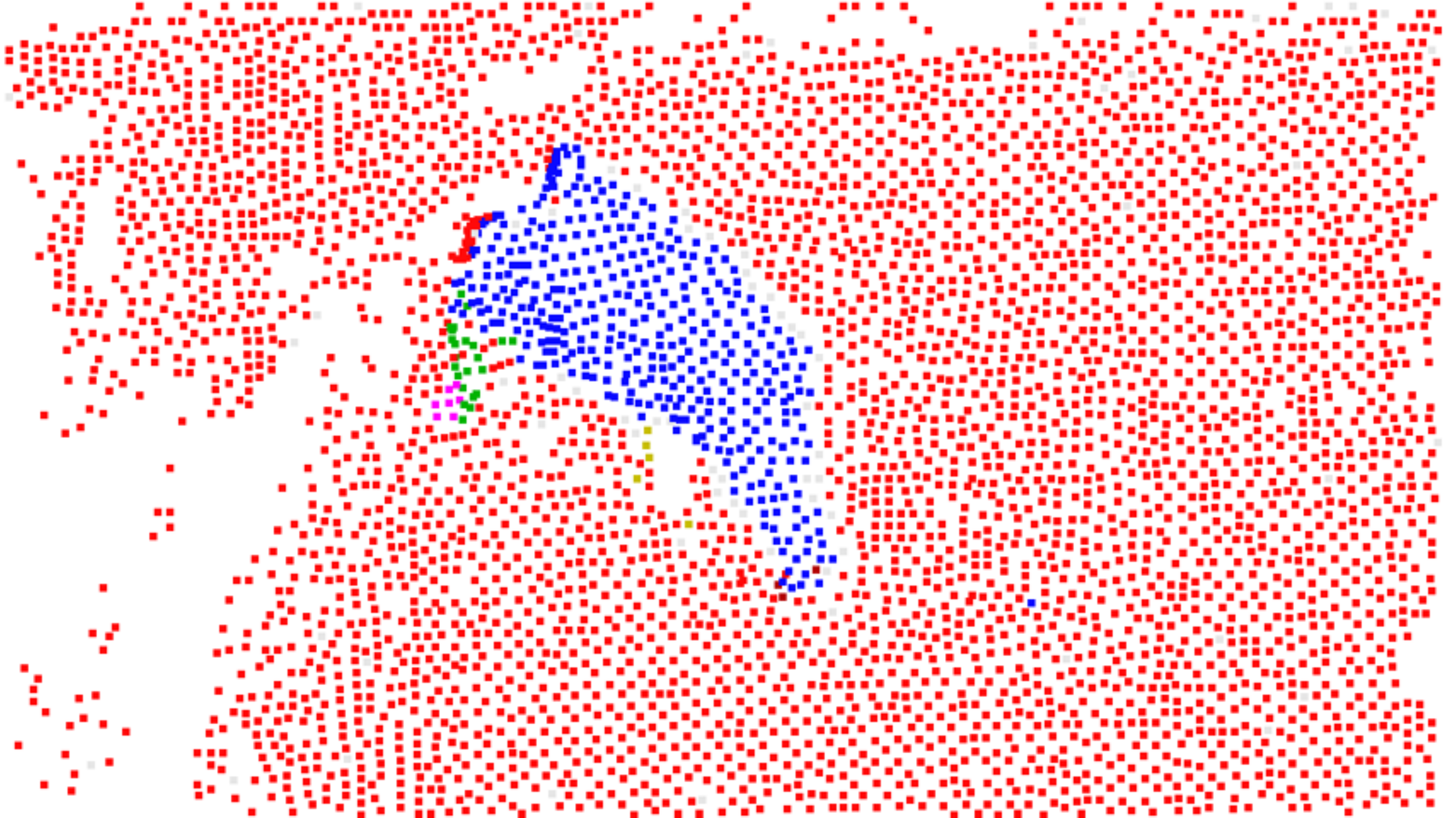}&
    \includegraphics[width=0.2\linewidth]{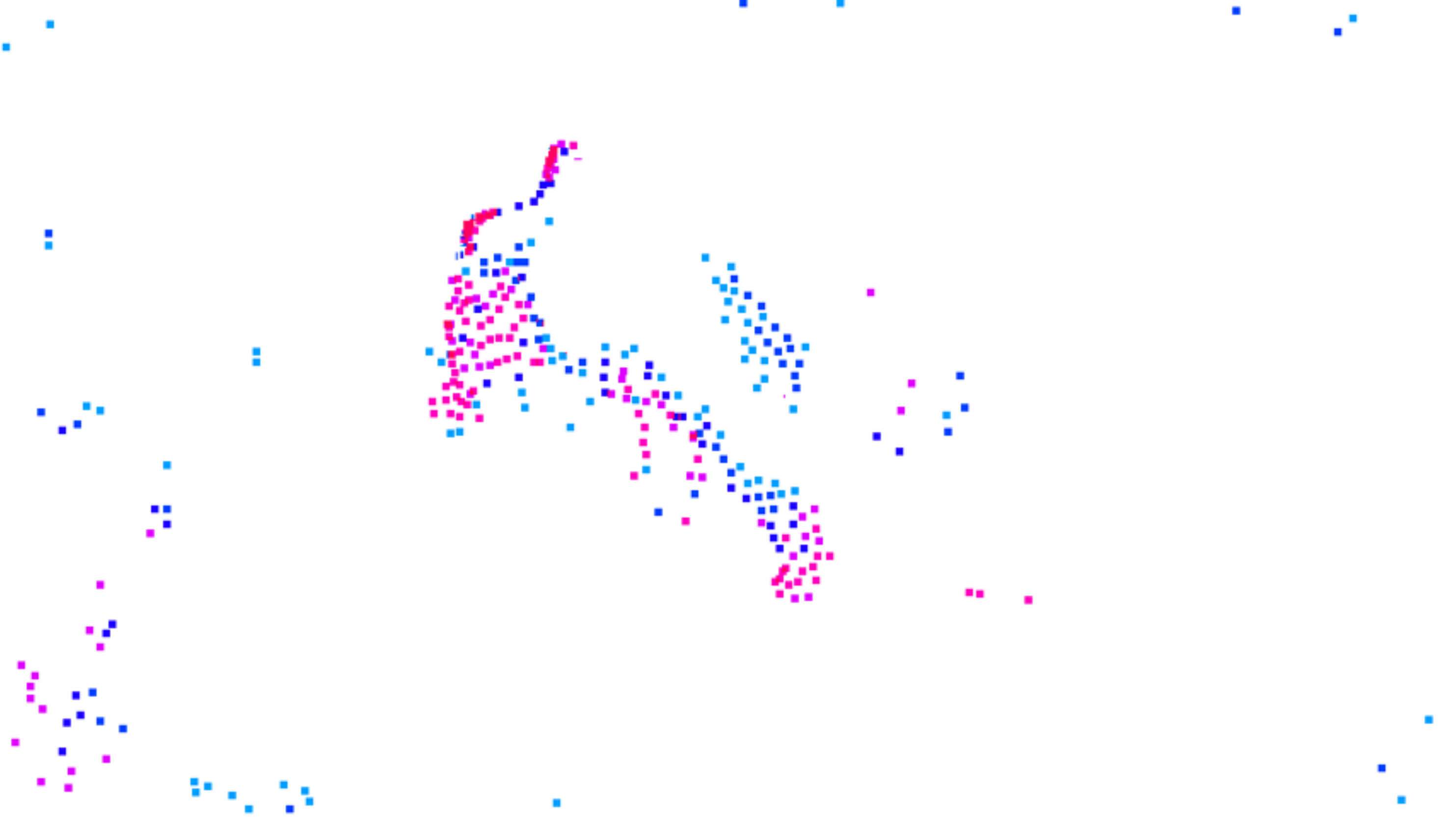}&
    \includegraphics[width=0.2\linewidth]{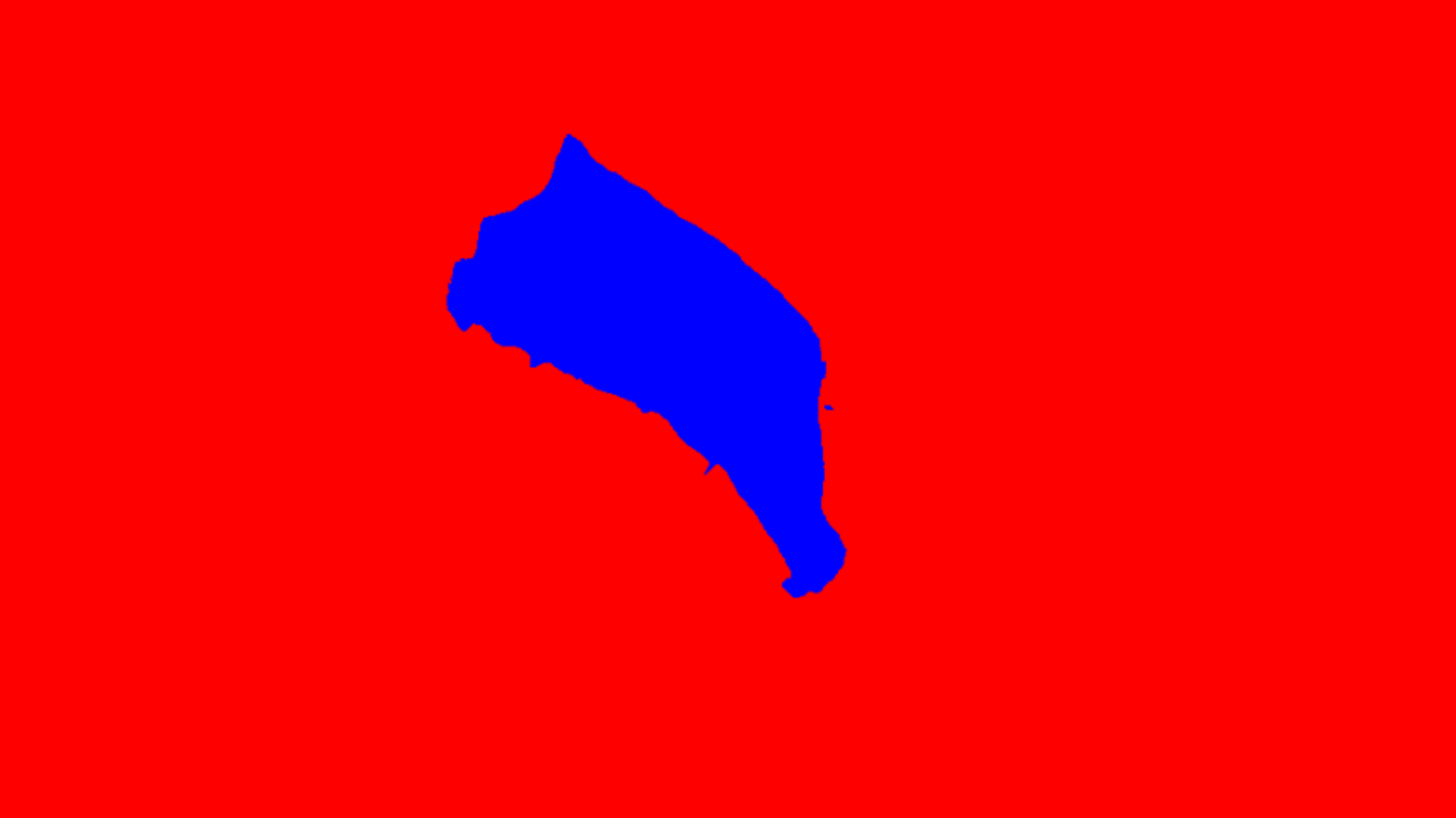}&
    \includegraphics[width=0.2\linewidth]{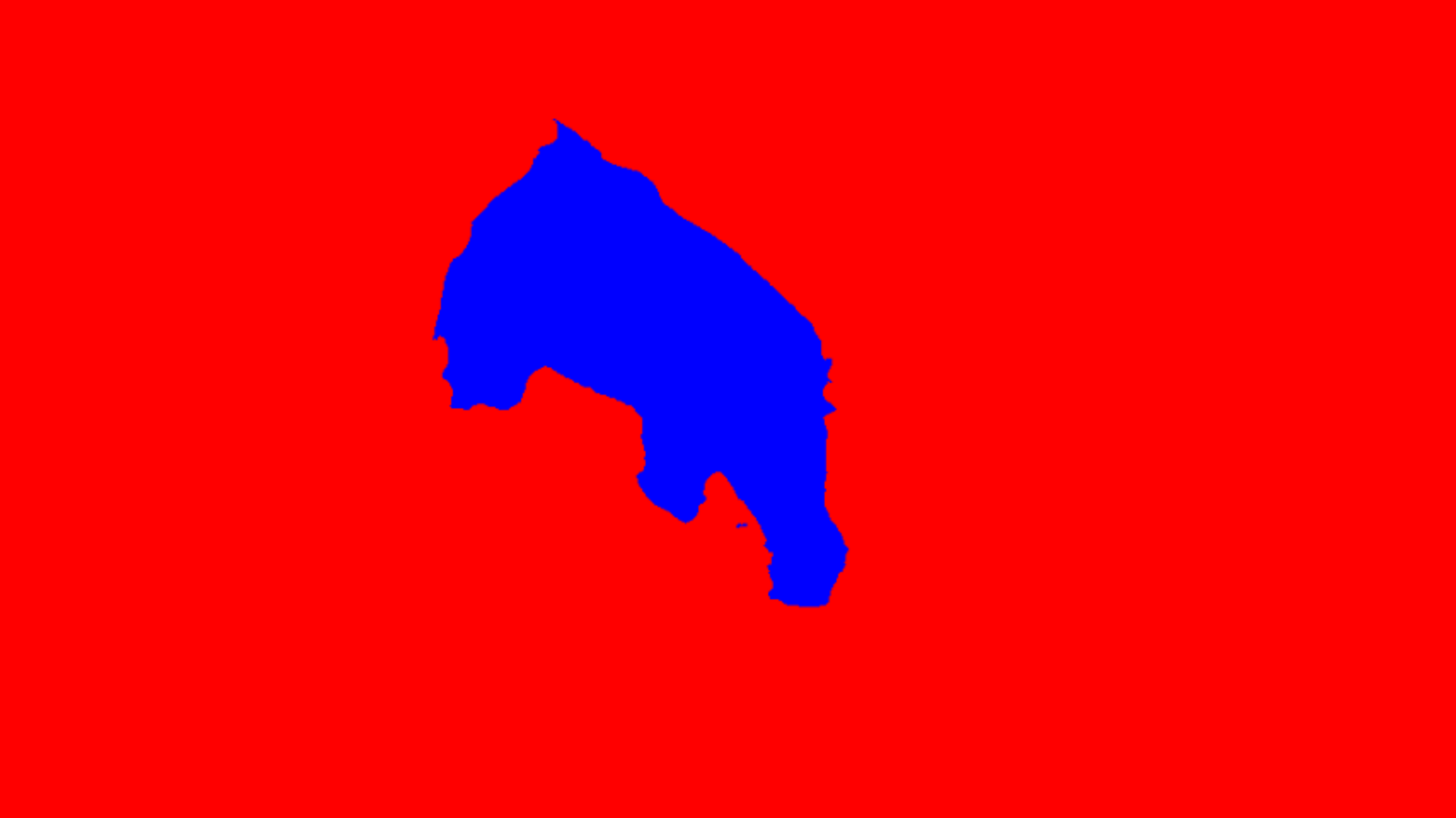}\\
    \includegraphics[width=0.2\linewidth]{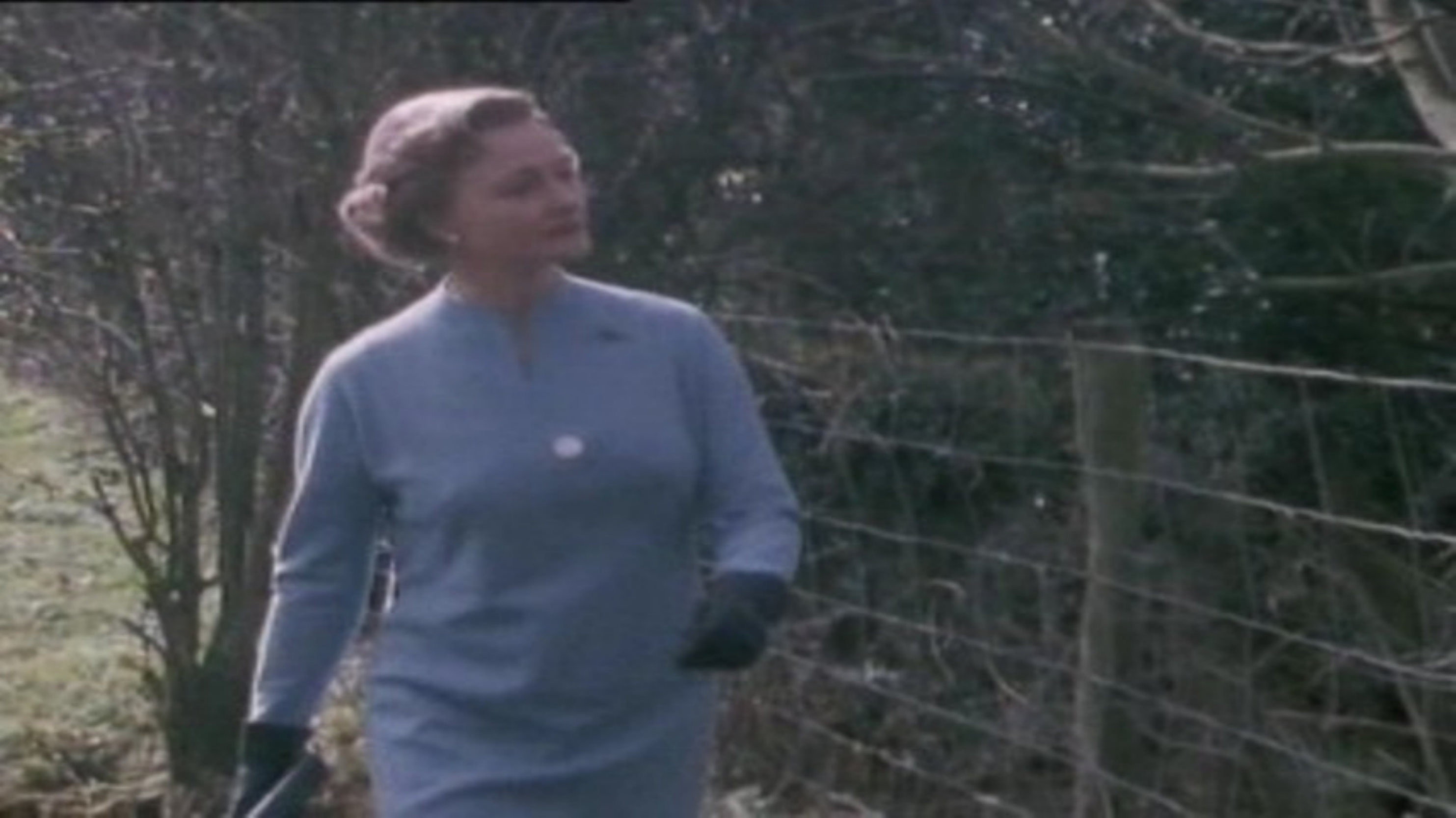}&
    \includegraphics[width=0.2\linewidth]{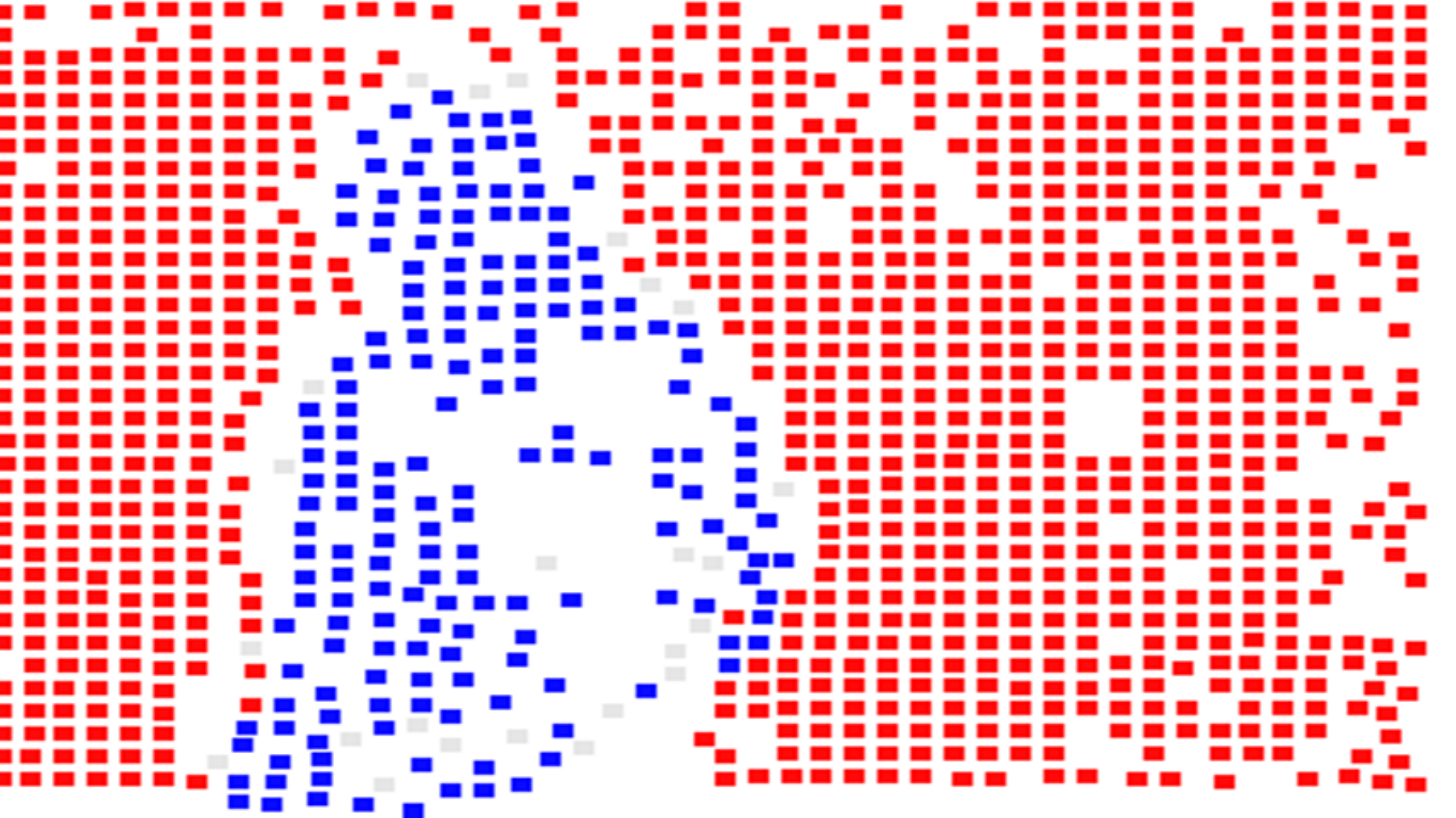}&
    \includegraphics[width=0.2\linewidth]{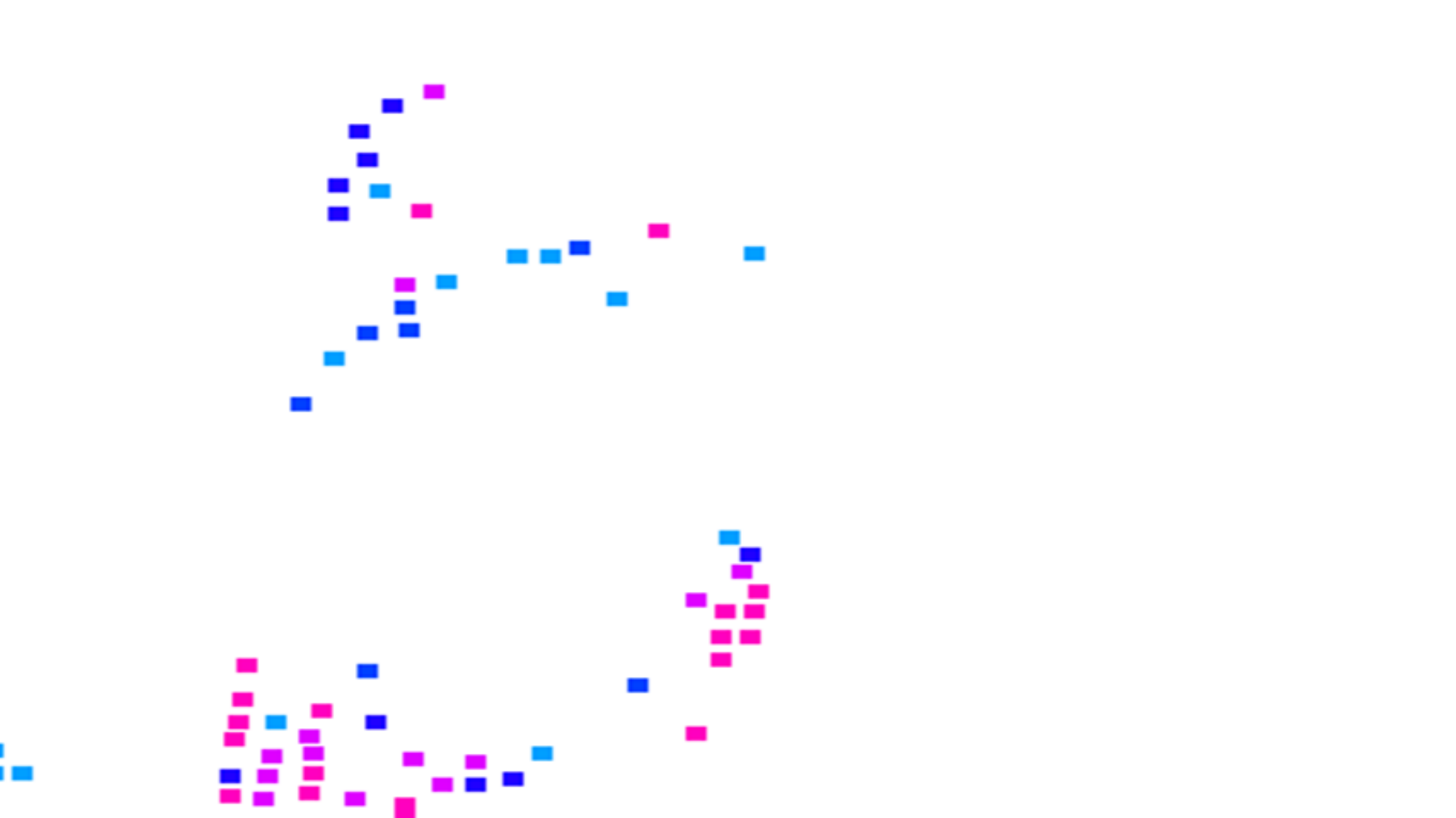}&
    \includegraphics[width=0.2\linewidth]{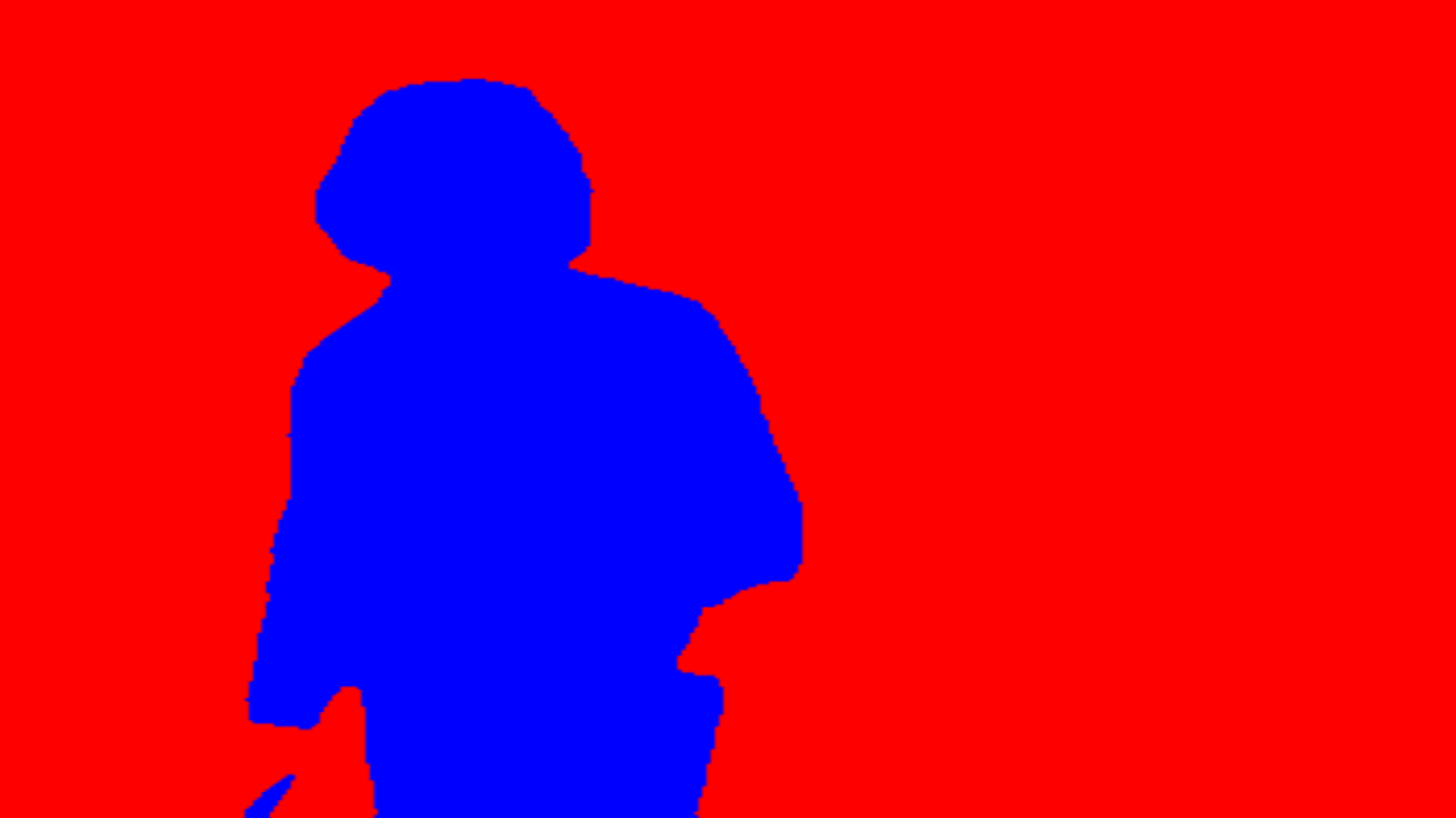}&
    \includegraphics[width=0.2\linewidth]{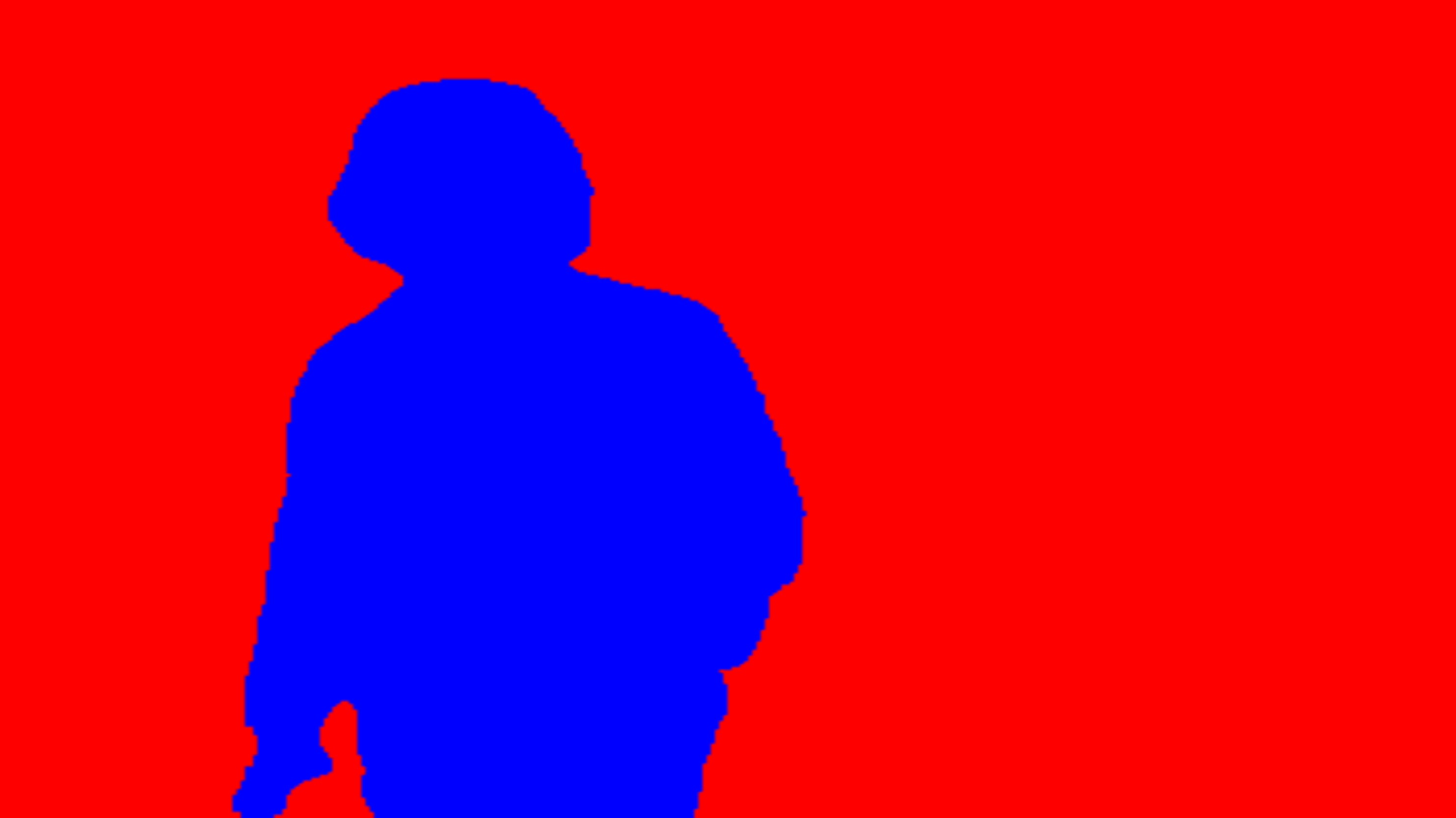}\\
    \includegraphics[width=0.2\linewidth]{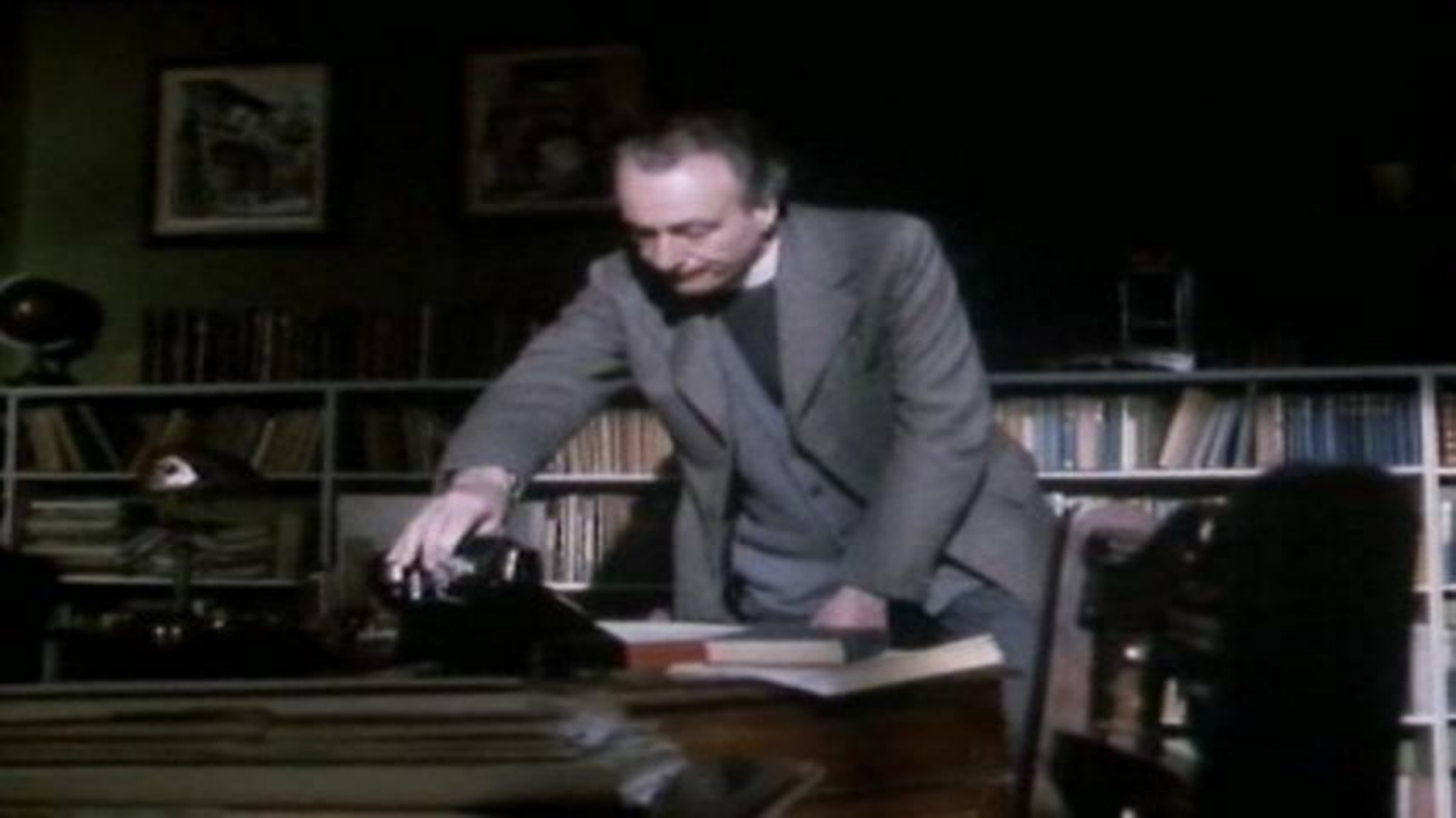}&
    \includegraphics[width=0.2\linewidth]{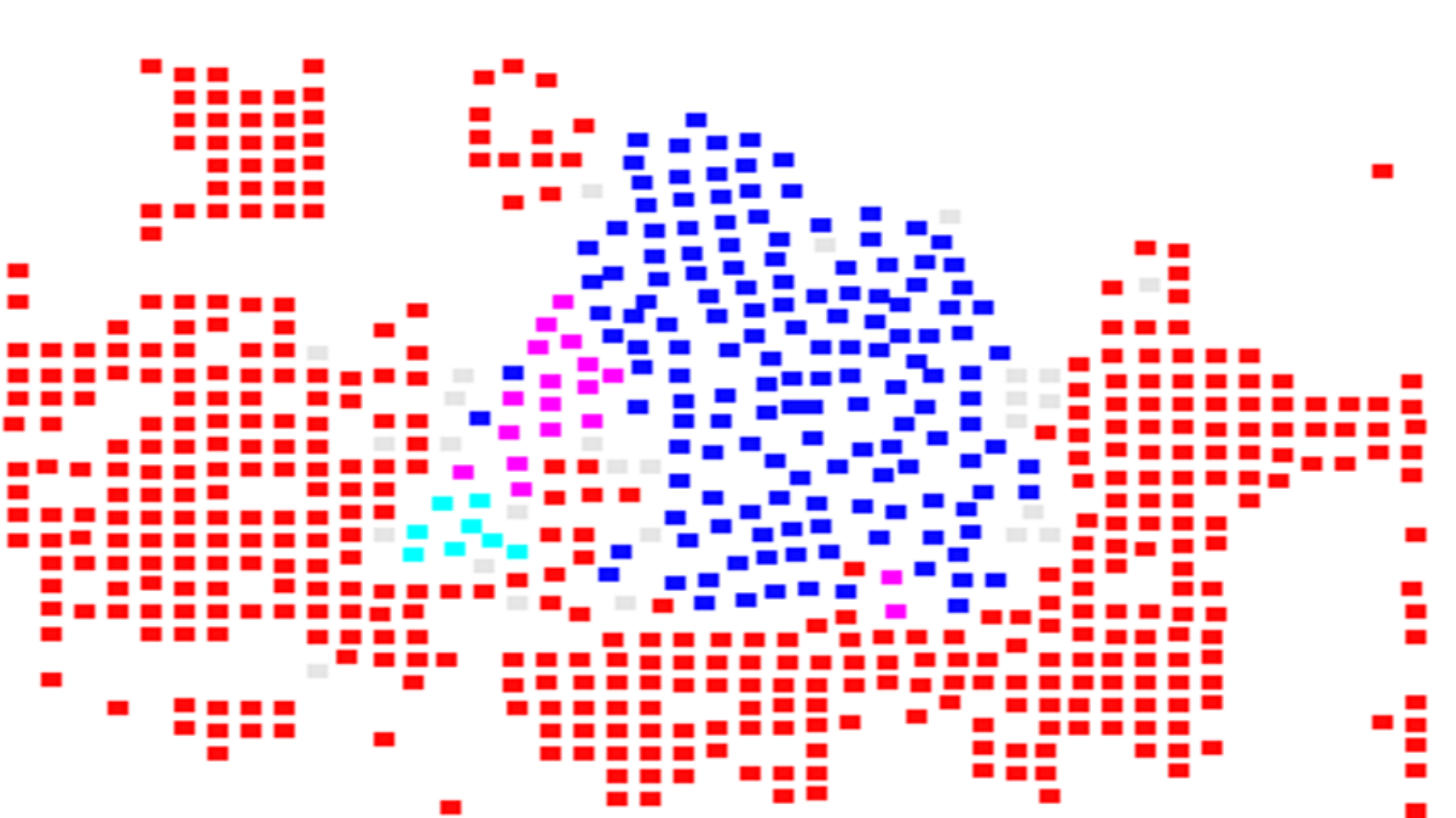}&
    \includegraphics[width=0.2\linewidth]{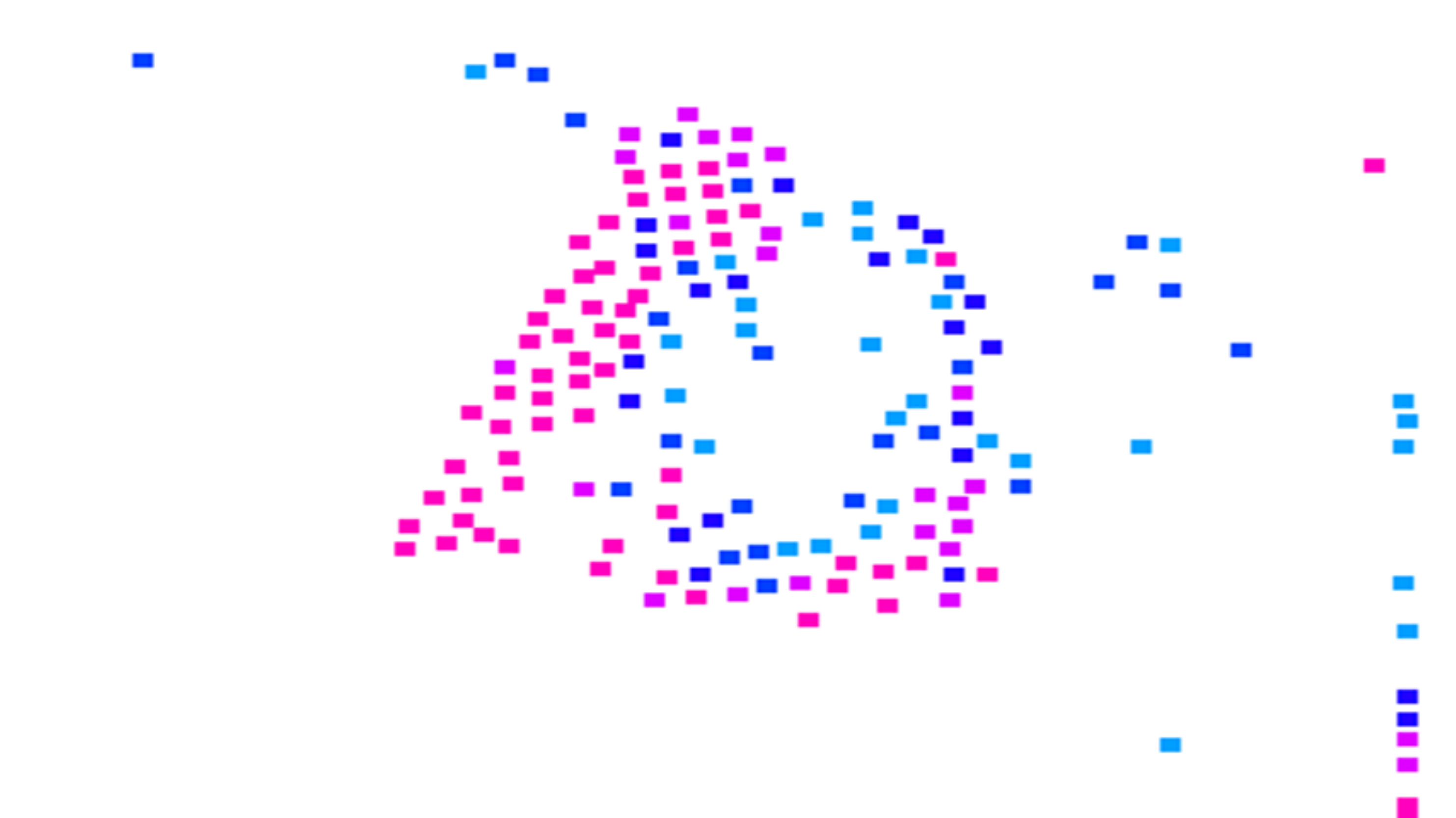}&
    \includegraphics[width=0.2\linewidth]{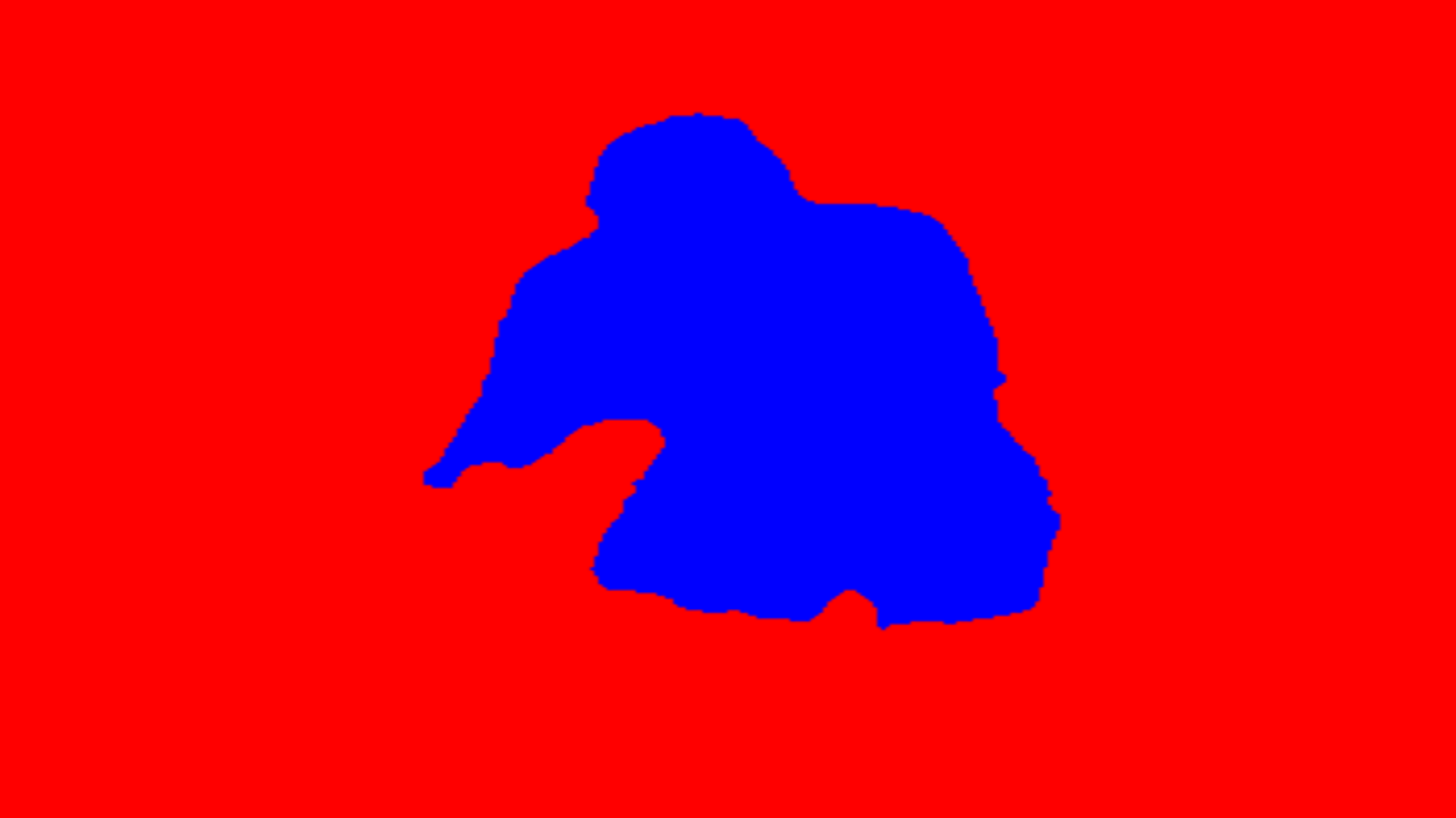}&
    \includegraphics[width=0.2\linewidth]{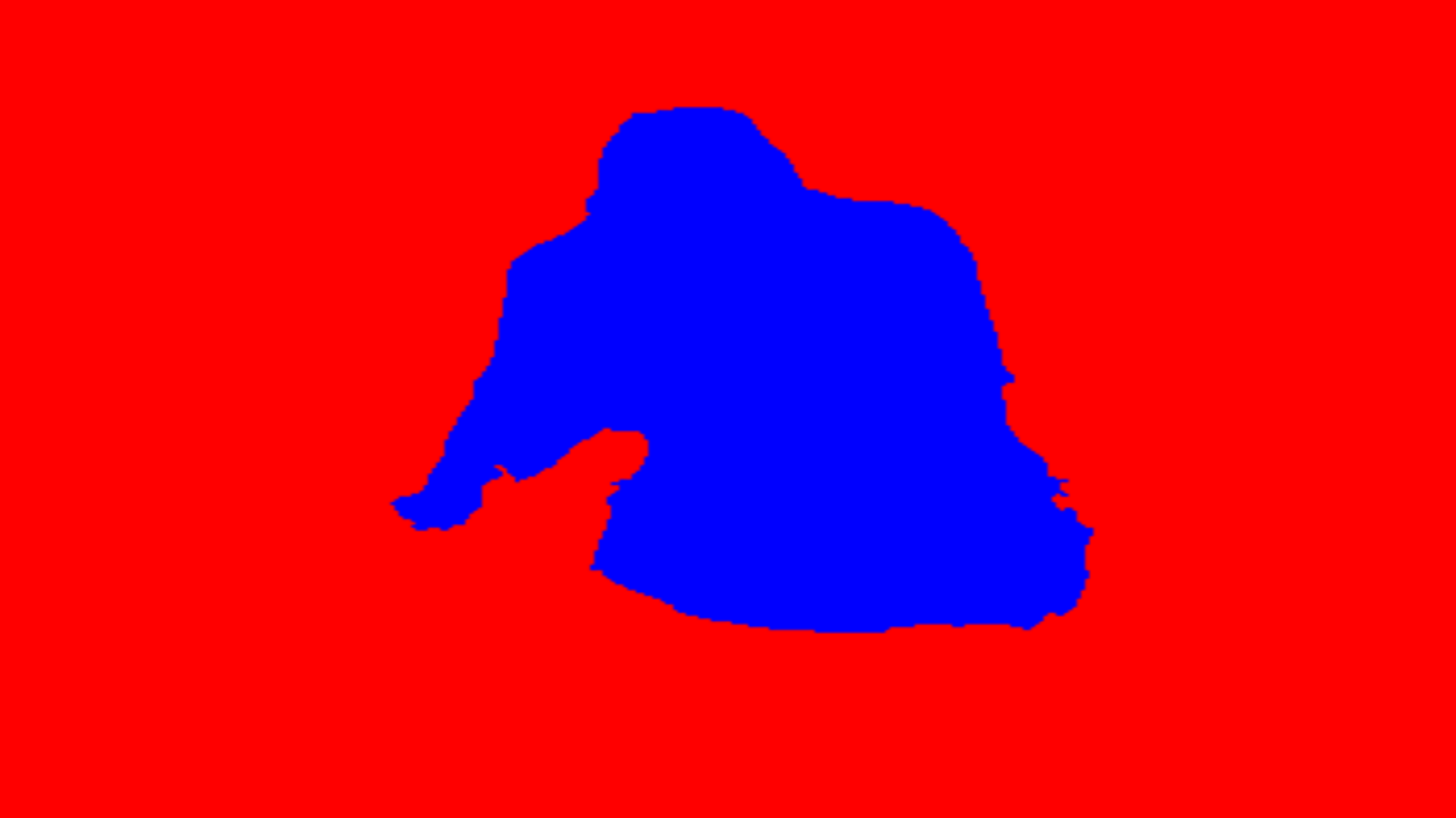}\\
    image&multicut&uncertainty&baseline&proposed
    \end{tabular}
    \end{center}
    \caption{Densification of sparse segmentations using uncertainties. We compare the result from~\cite{selfsupervised} (baseline) with the proposed method, which uses uncertainties in the model training. Improvements can be observed especially on thin structures such as limbs. 
    }
    \label{fig:fbms_davis_uncertain_combined}
\end{figure}
\begin{table}[t!]
\centering
\caption{Densification of the sparse trajectory segmentations on the FBMS$_{59}$ train set. We compare the model of ~\citet{selfsupervised} to the variant trained using uncertainties.}
\small
\begin{tabular}{l|c|c|c}
\toprule
                             & Precision     & Recall       & F-measure  \\ \midrule
~\cite{selfsupervised}     & 89.35         & 67.67        & 77.01      \\
Ours (uncertainty-aware)                 & 88.17         & 68.96        & 77.40      \\
\bottomrule
\end{tabular}
\label{fbms_uncertainty_added}
\end{table}

%
%
To create actual pixel-level segmentations from sparse segmentations a separate model needs to be learned. 
For example~\citet{Ochs14} propose a variational model while~\citet{selfsupervised} employs a U-Net~\cite{unet} based model, which is trained in a self-supervised manner, using the sparsely segmented trajectories as labels. In the following, we employ our uncertainties in order to 
enhance the self-supervised training signal in this model in the FBMB59 dataset. 
%
%
As a pre-processing, \cite{selfsupervised} remove small segments from the training data such that the remaining uncertain points are mostly on thin, articulated object parts. For each segment $A$ in the decomposition of graph $G$, the mean uncertainty is computed as $\mu_A$. The highly uncertain nodes, where $u_v > \mu_A$, in the highly uncertain segments $\mu_A > \phi$ are assumed to be hard examples for the network and are therefore used more often than other points during training (with factor 1000).
%
In Tab.~\ref{fbms_uncertainty_added}, we report Precision, Recall and F-measure of the proposed training and the original model from \citet{selfsupervised}. Using this simple trick, the F-Measure can be improved by 0.4\%. 
Qualitative results are provided in Fig.~\ref{fig:fbms_davis_uncertain_combined}.

\subsection{Image Decomposition}
\begin{figure}[t]
    \begin{center}
    \begin{tabular}{@{}cccc@{}}
    \includegraphics[width=0.20\linewidth]{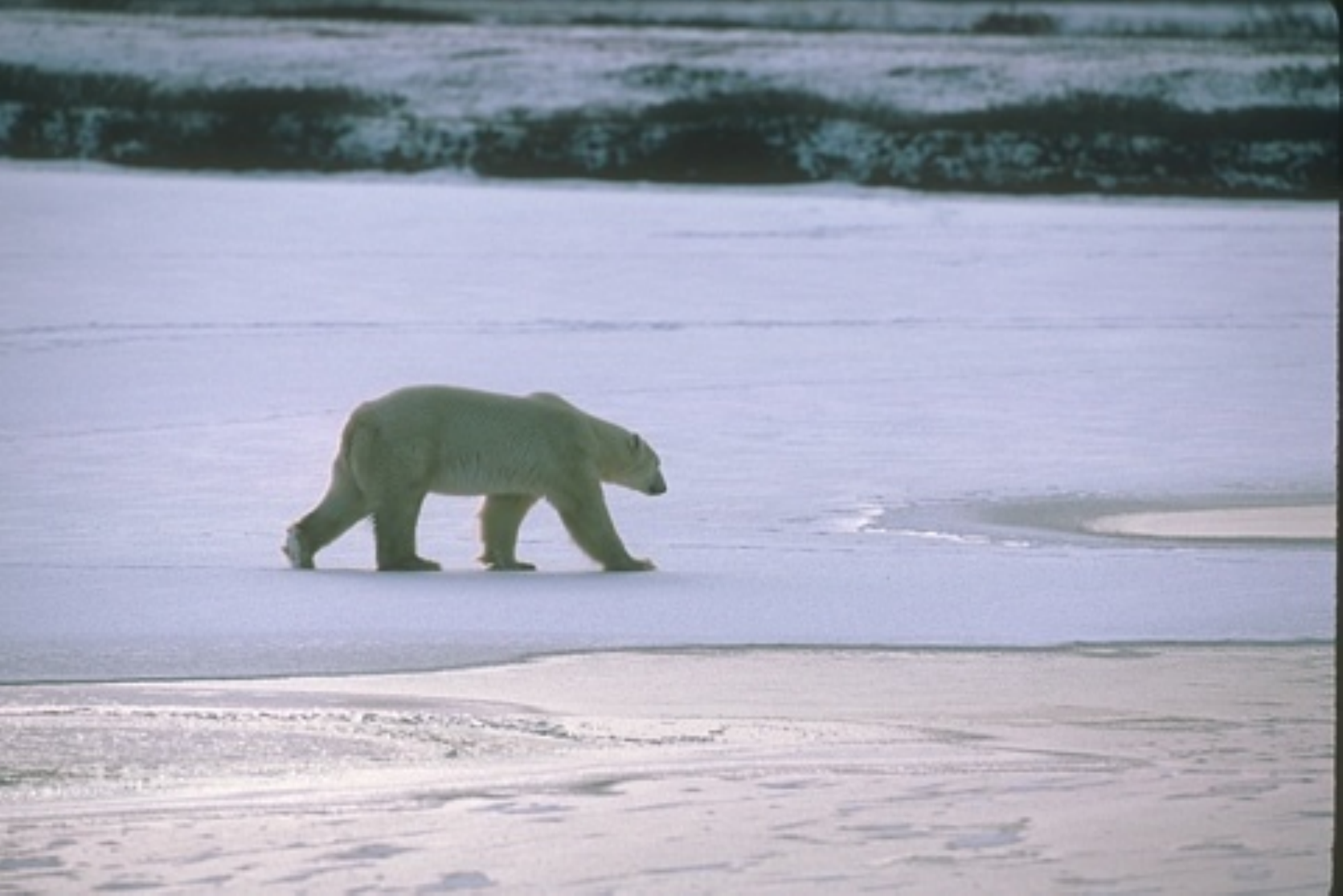}&
    \includegraphics[width=0.20\linewidth]{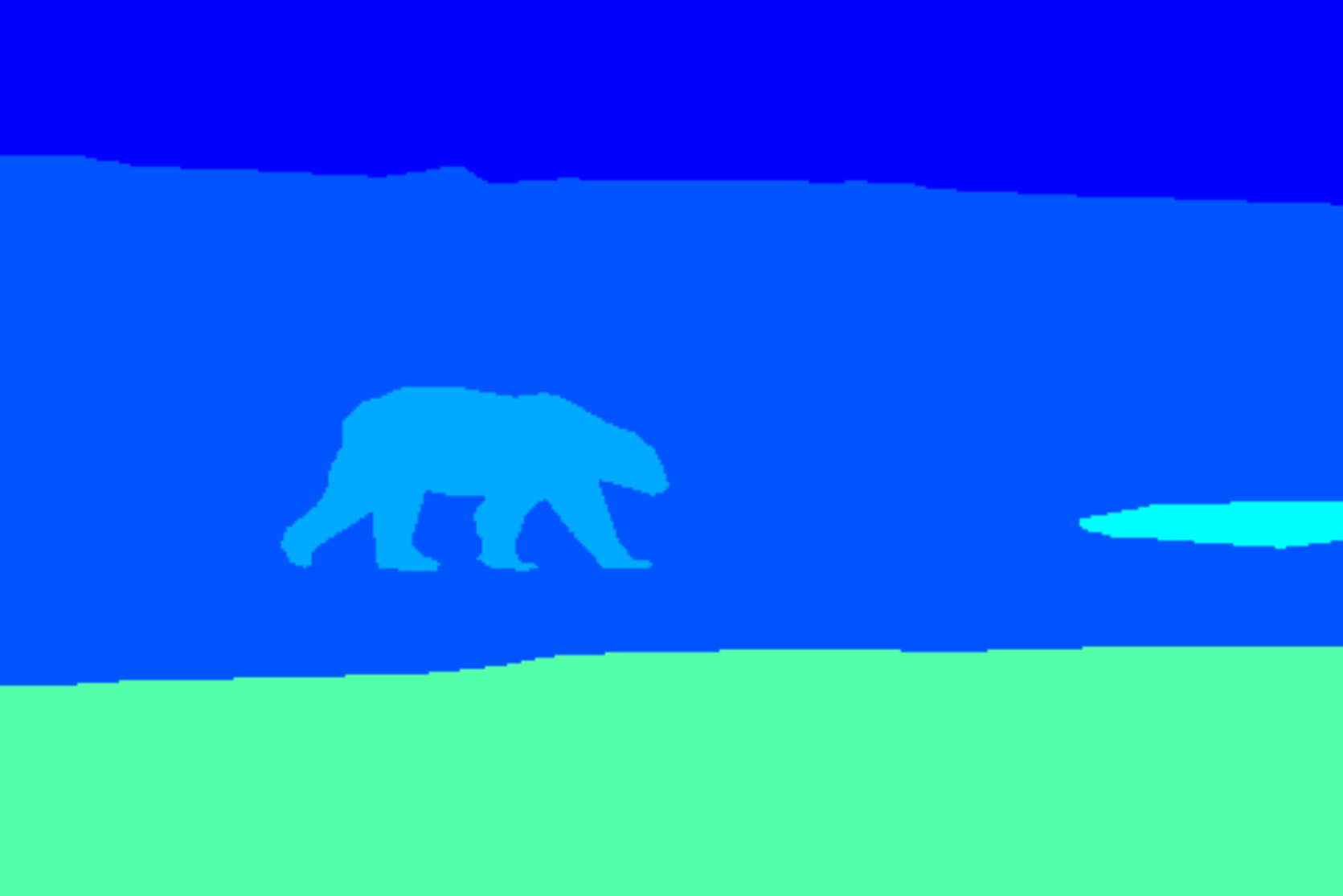}&
    \includegraphics[width=0.20\linewidth]{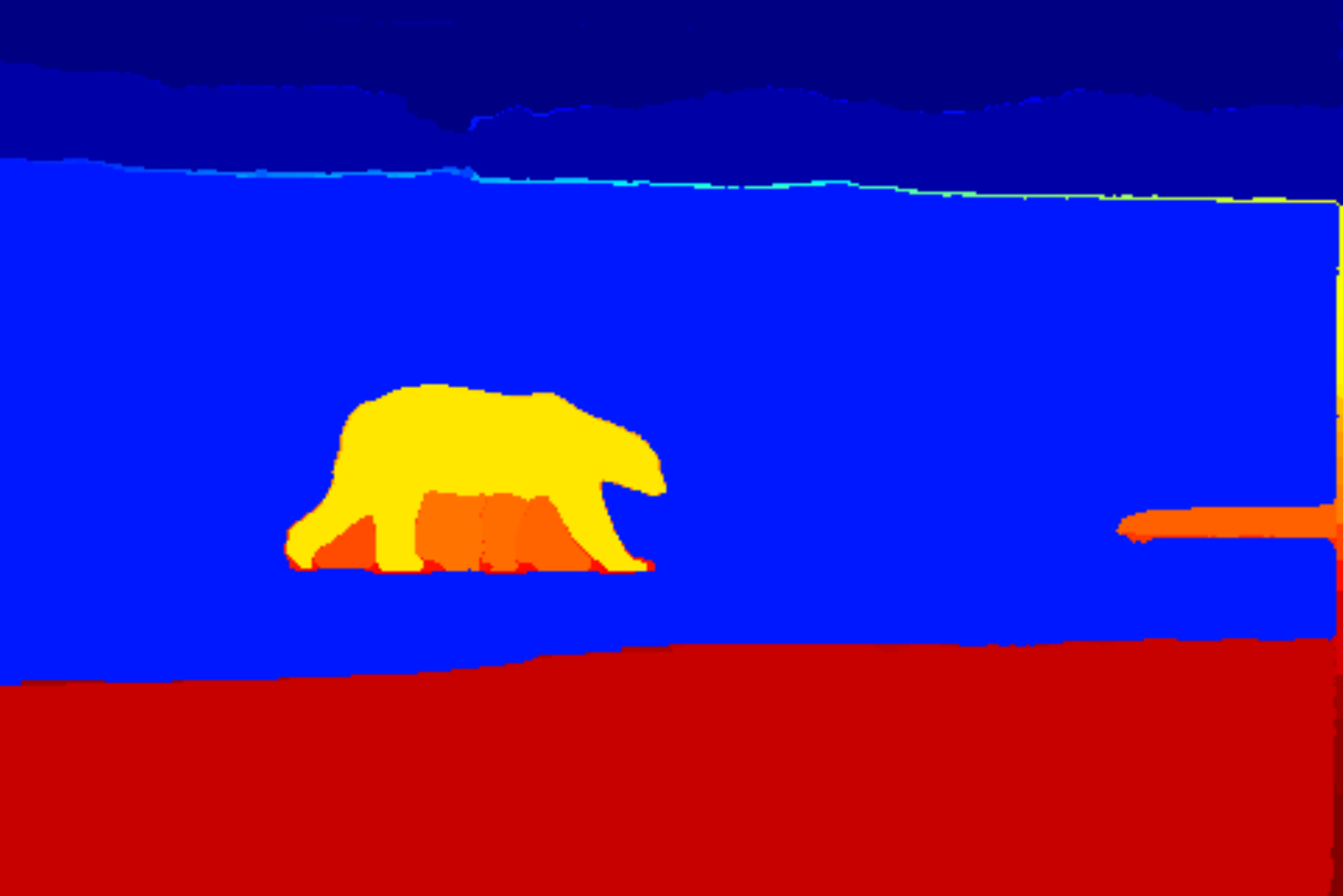}&
    \includegraphics[width=0.20\linewidth]{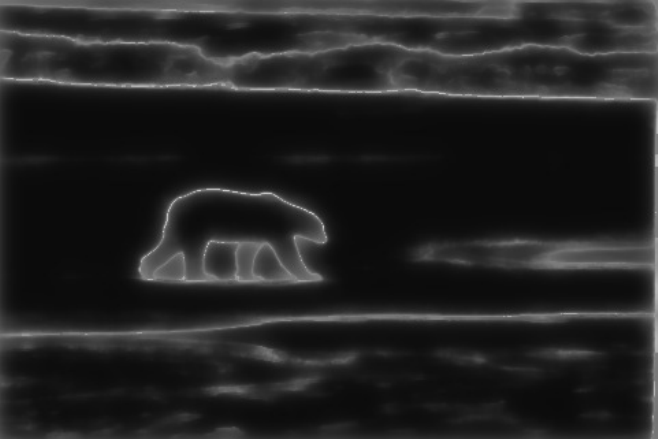}\\
    \includegraphics[width=0.20\linewidth]{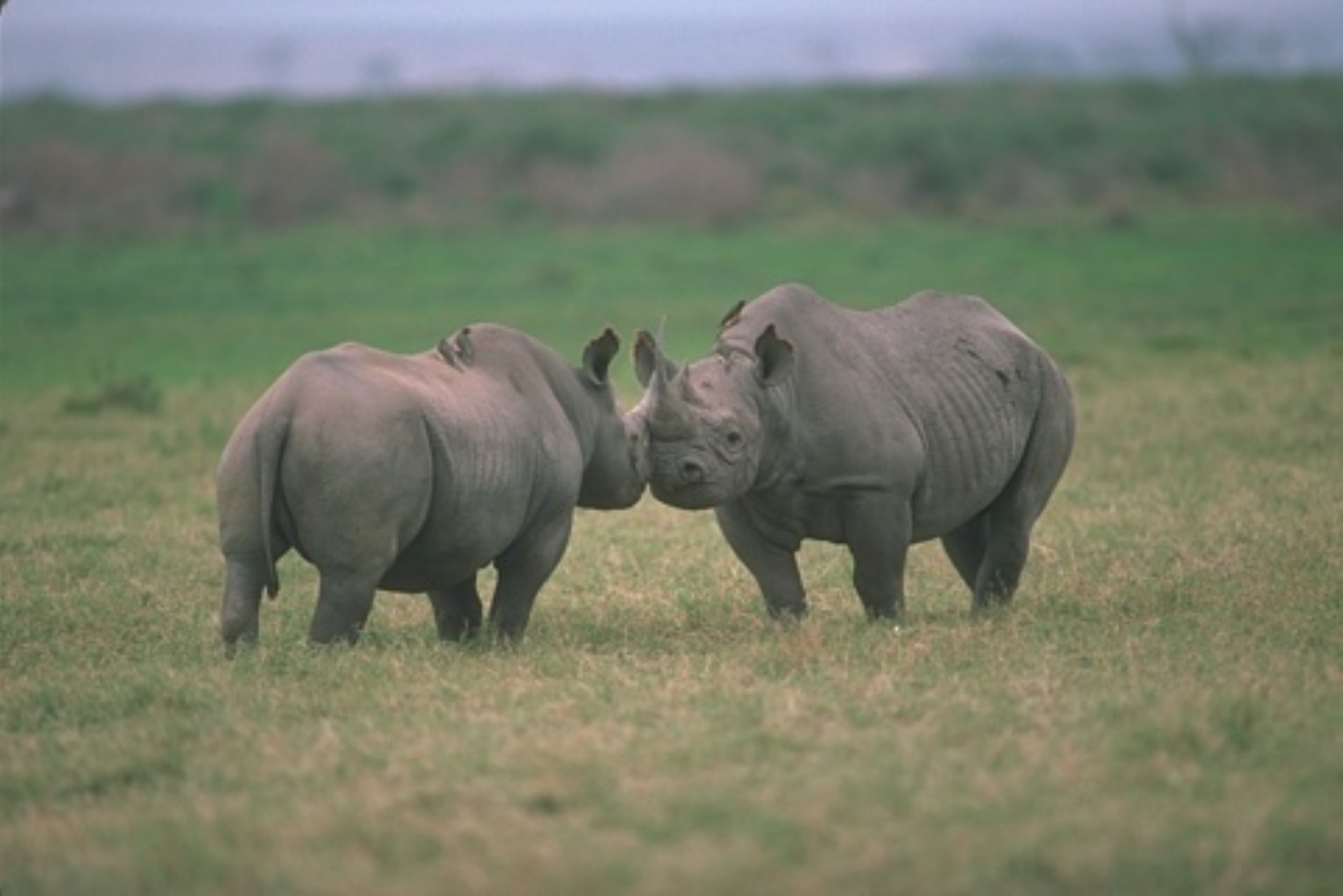}&
    \includegraphics[width=0.20\linewidth]{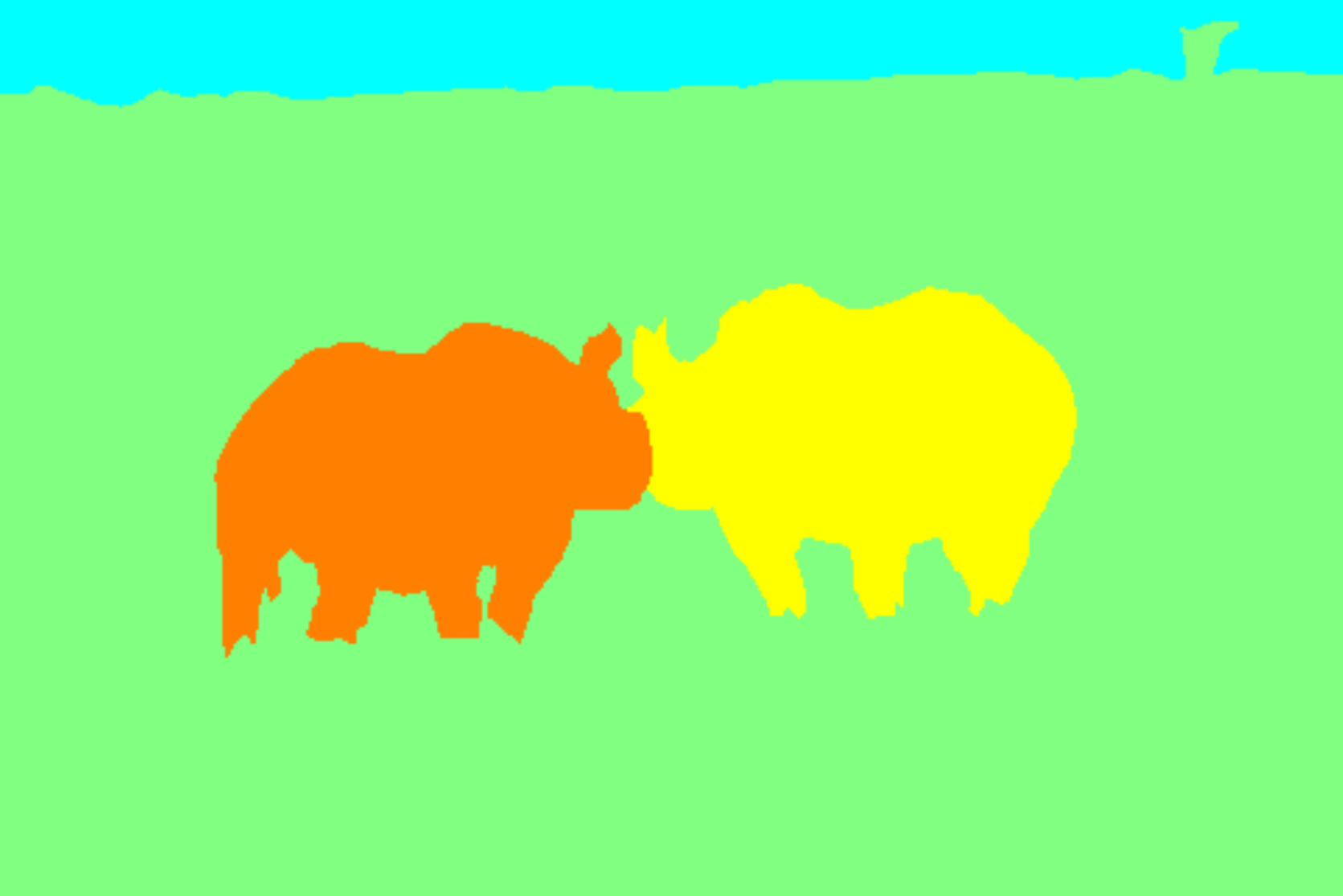}&
    \includegraphics[width=0.20\linewidth]{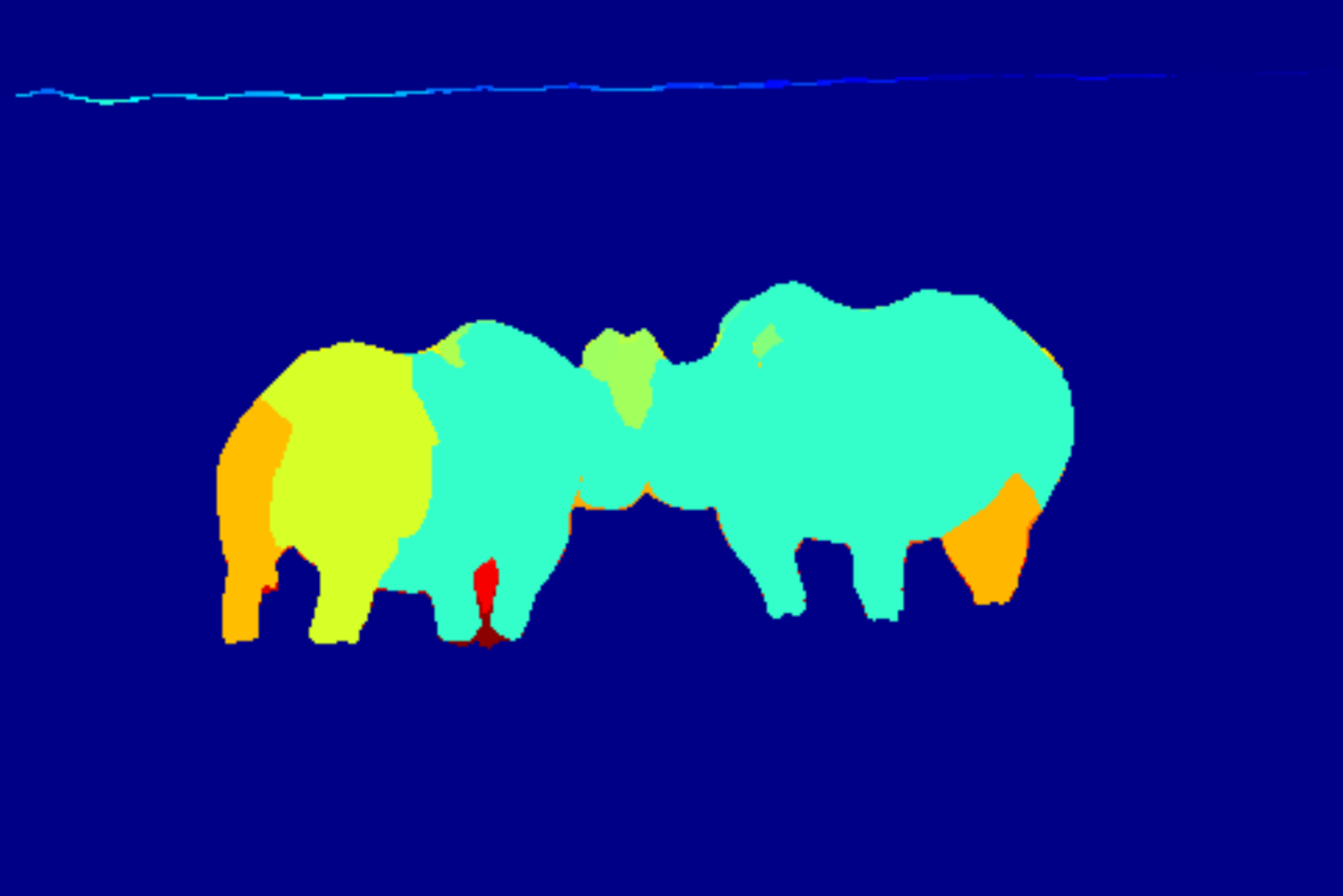}&
    \includegraphics[width=0.20\linewidth]{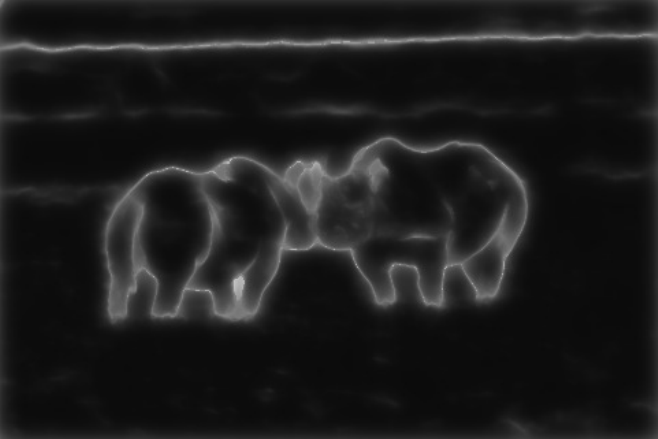}\\
    \includegraphics[width=0.20\linewidth]{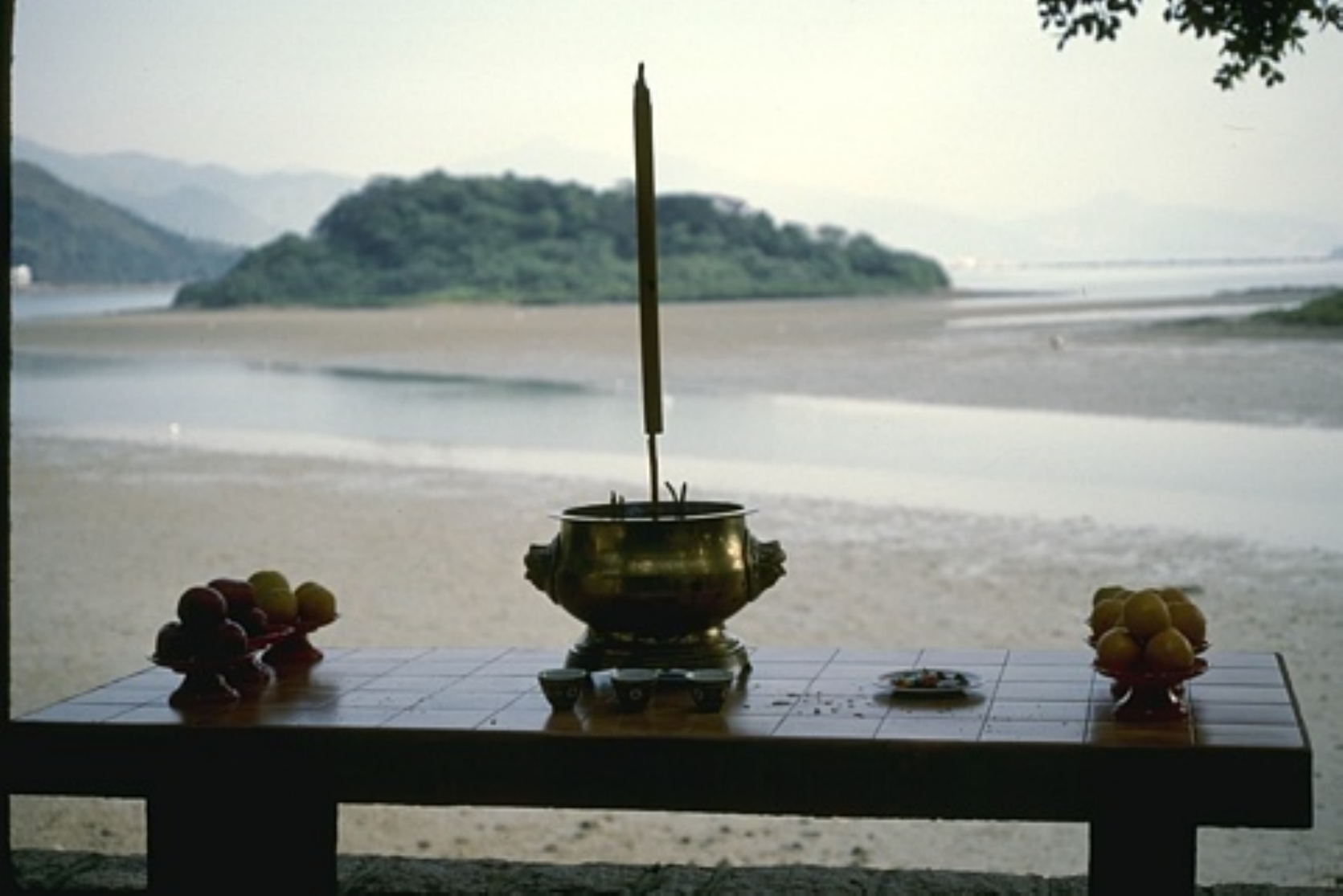}&
    \includegraphics[width=0.20\linewidth]{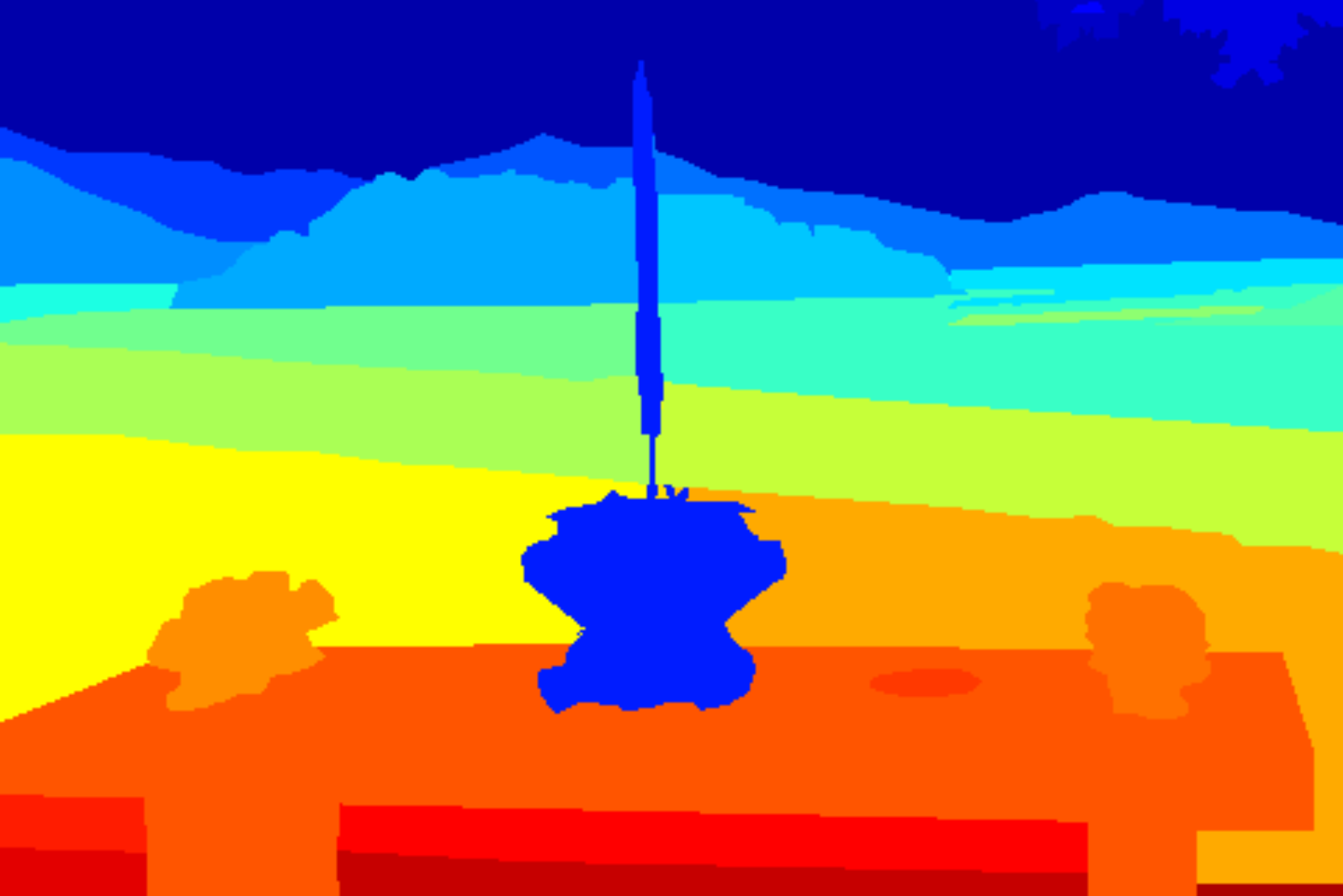}&
    \includegraphics[width=0.20\linewidth]{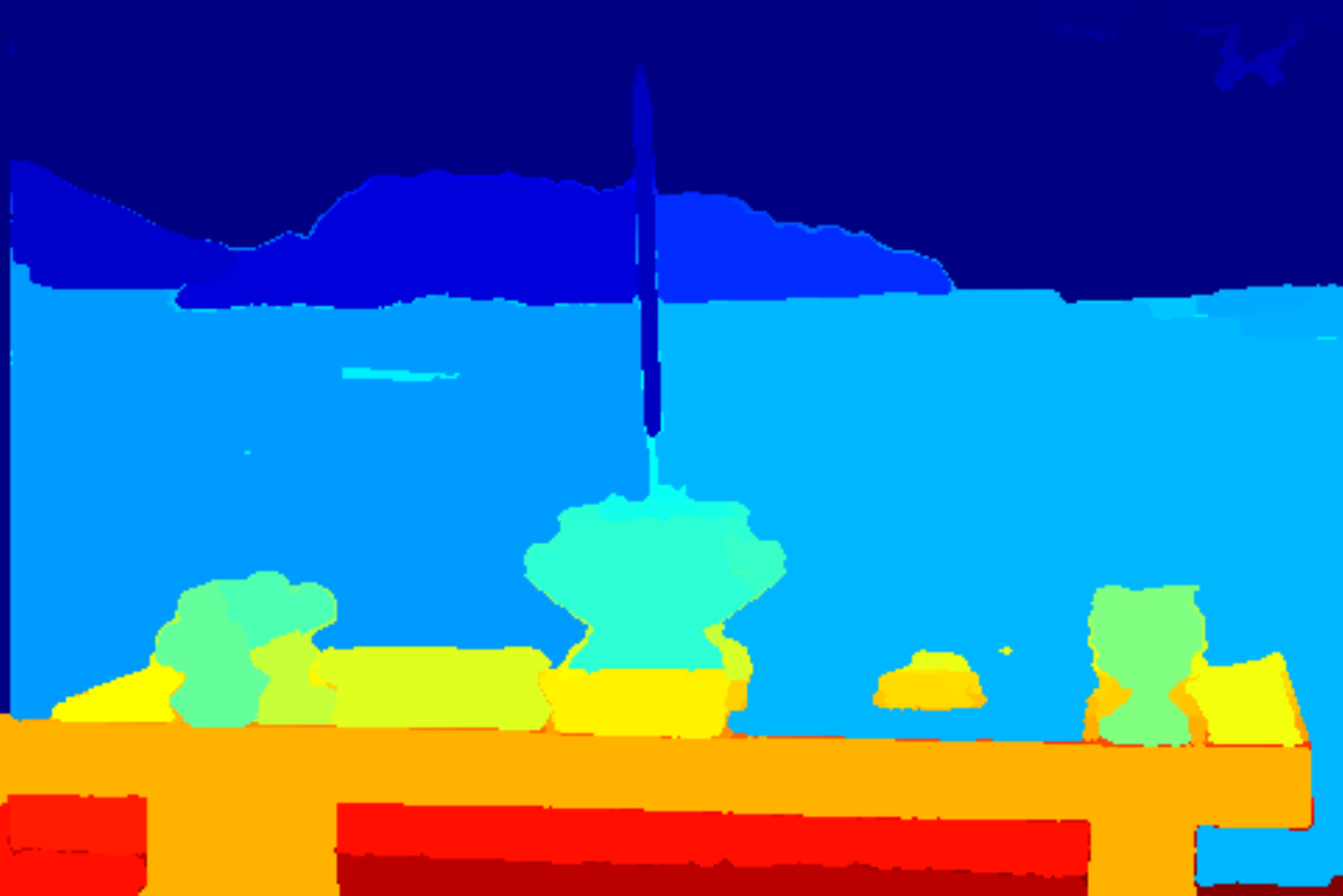}&
    \includegraphics[width=0.20\linewidth]{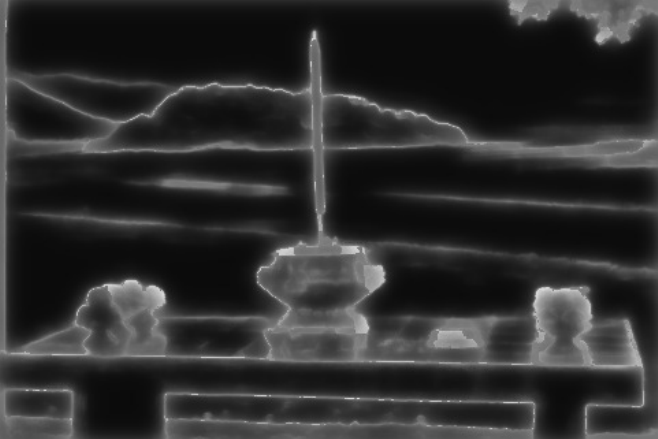}\\
    \includegraphics[width=0.20\linewidth]{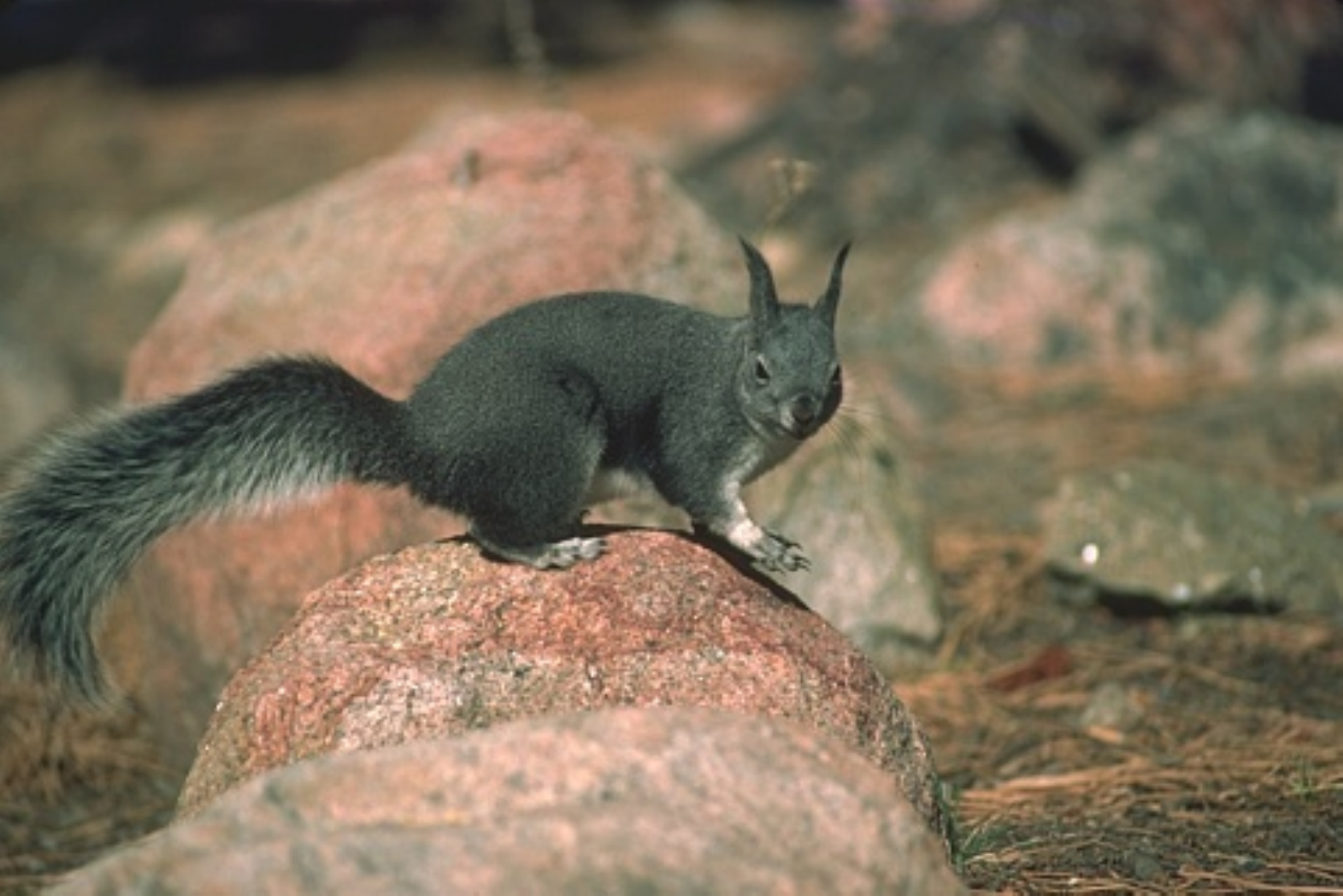}&
    \includegraphics[width=0.20\linewidth]{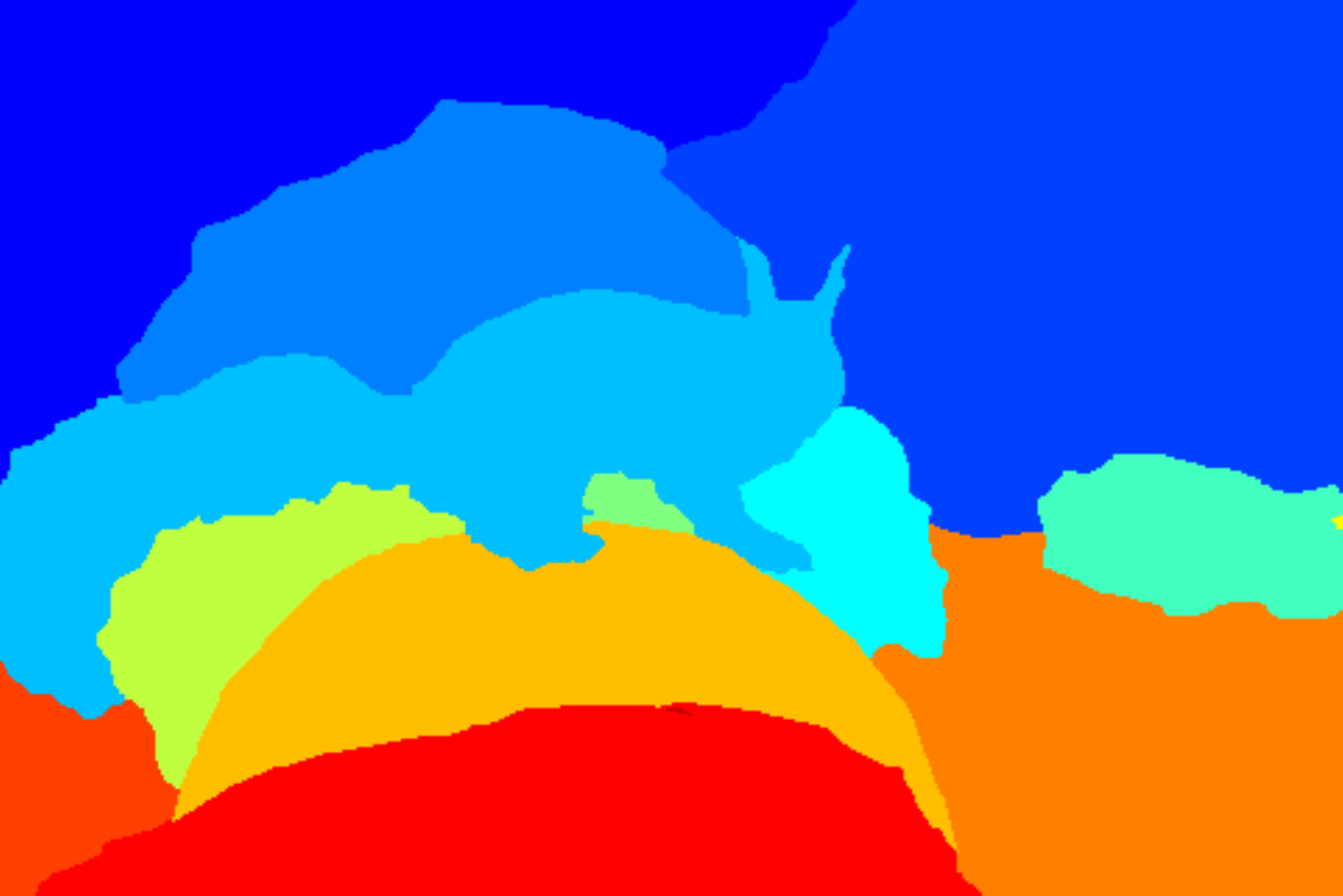}&
    \includegraphics[width=0.20\linewidth]{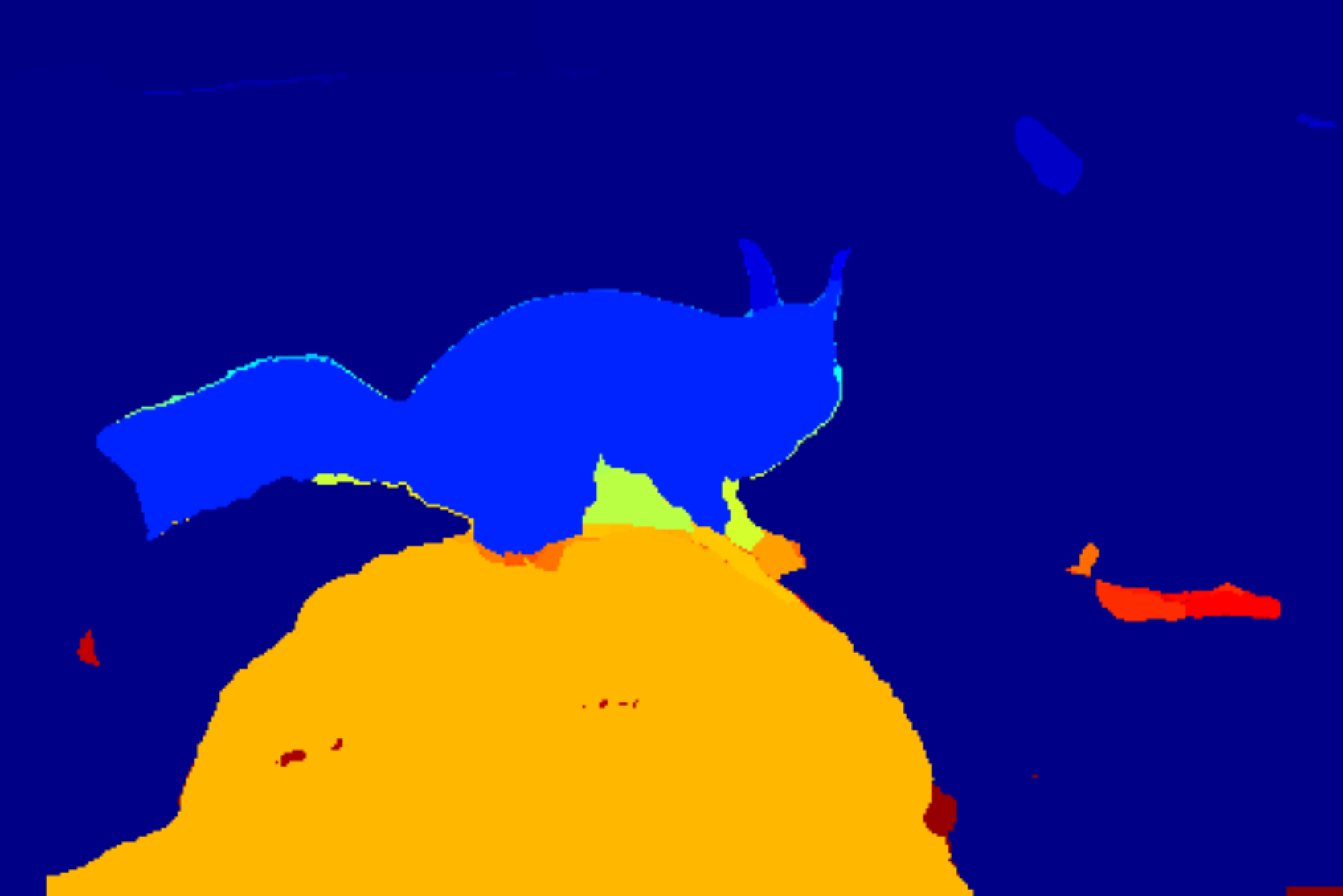}&
    \includegraphics[width=0.20\linewidth]{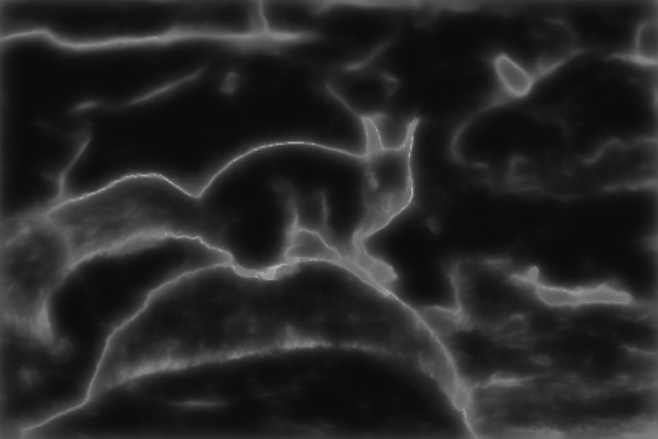}\\
    \end{tabular}
    \end{center}
    \caption{Exemplary images and segmentation uncertainties on the BSDS-500~\cite{BSDS500} dataset. In each row, from left to right, the original images, ground-truth segmentation, the resulting minimum cost lifted multicut segmentation and the proposed uncertainties are given. Bright areas in the uncertainty images represent uncertain pixels.}
    \label{fig:bsds_node}
\end{figure}

As a last application, we evaluate the proposed uncertainty measure on the image segmentation task in the context of minimum cost lifted multicuts. The evalutation is computed on the BSDS500 dataset using the problem instances and solutions from \cite{Keuper2015}.
Fig.~\ref{fig:bsds_node} visualizes the node (pixel) level uncertainty on several example images from the BSDS-500~\cite{BSDS500} dataset. As expected, the node level uncertainty is higher along the object boundaries, where label changes are likely to happen. 

\begin{figure}[t]
    \begin{center}
    \begin{tabular}{@{}ccc@{}}
    \includegraphics[width=0.3\linewidth]{1_resized_123057.jpg.pdf}&
    \includegraphics[width=0.3\linewidth]{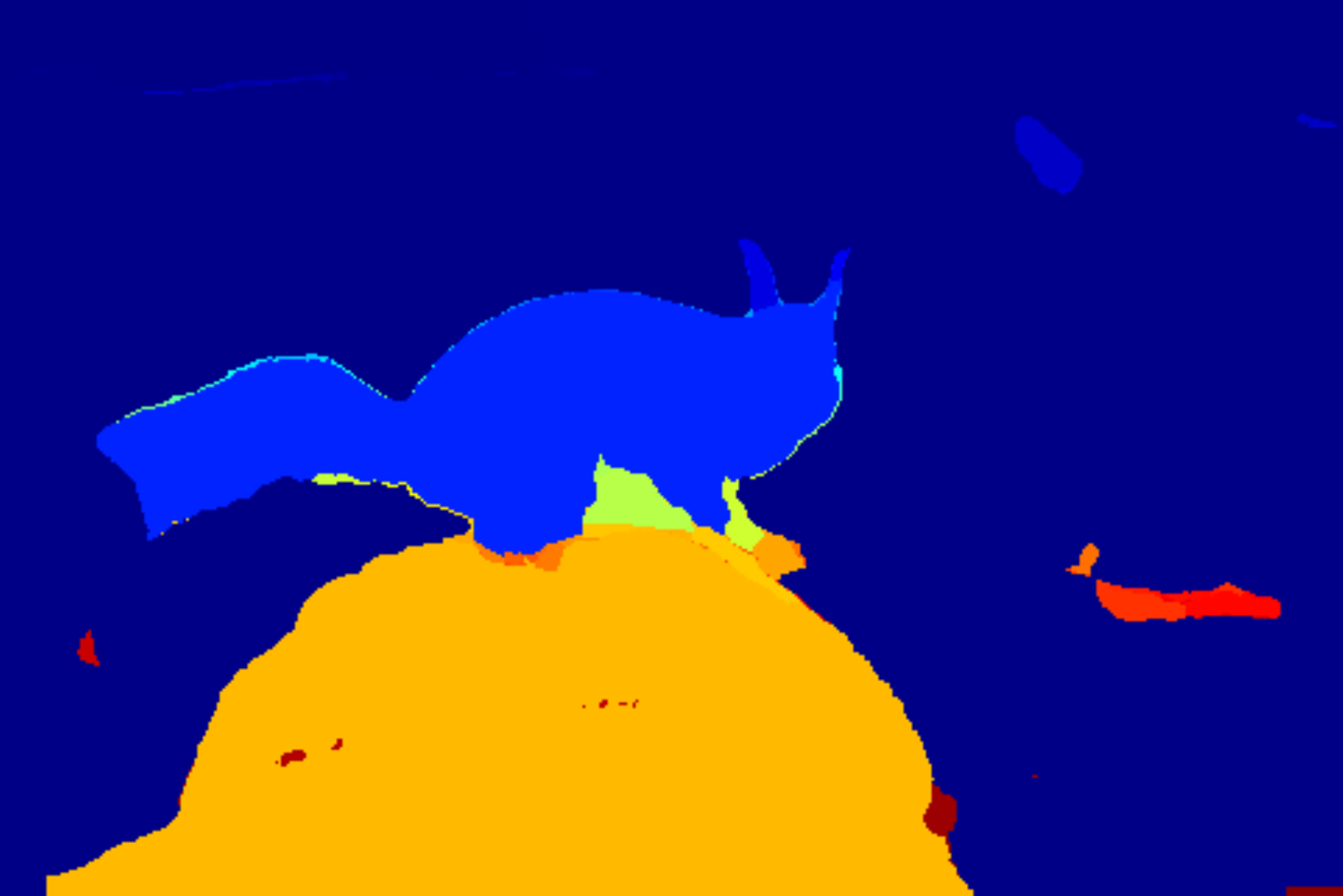}&
    \includegraphics[width=0.3\linewidth]{1_resized_123057.bmp.pdf}
    \\
    \includegraphics[width=0.3\linewidth]{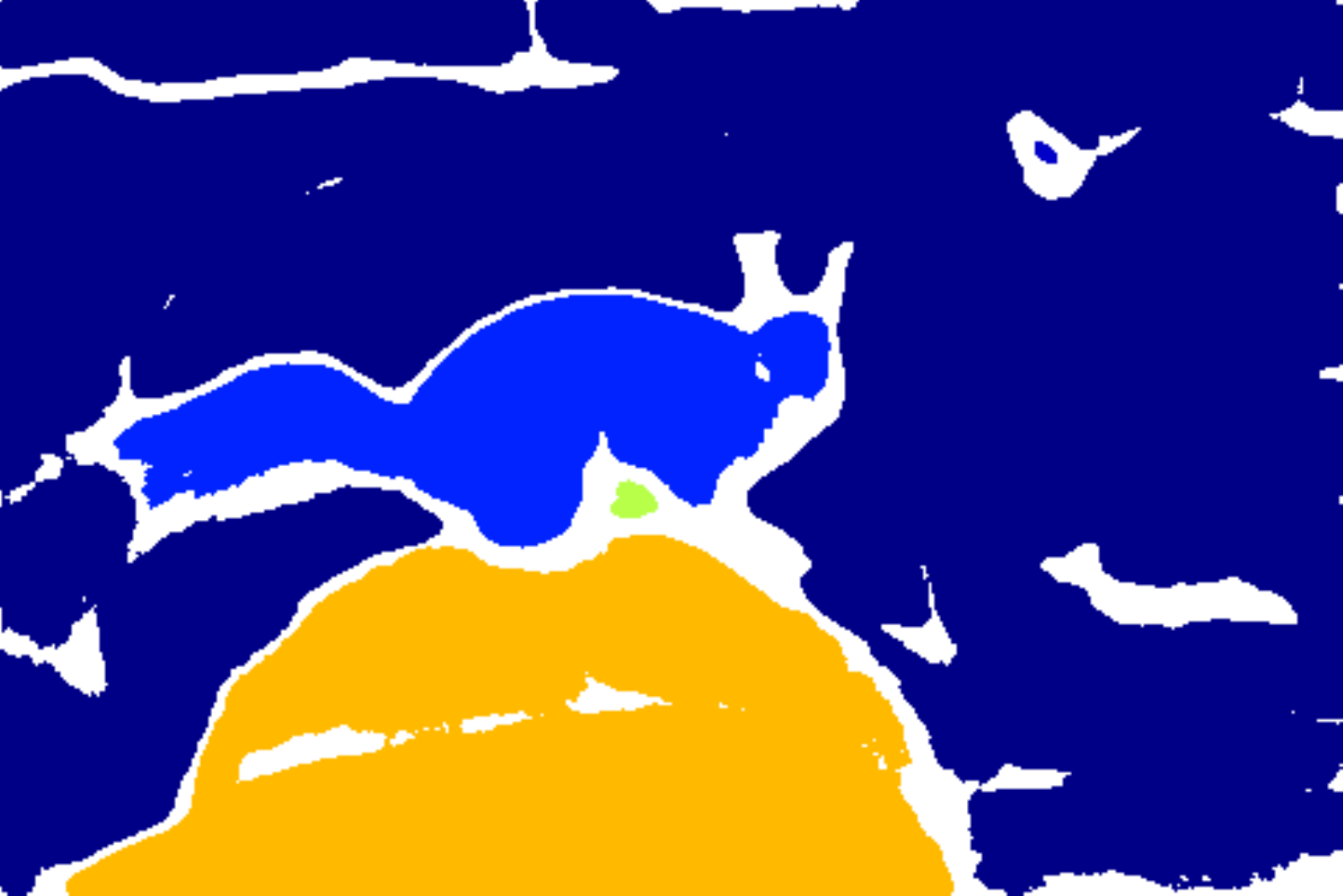}&
    \includegraphics[width=0.3\linewidth]{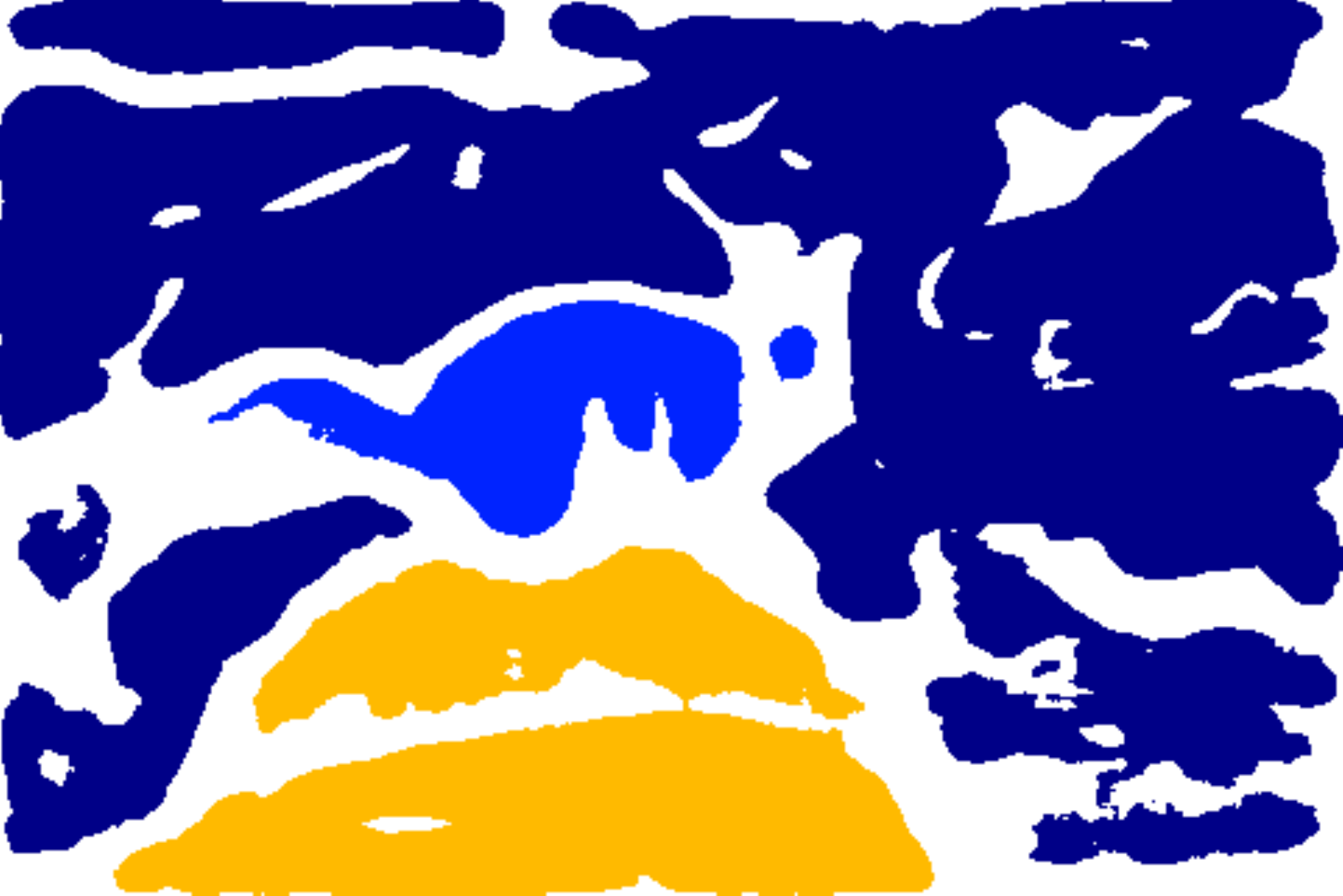}&
    \includegraphics[width=0.3\linewidth]{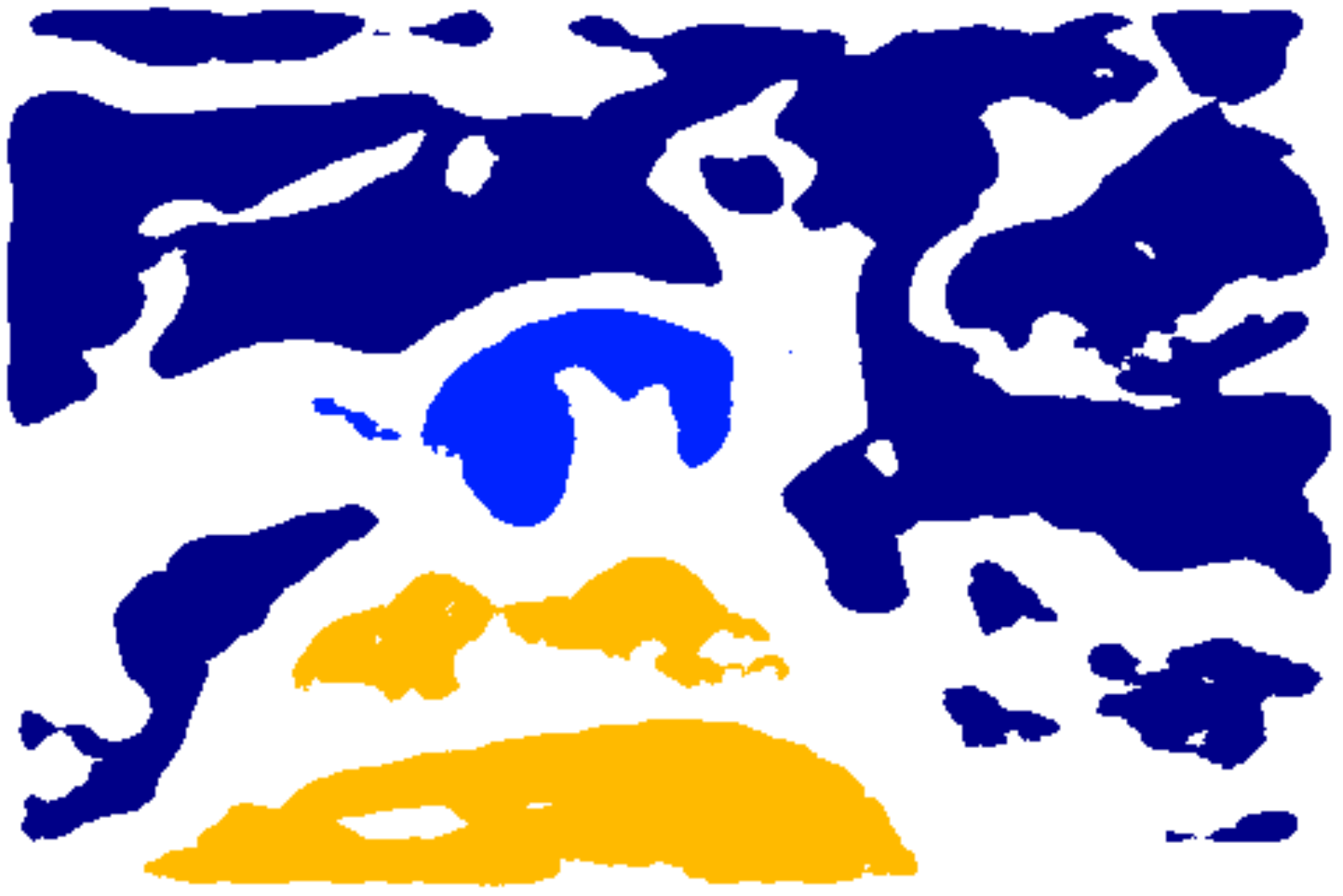}\\
    \end{tabular}
    \end{center}
    \caption{Visualization of removing uncertain pixels in BSDS-500~images (\cite{BSDS500}). Notice that removing uncertain pixels corresponds to removing pixels along the object boundaries. The original image (left), its multicut solution (middle) and the uncertainty measure (right) based on our model are shown in the first row. The second row visualizes (from left to right) removing 10, 30 and 50 \% most uncertain pixels.}
    \label{fig:bsds_node_remove}
\end{figure}
\begin{figure}[t]
    \begin{center}
    \small
    \begin{tabular}{@{}c@{}c@{}}
    \includegraphics[width=0.51\linewidth]{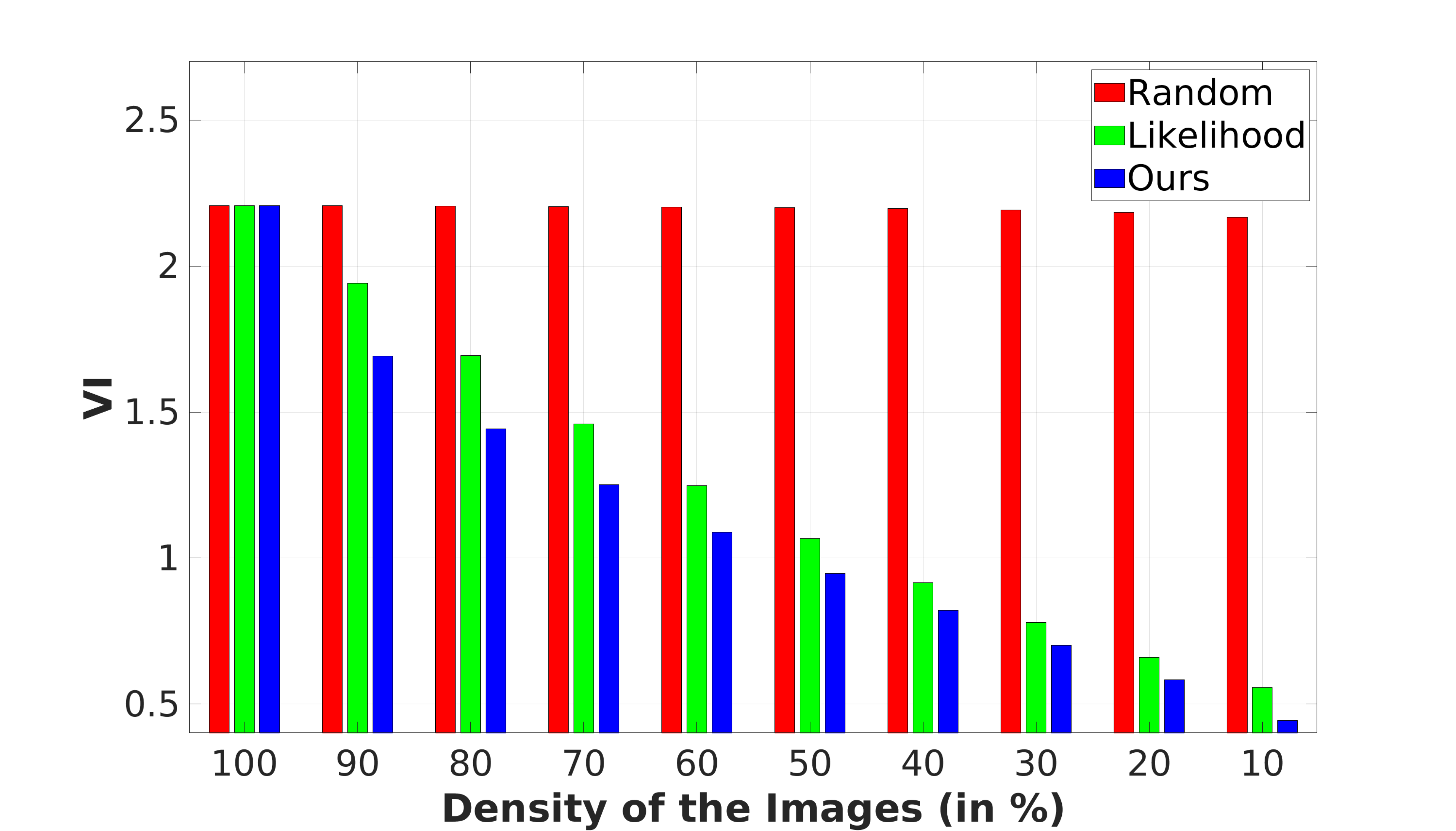}&
    \includegraphics[width=0.51\linewidth]{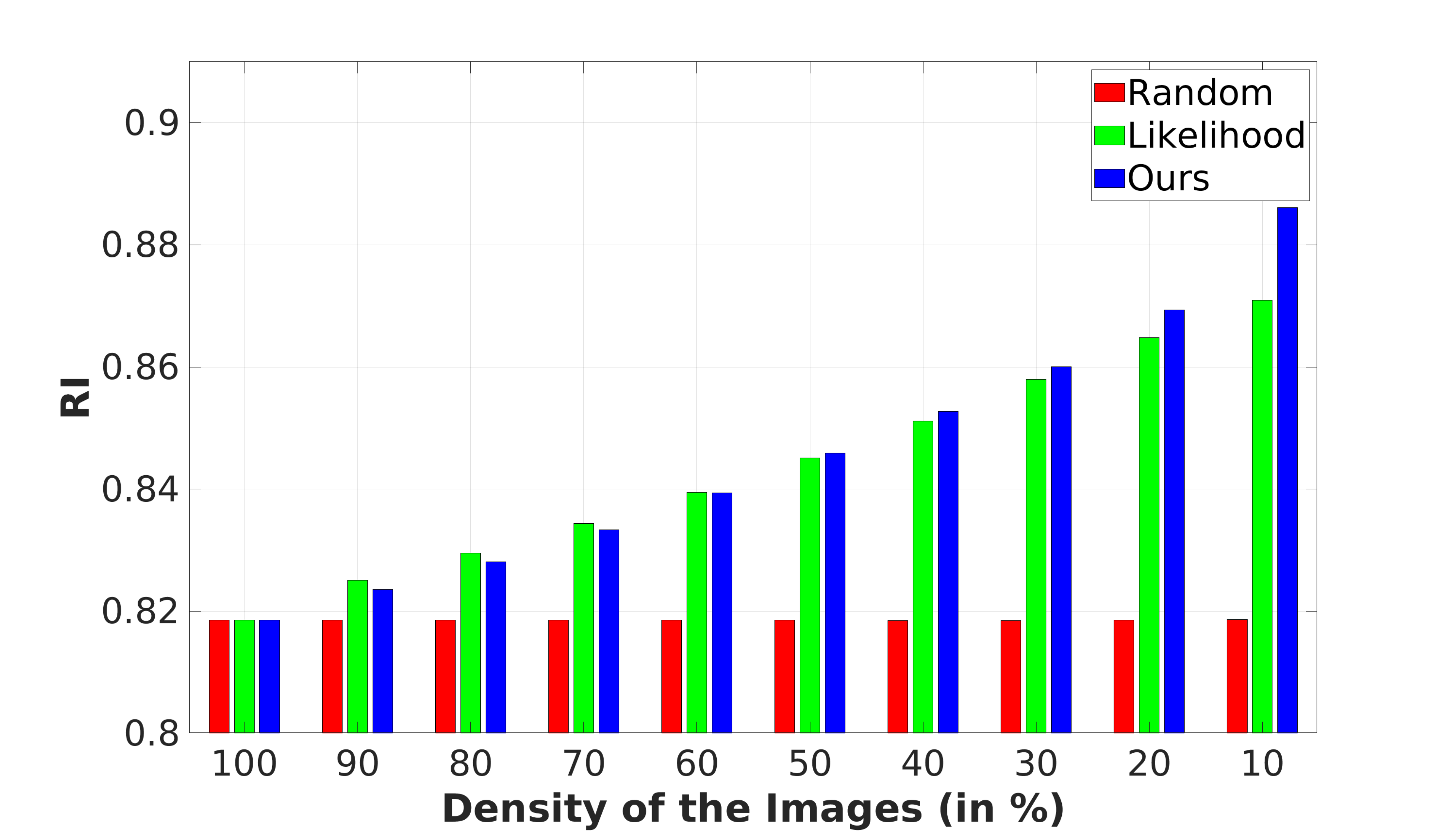}\\
    VI (lower is better)& RI (higher is better)
    \end{tabular}
    \end{center}
    \caption{Sparsification analysis in VI (\textit{left}) and RI (\textit{right}) on the BSDS-500~(\cite{BSDS500}) test data. The proposed method shows a faster decrease in VI than the baseline and reaches a higher RI.} 
    \label{fig:bsds_ri_vi_baselines}
\end{figure}

Removing the most uncertain pixels corresponds to removing object boundaries as it is depicted in Fig.~\ref{fig:bsds_node_remove} for an example. As the segmentation becomes sparser, entire object parts such as the tail of the squirrel will get removed from the solution. In Fig.~\ref{fig:bsds_ri_vi_baselines}, the VI and RI metrics are provided with respect to the pixel densities. As expected, removing highly uncertain pixels results in increasing the RI and decreasing VI. Thus, an improvement can be observed in both metrics. Again, we compare our to the probability driven baseline and to a random baseline. Unlike before, the probability driven baseline performs reasonably in this setting. Yet, the proposed measure can achieve a faster decrease in VI and a higher increase in RI and thus shows more robust behavior. The relatively good performance of the baseline in comparison to its results on motion segmentation can be explained by the way local cut probabilities are computed in this setting. Specifically, they are derived by \cite{Keuper2015} from edge maps and thus have consistently low values in homogeneous regions. Therefore, the normalization by the denominator in the proposed measure (refer to Eq.~\eqref{eq:uncertainty_eq4_1}) becomes less important.
	

\section{Conclusion}
\label{sec:conclusion}
The minimum cost \textit{lifted} multicut problem has been widely studied in different computer vision and data analysis applications such as motion segmentation and image decomposition. Due to the probabilistic behavior of the multicut problem the final decomposition of the nodes in the proposed graph can be seen as relaxed decisions rather than hard decisions on label assignment to the nodes of the graph. We studied this probabilistic behavior and provide an informative uncertainty measure in the node uncertainties. We showed that removing uncertain nodes according to the proposed measure leads to an improvement in the variation of information (VI) and the Rand index (RI) in two applications of motion segmentation and image decomposition. Further, we showed the application of such uncertainties to train a self-supervised model for motion segmentation. The proposed uncertainty measure can be combined with any minimum cost multicut based formulation. 

\subsubsection*{Acknowledgment}
We acknowledge funding by the DFG project KE 2264/1-1 and thank the NVIDIA Corporation for the donation of a Titan Xp GPU.

\bibliography{uai2021-template}

\clearpage

\title{Supplementary Material:\\
``Uncertainty in Minimum Cost Multicuts for Image and Motion Segmentation''}

\maketitle
\vspace*{-1cm}
  In this supplementary document, we first provide a detailed derivation of the proposed uncertainty measure in terms of the underlying local cut probabilities. Then, we provide additional evaluations of the uncertainties in the context of minimum cost multicuts for motion segmentation when the GAEC heuristic (Keuper et al. [2015b]) is applied as a solver.
\section{Uncertainty Estimation}

Given an instance of the (lifted) multicut problem and its solution, we employ the probability measures in equations (6) (\textit{MP}) and (7) \textit{(LMP)} of the main paper. We iterate through nodes $v_i\in\{1, \dots, |V|\}$ in vicinity of a cut, i.e. $\exists e\in \pazocal{N}_{E}(v_i)$ with $e\in E$ and $y_e=1$.
Assuming that $v_i$ belongs to segment $A$ and its neighbour $v_j$ according to $E$ belongs to the segment $B$, the amount of cost change $\gamma_B$ is computed in the linear cost function (defined in equations (6) and (7) of the main paper) by moving $v_i$ from cluster $A$ to cluster $B$ as
%
\begin{align}
\gamma_B=\sum_{v_j\in\pazocal{N}_{E'}(v_i)\cap A}c_{(v_i,v_j)} - \sum_{v_j\in\pazocal{N}_{E'}(v_i)\cap B}c_{(v_i,v_j)}.
\label{eq:costDifference}
\end{align}
Thus, in $\gamma_{B}$, we accumulate all costs of edges from $v_i$ that are not cut in the current decomposition and subtract all costs of edges that are cut in the current decomposition but would not be cut if $v_i$ is moved from $A$ to $B$. 
Note that, while the cost change is computed over all edges in $E'$ for lifted graphs, only the uncertainty of nodes with an adjacent cut edge in $E$ can be considered in order to preserve the feasibility of the solution. For each node $v_i$ the number of possible moves depends on the labels of its neighbours $\pazocal{N}_{E}(v_i)$, and Eq.~\eqref{eq:costDifference} allows us to assign a cost to any such node-label change. Altogether, we assess the uncertainty of a given node label by the cheapest, i.e. the most likely, possible move
\begin{align}
\gamma_i = \min_B \gamma_B.
\label{eq:uncertain_node}
\end{align}
and set $\gamma_i$ to $\infty$ if no move is possible. 
The minimization in Eq.~\eqref{eq:uncertain_node} corresponds to considering the local move of $v_i$ which maximizes
\begin{align}
\prod_{e=(v_i,v_j), v_j\in A}
\frac{p_{\pazocal{Y}_e | \pazocal{X}_e}(1, x_e)}{p_{\pazocal{Y}_e | \pazocal{X}_e}(0, x_e)} \cdot\prod_{e=(v_i,v_j), v_j\in B}\frac{p_{\pazocal{Y}_e | \pazocal{X}_e}(0, x_e)}{p_{\pazocal{Y}_e | \pazocal{X}_e}(1, x_e)}.
\label{eq:uncertain_nodeProb}
\end{align}
To produce an uncertainty measure for each node in the graph, we apply the logistic function on \eqref{eq:uncertain_node} 
\begin{align}
\mathrm{uncertainty=\frac{1}{1+\exp{(-\gamma_i})}} 
\label{eq:uncertainty_eq1}
\end{align}
as it is the inverse of the logit function used in the cost computation in \eqref{eq:weights-map}.

\begin{figure}[t]
\begin{center}
    \includegraphics[width=50mm]{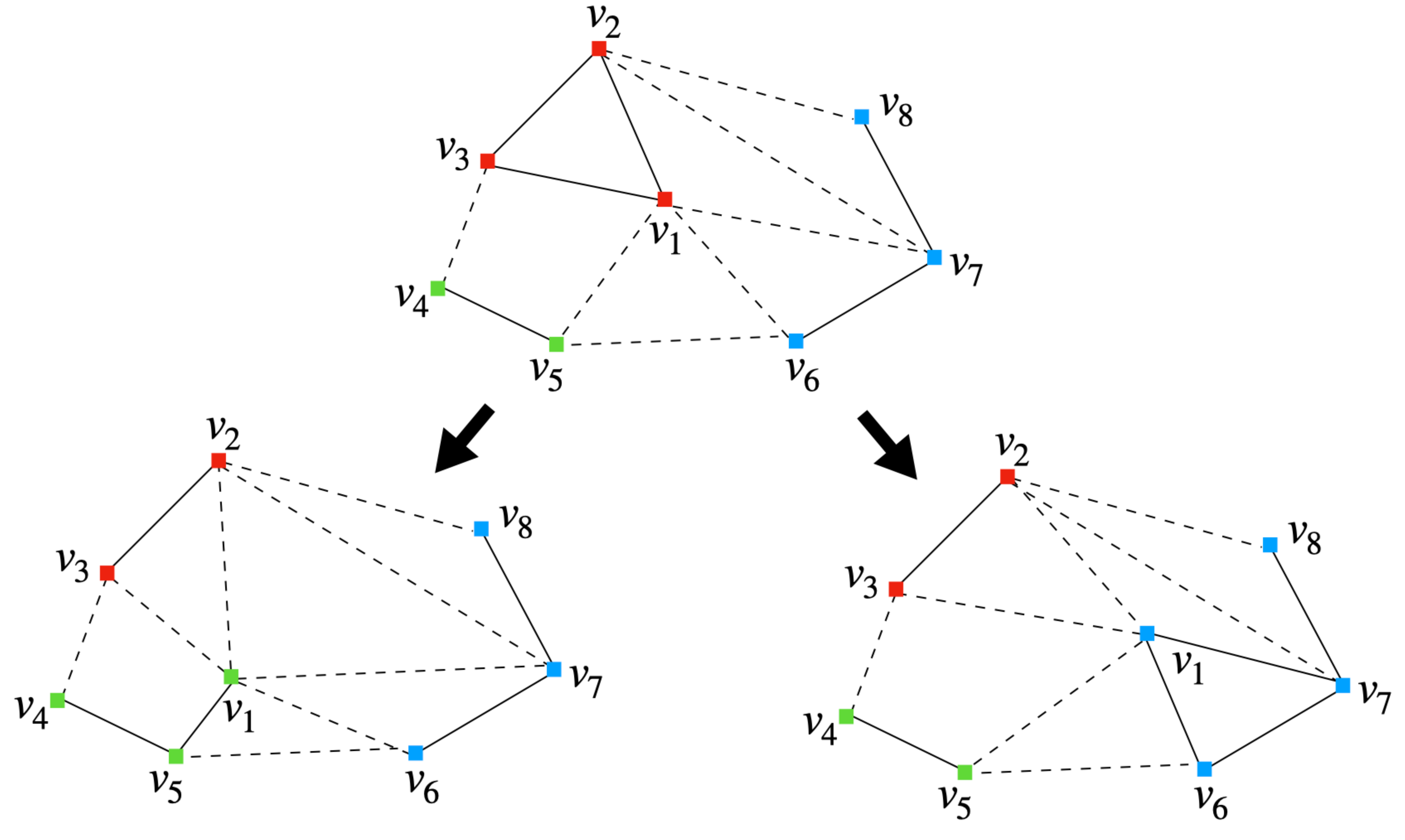}
\end{center}
\caption{In the current decomposition of the exemplary graph $G=(V,E)$ (top figure), we study the node uncertainties as represented in equation \eqref{eq:costDifference}. For instance, $v_1$ is moved from one partition (red label) to the new possible partitions (blue and green labels) and the cost change is estimated. The $\gamma_\alpha$ represents the cost which minimize the cost among these moves.}
\label{fig:graph_fig}
\end{figure}

In the following we show the expansion of equation \eqref{eq:uncertainty_eq1}, by injecting the corresponding values for $\gamma_i$.
The determine $\gamma_i$, the minimum change in the cost is computed among all the possible changes between the clusters for each node $v_i \in V$ (refer to Fig. \ref{fig:graph_fig} for an example), such that

\begin{align}
\mathrm{uncertainty}=\frac{1}{1+\exp{(- \displaystyle \min_B \gamma_B)}}
\label{eq:uncertainty_eq2}
\end{align}
where
\begin{align}
\gamma_B=\sum_{v_j\in\pazocal{N}_{E'}(v_i)\cap A}c_{(v_i,v_j)} - \sum_{v_j\in\pazocal{N}_{E'}(v_i)\cap B}c_{(v_i,v_j)}.
\label{eq:costDifference}
\end{align}

The resulting uncertainty measure is thus of the form 
\begin{align}
&\mathrm{uncertainty}&\nonumber\\
%
&= \frac{1}{1+\exp{\left(\displaystyle \sum_{v_j\in\pazocal{N}_{E'}(v_i)\cap B}c_{(v_i,v_j)} -  \displaystyle \sum_{v_j\in\pazocal{N}_{E'}(v_i)\cap A}c_{(v_i,v_j)}\right)}}.
\label{eq:uncertainty_eq3}
\end{align}

According to the Bayesian model and the findings in (Andres
et al. [2012]), the costs $c_{(v_i,v_j)}$ for each $e:= (v_i,v_j)\in E$ are computed via (refer to the equation (9) in the main paper),

\begin{align}
\forall e \in E: \quad
c_e = \log \frac{p_{\pazocal{Y}_e | \pazocal{X}_e}(0, x_e)}{p_{\pazocal{Y}_e | \pazocal{X}_e}(1, x_e)} + \log \frac{1-\beta}{\beta}
\enspace .
\label{eq:weights-map}
\end{align}

For simplicity and to be compatible with our experiments, we set the value of $\beta = 0.5$ (i.e. we assume an unbiased decomposition), which makes $\log \frac{1-\beta}{\beta} = 0$.

We insert $c(v_i, v_j) = \log \frac{p_{\pazocal{Y}_e | \pazocal{X}_e}(0, x_e)}{p_{\pazocal{Y}_e | \pazocal{X}_e}(1, x_e)}$, ($e=(v_i, v_j)$) into Eq. \eqref{eq:uncertainty_eq3}  and denote
$v_j\in\pazocal{N}_{E'}(v_i)\cap B$ by $e,B$ where $e=(v_i,v_j), v_j \in B$ and $v_j\in\pazocal{N}_{E'}(v_i)\cap A$ by $e,A$ where $e=(v_i,v_j), v_j \in A$, and get

\begin{align}
&\mathrm{uncertainty} \nonumber\\
&=\frac{1}{1+\exp{\left(\displaystyle \sum_{e,B}\log \frac{p_{\pazocal{Y}_e | \pazocal{X}_e}(0, x_e)}{p_{\pazocal{Y}_e | \pazocal{X}_e}(1, x_e)} - \displaystyle \sum_{e,A}\log \frac{p_{\pazocal{Y}_e | \pazocal{X}_e}(0, x_e)}{p_{\pazocal{Y}_e | \pazocal{X}_e}(1, x_e)}\right)}}\nonumber
\label{eq:uncertainty_eq3_1}
\end{align}

\begin{align}
& = \frac{1}{1+\exp{\left(\displaystyle \log \prod_{e,B} \frac{p_{\pazocal{Y}_e | \pazocal{X}_e}(0, x_e)}{p_{\pazocal{Y}_e | \pazocal{X}_e}(1, x_e)} - \displaystyle \log \prod_{e,A} \frac{p_{\pazocal{Y}_e | \pazocal{X}_e}(0, x_e)}{p_{\pazocal{Y}_e | \pazocal{X}_e}(1, x_e)}\right)}}\nonumber
\end{align}

\begin{align}
&= \frac{1}{1+\exp{\left(\displaystyle \log \prod_{e,B} \frac{p_{\pazocal{Y}_e | \pazocal{X}_e}(0, x_e)}{p_{\pazocal{Y}_e | \pazocal{X}_e}(1, x_e)} + \displaystyle \log \prod_{e,A} \frac{p_{\pazocal{Y}_e | \pazocal{X}_e}(1, x_e)}{  p_{\pazocal{Y}_e | \pazocal{X}_e}(0, x_e)}\right)}}&\nonumber
\end{align}

\begin{align}
& = \frac{1}{1+\exp{\left(\displaystyle \log \left(\prod_{e,B} \frac{p_{\pazocal{Y}_e | \pazocal{X}_e}(0, x_e)}{p_{\pazocal{Y}_e | \pazocal{X}_e}(1, x_e)} . \displaystyle \prod_{e,A} \frac{p_{\pazocal{Y}_e | \pazocal{X}_e}(1, x_e)}{  p_{\pazocal{Y}_e | \pazocal{X}_e}(0, x_e)}\right)\right)}}\nonumber\\
&= \frac{1}{1+{\left(\displaystyle \prod_{e,B} \frac{p_{\pazocal{Y}_e | \pazocal{X}_e}(0, x_e)}{p_{\pazocal{Y}_e | \pazocal{X}_e}(1, x_e)} . \displaystyle \prod_{e,A} \frac{p_{\pazocal{Y}_e | \pazocal{X}_e}(1, x_e)}{  p_{\pazocal{Y}_e | \pazocal{X}_e}(0, x_e)}\right)}}\nonumber
\\
& = \frac{1}{1+{\left(\frac{\displaystyle \prod_{e,B} p_{\pazocal{Y}_e | \pazocal{X}_e}(0, x_e)}{\displaystyle \prod_{e,B} p_{\pazocal{Y}_e | \pazocal{X}_e}(1, x_e)} . \frac{\displaystyle \prod_{e,A} p_{\pazocal{Y}_e | \pazocal{X}_e}(1, x_e)}{\displaystyle \prod_{e,A} p_{\pazocal{Y}_e | \pazocal{X}_e}(0, x_e)}\right)}}
\end{align}

Note that in the denominator, we have exactly 1 + the term from Equation (12) in the main paper. With a slight reformulation, we get
\fontsize{7}{6}{
\begin{align}
&
=\frac{\displaystyle \prod_{e,A} p_{\pazocal{Y}_e | \pazocal{X}_e}(0, x_e) . \displaystyle \prod_{e,B} p_{\pazocal{Y}_e | \pazocal{X}_e}(1, x_e)}{{\displaystyle \prod_{e,A} p_{\pazocal{Y}_e | \pazocal{X}_e}(0, x_e) . \displaystyle \prod_{e,B} p_{\pazocal{Y}_e | \pazocal{X}_e}(1, x_e)+{\displaystyle \prod_{e,B} p_{\pazocal{Y}_e | \pazocal{X}_e}(0, x_e) . \displaystyle \prod_{e,A} p_{\pazocal{Y}_e | \pazocal{X}_e}(1, x_e)}}}
\end{align}
}%
or by a simplified notation $p_{\pazocal{Y}_e | \pazocal{X}_e}(1, x_e)=: p_c$ for $cut$ probabilities and $p_{\pazocal{Y}_e | \pazocal{X}_e}(0, x_e)=: p_j$ for \emph{join} probabilities. 
\begin{figure}[t!]
    \begin{center}
    \small
    \begin{tabular}{@{}c@{}c@{}}
    \includegraphics[width=0.49\linewidth]{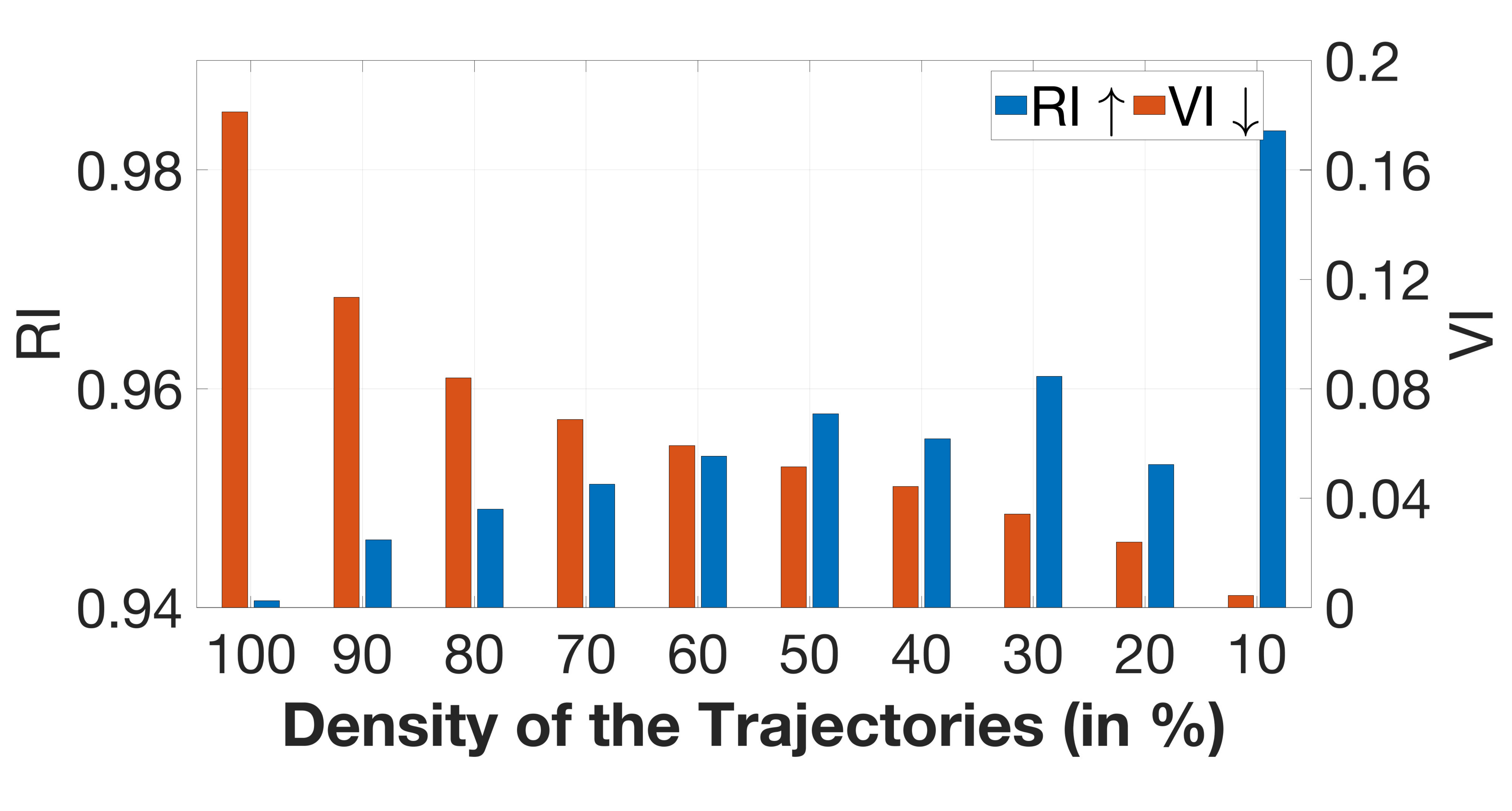}&
    \includegraphics[width=0.49\linewidth]{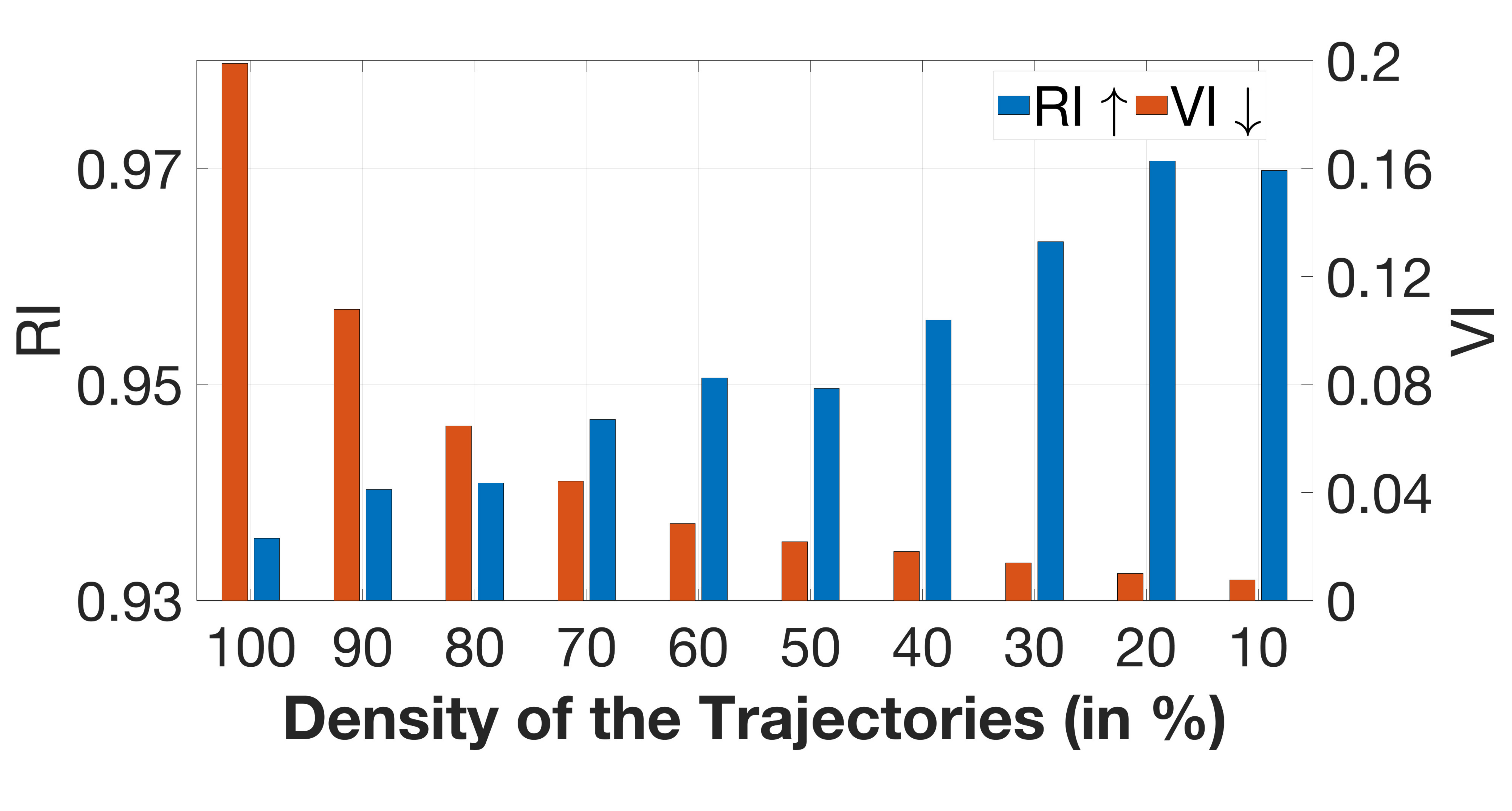}
    \end{tabular}
    \end{center}
    \caption{Study on the trajectory uncertainty on the GAEC~(Keuper et al. [2015b]) solver. The experiment relates to the Variation of Information (VI) and Rand Index (RI) on the train (left) and test (right) set of FBMS$_{59}$~(9 Ochs
et al. [2014]).}
    \label{fig:fbms_ri_vi_gaec}
\end{figure}
\begin{figure}[t!]
    \begin{center}
    \small
    \begin{tabular}{@{}c@{}c@{}}
    \includegraphics[width=0.49\linewidth]{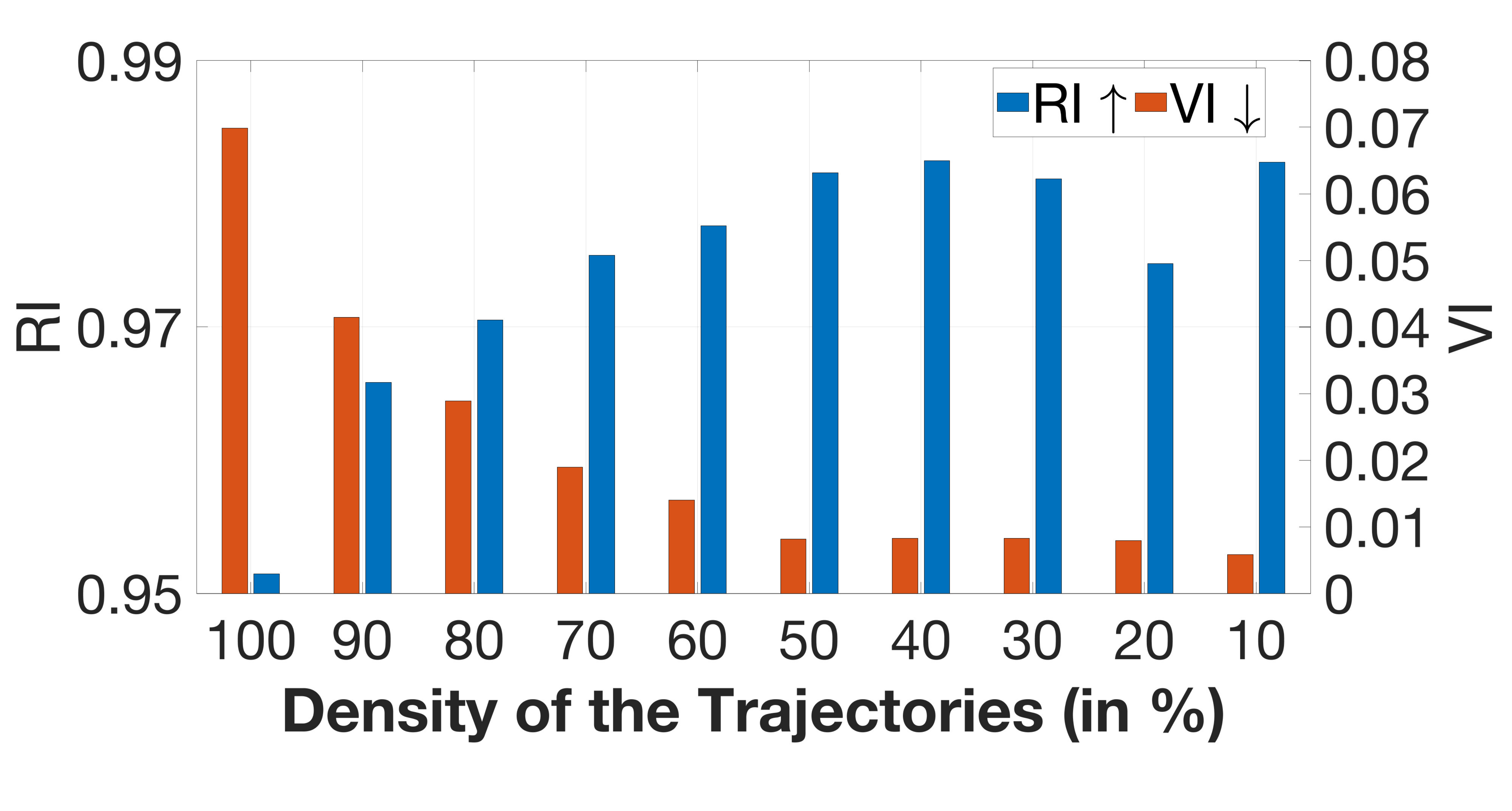}&
    \includegraphics[width=0.49\linewidth]{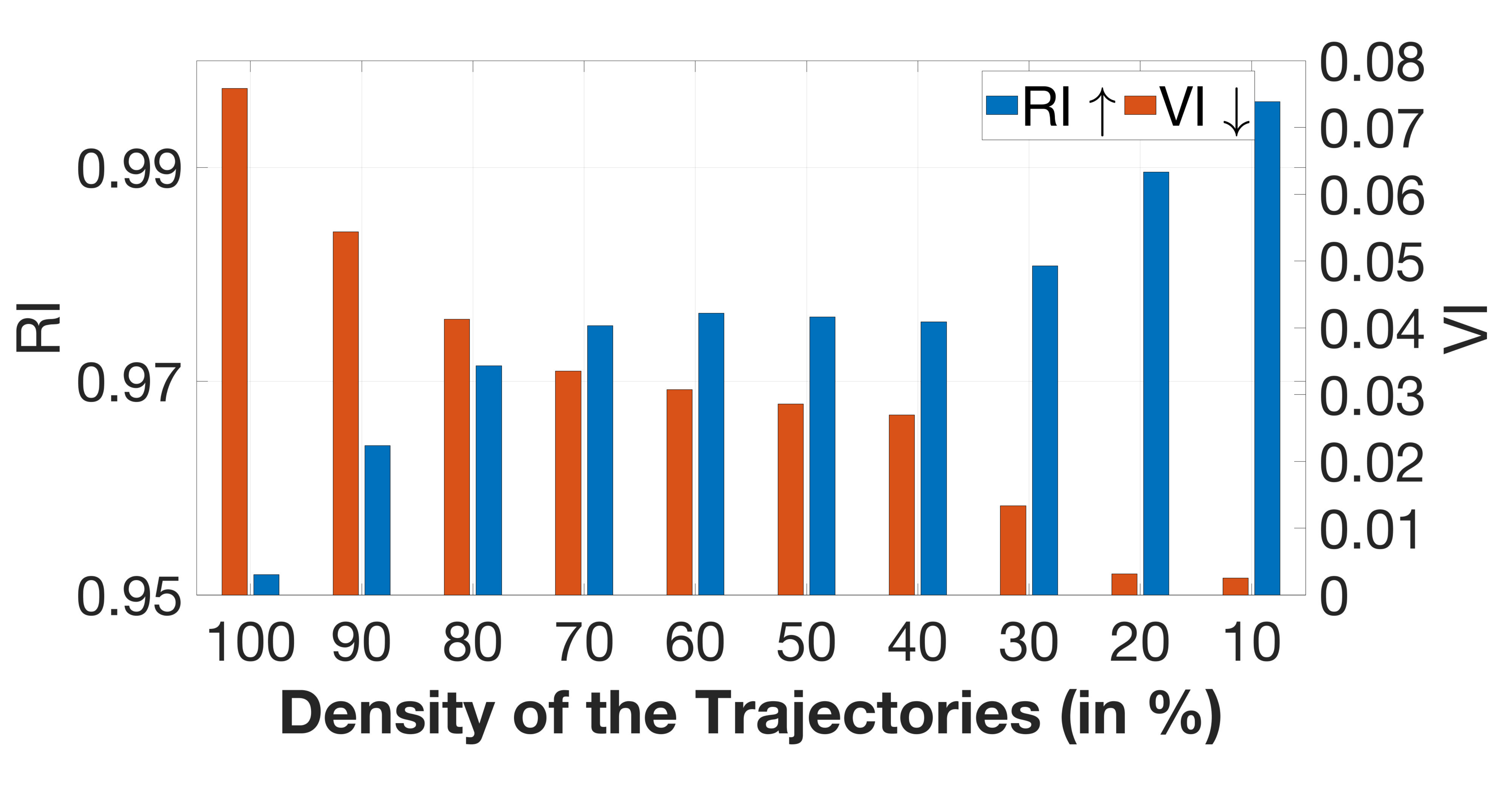}
    \end{tabular}
    \end{center}
    \caption{Study on the trajectory uncertainty on the GAEC~(Keuper et al. [2015b]) solver. The experiment relates to the Variation of Information (VI) and Rand Index (RI) on the train (left) and validation (right) set of DAVIS$_{2016}$~(Perazzi et al. [2016]).}
    \label{fig:davis_ri_vi_gaec}
\end{figure}

\begin{align}
&\mathrm{uncertainty} = \frac{\displaystyle \prod_{e,B} p_c \displaystyle \prod_{e,A} p_j     }{\displaystyle \prod_{e,B} p_c \displaystyle \prod_{e,A} p_j +  \displaystyle \prod_{e,B} p_j \displaystyle \prod_{e,A} p_c}&
\label{eq:uncertainty_eq4_1}
\end{align}
where the nominator is exactly the product of the local probabilities for the observed solution at node $v_i$ (compare Eq. (6) in the main paper) and the denominator sums trivially to one in the case of $|\pazocal{N}_{E'}(v_i)| =1$.

\section{Uncertainty on Minimum Cost Multicut Solutions from GAEC} In Figures \ref{fig:fbms_ri_vi_gaec} and \ref{fig:davis_ri_vi_gaec}, we provide an additional evaluation of the proposed uncertainty measure in the motion segmentation setting. Specifically, we compute solutions for the same motion segmentation problem instances as used in the main paper on the FBMS$_{59}$ and DAVIS$_{2016}$ datasets. While the main paper evaluates using the widely employed high quality solutions from the KLj heuristic, we here additionally assess uncertainties on a faster, lower quality solver, GAEC (Keuper et al. [2015b]). It can be seen that the sparsification plots behave as expected. The VI decreases as the segmentation becomes sparser and the RI increases. However, it can be seen that for the poorer segmentation results, the RI does not increase as monotonically as this was the case for KLj. Specifically when considering the high sparsity regime, the RI metric becomes brittle, indicating that entire labels might have been removed from the solution.

\end{document}